%% file: neurips_2026.tex
\documentclass{article}

\PassOptionsToPackage{numbers,compress}{natbib}
\usepackage[main,final]{neurips_2026}
\usepackage[utf8]{inputenc}
\usepackage[T1]{fontenc}
\usepackage{hyperref}
\usepackage{url}
\usepackage{booktabs}
\usepackage{amsfonts}
\usepackage{nicefrac}
\usepackage{microtype}
\usepackage{graphicx}
\usepackage{subcaption}
\usepackage{multirow}
\usepackage{float}
\usepackage{placeins}
\usepackage{needspace}
\usepackage{fvextra}
\usepackage{amssymb}
\usepackage{mathtools}
\usepackage[table]{xcolor}
\usepackage[ruled,vlined]{algorithm2e}

\title{Causal Attribution via Activation Patching}

\author{%
  {\small
    \textbf{Amirmohammad Izadi}\textsuperscript{1,}%
    \thanks{These authors contributed equally.}\hspace{0.4em},
    \textbf{Mohammadali Banayeeanzade}\textsuperscript{1,*},
    \textbf{Alireza Mirrokni}\textsuperscript{1,*},
    \textbf{Hosein Hasani}\textsuperscript{1,*}
  }
  \\[0.2em]
  {\small
    \textbf{Mobin Bagherian}\textsuperscript{1},
    \textbf{Faridoun Mehri}\textsuperscript{2,3},
    and \textbf{Mahdieh Soleymani Baghshah}\textsuperscript{1}
  }
  \\[0.8em]
  \textsuperscript{1}Computer Engineering Department,
  Sharif University of Technology
  \\
  \textsuperscript{2}Center for Information and Language Processing,
  Ludwig-Maximilians-Universit\"at M\"unchen
  \\
  \textsuperscript{3}Munich Center for Machine Learning (MCML)
}

\begin{document}

\maketitle

\vspace{-2mm} 

\begin{abstract}
Attribution methods for Vision Transformers (ViTs) aim to identify image regions that influence model predictions, but producing faithful and well-localized attributions remains challenging. Existing attribution methods face several limitations, with gradient-based, relevance-propagation, and attention-based methods relying on local approximations, while perturbation or optimization-based methods intervene on inputs, tokens, or surrogates rather than internal patch representations. The key challenge is that class-relevant evidence is formed through interactions between patch tokens across layers; methods that operate only on input changes, attention weights, or backward relevance signals may therefore provide indirect proxies for patch importance rather than directly testing the predictive effect of contextualized patch representations.
We propose \textit{Causal Attribution via Activation Patching (CAAP)}, which estimates the contribution of individual image patches to the ViT’s prediction by directly intervening on internal activations rather than using learned masks or synthetic perturbation patterns. For each patch, CAAP inserts the corresponding source-image activations into a neutral target context over an intermediate range of layers and uses the resulting target-class score as the attribution signal. The resulting attribution map reflects the causal contribution of patch-associated internal representations on the model’s prediction. The causal intervention serves as a principled measure of patch influence by capturing semantic evidence after initial representation formation, while avoiding late-layer global mixing that can reduce spatial specificity.
Across multiple ViT backbones and standard metrics, CAAP consistently outperforms existing methods in various settings and produces more faithful and localized attributions.
\end{abstract}

\vspace{-2mm} 

\section{Introduction}

Vision Transformers (ViTs) have become a dominant architecture for image recognition and representation learning, including supervised and foundation models such as ViT~\cite{vit}, CLIP~\cite{clip}, DINOv2~\cite{dino2} and DeiT3~\cite{deit}.
Their patch-based tokenization and global self-attention naturally motivate attribution methods that operate at the patch or token level rather than on individual pixels~\cite{dosovitskiy2021imageworth16x16words}.
Input attribution methods aim to quantify how specific input features contribute to a model’s prediction and are commonly used to analyze model behavior, identify reliance on spurious cues, and compare learned representations~\cite{samek2021explaining}.
Existing attribution approaches for vision models include gradient-based and relevance-propagation methods~\cite{shrikumar2016not,sundararajan2017axiomatic,grad_cam,chefer2021transformer,mehri-skipplus-cvpr24,Mehri_2025_CVPR}, attention-flow and token-mixing methods~\cite{attn_r,tam,qiang2022attcat,mutex}, and perturbation or optimization-based methods that modify inputs, tokens, attention mechanisms, or patch embeddings~\cite{Petsiuk2018RISE,fong2017meaningful,fong2019extremal,Englebert_2023_ICCV,deb2023atman,vit_cx,mda}.
Despite this broad progress, producing attribution maps that are both faithful to ViT predictions and spatially localized remains challenging.

\begin{figure}[t]
  \centering
  \includegraphics[width=0.9\linewidth]{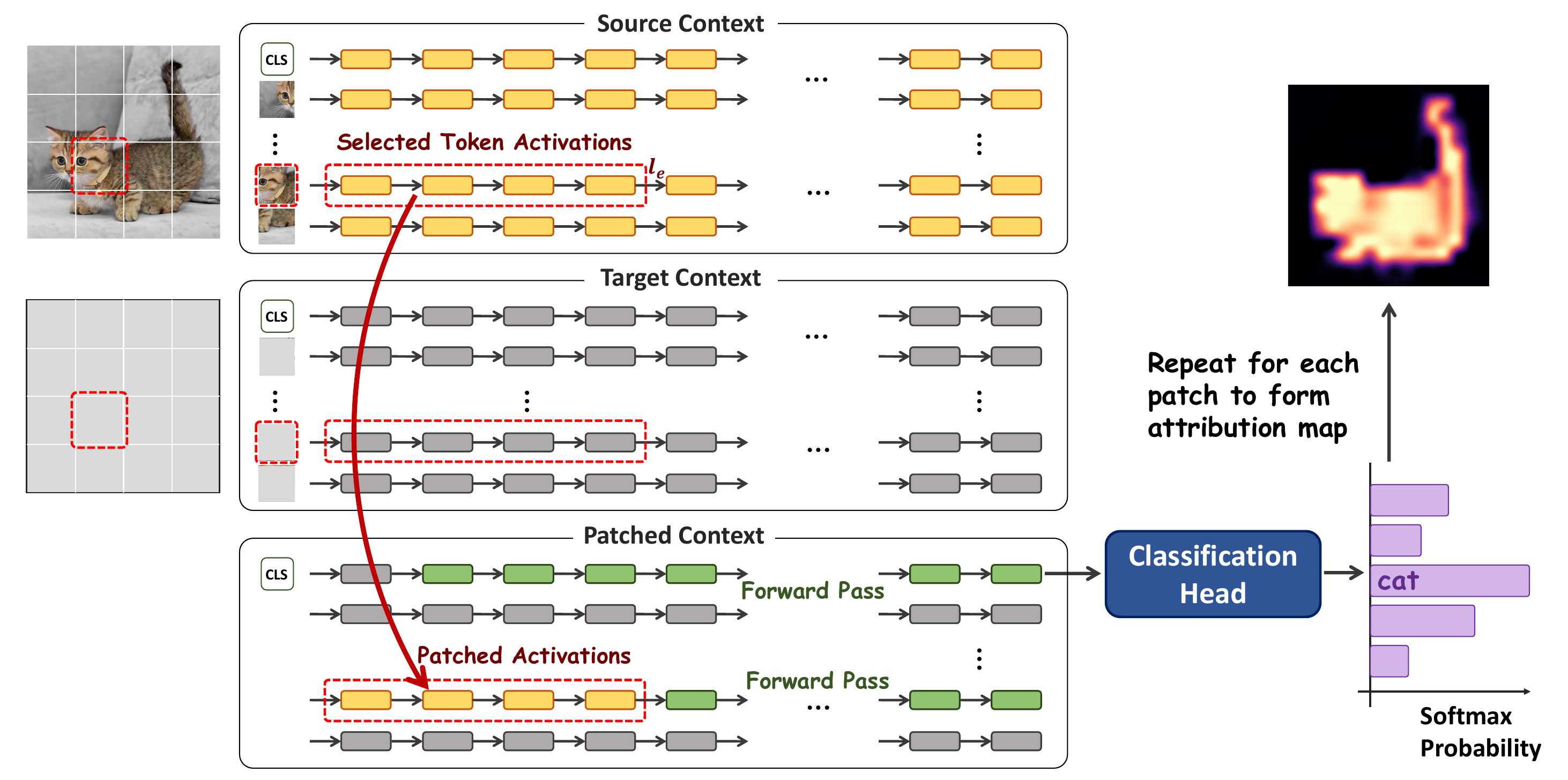}
  \vspace{-1.0mm}
    \caption{
    Overview of CAAP. Given a source image (top) and a blank target image (middle), we extract internal activations from the source image corresponding to a selected patch and a specified range of layers. These activations are injected into the target context to form a patched context (bottom), while all other activations remain unchanged. A forward pass on the patched sequence produces a class score for the CLS token (green), which measures the causal effect of the selected patch on the target prediction. Repeating this intervention for all patches yields a patch-level attribution.
    }
  \label{fig:graphical_abstract}
    \vspace{-3.0mm}
\end{figure}


A central difficulty in ViT attribution is that patch relevance is not determined only by the raw visual content of a patch, but by contextualized representations formed through self-attention across layers. This creates limitations for several attribution families. Gradient-based and relevance-propagation methods provide efficient class-specific signals, but they measure local sensitivity or propagate relevance through the computation graph rather than directly testing whether a contextualized patch representation is sufficient to support the prediction~\cite{shrikumar2016not,sundararajan2017axiomatic,grad_cam,chefer2021transformer,Mehri_2025_CVPR}. Attention-based methods expose token-mixing patterns, but attention weights or attention flow are indirect proxies for feature contribution, especially after information from different tokens has been repeatedly mixed across layers~\cite{attn_r,tam,qiang2022attcat}. 

Perturbation and masking methods measure causal output changes under interventions, but input-space perturbations can introduce distribution shift~\cite{Petsiuk2018RISE,fong2017meaningful,fong2019extremal,Englebert_2023_ICCV}. Input-based deletion does not account for redundancy across patches: when multiple patches encode similar class evidence, removing each patch from a group of redundant patches may cause little change in the output, leading to underestimation of the group’s importance. Insertion into a blank or blurred image overlooks interactions among patches and may fail to reproduce the functional role a patch plays in the full image context. More ViT-specific perturbation or optimization methods operate over tokens, attention mechanisms, patch embeddings, or attribution metrics~\cite{deb2023atman,vit_cx,mda}, but they still do not directly evaluate the class evidence carried by the original internal patch representation within a controlled target context.

In this work, we propose \textit{Causal Attribution via Activation Patching (CAAP)}, a novel attribution method motivated by mechanistic interpretability. Rather than attributing through backward sensitivity, attention propagation, input masking, or optimized surrogate maps, CAAP directly intervenes on the hidden activations of a trained ViT. For each patch, we insert the corresponding source-image activations into a neutral blank context and use the resulting target-class score as the attribution signal~\cite{ZhangNanda2024Patching}.
Crucially, we apply activation patching up to the intermediate layers: late enough that patch representations contain enriched class-relevant object information, but early enough to avoid late-layer contextual mixing which reduces spatial specificity. This design allows CAAP to better isolate the causal contribution~\cite{causality,chattopadhyay2019causal} of class-relevant patch representations to the final prediction.

Causal intervention offers a practical advantage for CAAP over existing attribution techniques. It derives patch scores directly from information already encoded in contextualized token representations, rather than relying on indirect evidence such as attention weights, approximate gradient signals, input-space masks, or metric-optimized surrogate maps.
The procedure can be efficiently parallelized, enabling practical patch-level attribution for large ViT backbones.
Empirical results across multiple Vision Transformer backbones show that our method improves over existing approaches on most evaluated settings, achieving state-of-the-art results on deletion, insertion, and foreground and background localization metrics. Quantitative and qualitative results demonstrate that this method provides a reliable and effective approach to spatial attribution in ViTs, offering a new perspective based on mechanistic interpretability tools.

\vspace{-2.0mm}
\section{Preliminaries}

\vspace{-1.0mm}
\subsection{Problem Definition}

We consider an image classification model $f$ that maps an image $x \in \mathbb{R}^{C \times H \times W}$ to the class scores $f(x) \in \mathbb{R}^K$, where $K$ is the number of classes. For a target class $y$, we denote the corresponding prediction score by $p(y \mid x) = f(x)_y$. The image $x$ is partitioned into a set of non-overlapping visual patches $\mathcal{P} = \{p_1, \dots, p_N\}$, where each patch corresponds to a spatial region in the input.
The goal of visual attribution is to assign a scalar attribution score $A(p_i) \in \mathbb{R}$ to each patch $p_i$, measuring the contribution of that patch to the prediction $f(x)_y$. An attribution method produces an attribution map $A \in \mathbb{R}^N$ such that higher values indicate greater relevance to the target prediction. 

\vspace{-2.0mm}
\subsection{Activation Patching}

Let $f$ be a neural network composed of $L$ layers. For an input $x$, we denote by $h^l(x)$ the activation tensor at layer $l \in \{1, \dots, L\}$. Activation patching is an intervention-based procedure that modifies a forward pass by replacing selected internal activations of one input with those from another, while keeping all other computations unchanged. We refer to the input providing the replaced activations as the source context $c$, and the input receiving these activations as the target context $c'$.

The patched context $c^*$ is first initialized with the target context $c'$. Given a set of layers $\mathcal{L}$ and selection operator $\mathcal{S}$ over token indices, the patched context $c^*$ is then constructed by setting
\begin{equation}
h^l(c^*)_{\mathcal{S}} = h^l(c)_{\mathcal{S}}, \quad h^l(c^*)_{\neg \mathcal{S}} = h^l(c')_{\neg \mathcal{S}}, \quad \forall l \in \mathcal{L},
\end{equation}
while all remaining activations are propagated according to the model’s standard forward computation. The model output $f(c^*)$ reflects the effect of intervening on the selected internal activations of $c$, enabling causal analysis of their influence on the final prediction under activation interventions~\cite{causality}.

\vspace{-1.0mm}
\section{Method}
\vspace{-2.0mm}
We propose CAAP for ViT models, a patch-level attribution method that assigns relevance scores by intervening on the internal activations of a trained classifier. Rather than measuring local sensitivity at the input level, CAAP quantifies the causal effect of internal representations associated with a visual patch on the final prediction. This addresses a core limitation of input-level insertion and deletion methods. Deletion-based methods can underestimate importance when class-relevant information is redundant and spread across multiple patches, and insertion into a blank image produces an out-of-distribution (OOD) input and ignores token interactions across transformer layers.

Let $x$ be the input image and $x_0$ a blank image of the same resolution, used as a neutral target context that contains no class-specific visual evidence. We treat $x$ as the source context ($c$) and $x_0$ as the target context ($c'$). For each visual patch $p_i \in \mathcal{P}$, we identify the corresponding spatial indices in the activation tensors and define a patch selection operator $\mathcal{S}(p_i)$. For a chosen set of layers $\mathcal{L} = \{l_s, \dots, l_e\}$, the patched context $c^*(p_i)$ is constructed by replacing activations as
\begin{equation}
h^l(c^*(p_i))_{\mathcal{S}(p_i)} = h^l(x)_{\mathcal{S}(p_i)}, \quad
h^l(c^*(p_i))_{\neg \mathcal{S}(p_i)} = h^l(x_0)_{\neg \mathcal{S}(p_i)}.
\end{equation}

The intervention is performed on the patch-token residual
stream at the selected layers.
The transferred residual activations are then used by the
subsequent attention and FFN computations.
Starting from layer $l_s+1$, we resume the forward computation by propagating the $\mathrm{CLS}$ token through all subsequent layers up to the output, and after layer $l_e+1$ we also continue the forward computation for the patched tokens $\mathcal{S}(p_i)$ so that their residual streams are updated consistently with the intervened activations.
The attribution score for patch $p_i$ with respect to class $y$ is defined as the prediction score obtained from the patched context,
\begin{equation}
\label{eq:attr}
A(p_i) = f(c^*(p_i))_y.
\end{equation}
This score measures how predictive the contextualized token embedding associated with patch $p_i$ is for the target class, by inserting it into a neutral context and observing its effect on the output. Repeating this procedure for all patches yields a patch-level attribution map over the input image.

An important property of activation patching, compared to input-based perturbation methods, is that the patched activations are taken directly from the original context, so they already encode interactions between patches and avoid introducing OOD inputs in the source context.
The overall procedure of the proposed method is summarized in Algorithm~\ref{alg:cap} and shown in Fig.~\ref{fig:graphical_abstract}.

\begin{algorithm}[t]
\footnotesize
\caption{\small CAAP: \underline{C}ausal \underline{A}ttribution via \underline{A}ctivation \underline{P}atching}
\label{alg:cap}
\KwIn{
Classifier $f$; image $x$; blank image $x_0$; target class $y$;
patch set $\mathcal{P}=\{p_1,\dots,p_N\}$;
intervened layers $\mathcal{L}$; selection operator $\mathcal{S}(\cdot)$.
}
\KwOut{Patch attribution map $A \in \mathbb{R}^N$}

Compute and cache activations $\{h^l(x)\}_{l=1}^L$ and $\{h^l(x_0)\}_{l=1}^L$\;

\For{$i \leftarrow 1$ \KwTo $N$}{
Initialize patched context from target activations:
\vspace{-1.0mm}
\[
h^l(c^*) \leftarrow h^l(x_0), \quad \forall l \in \{1,\dots,L\}
\]

Replace source activations at selected indices:
\vspace{-1.0mm}
\[
h^l(c^*(p_i))_{\mathcal{S}(p_i)} \leftarrow h^l(x)_{\mathcal{S}(p_i)}, 
\quad \forall l \in \mathcal{L}
\]

Resume forward computation from layer $l_s+1$ to $L$:\\
\quad Propagate the $\mathrm{CLS}$ token through all layers $l_s+1,\dots,L$,\\
\quad and propagate the patched tokens $\mathcal{S}(p_i)$ through layers $l_e+1,\dots,L$.\\

Compute the target-class score:
\vspace{-1.0mm}
\[
A(p_i) \leftarrow f(c^*(p_i))_{y}.
\]
}
\Return{$A$}\;
\end{algorithm}

\vspace{-1.0mm}
\subsection{Patch Selection Operator $\mathcal{S}$}
\label{sec:selection_op}
In the naive implementation, the patch selection operator $\mathcal{S}(p_i)$ corresponds exactly to the patch $p_i$. However, patching a single token often does not transfer sufficient predictive information from the source context. To account for spatial feature spread across neighboring tokens, we consider a padded variant in which $\mathcal{S}(p_i)$ includes the patch $p_i$ and its immediate spatial neighbors. This choice preserves both contextual information and spatial locality, making the patch-level intervention a more faithful measure of localized contribution. 

\subsection{Choice of Intervened Layers $\mathcal{L}$}
\label{sec:layer_choice}

ViTs build patch representations through repeated self-attention, causing each patch token to encode information aggregated from other patches.
This perspective suggests distinguishing two representation regimes in ViTs~\cite{pan2024dissecting}. First, patch tokens undergo a grouping process in which class-consistent and object-related features become organized across patches.
Second, tokens increasingly encode broader context through interactions that may involve tokens from different objects or background regions. For most pretrained ViTs, intra-object grouping occurs in the early layers, while inter-object or background contextualization emerges in the later layers~\cite{pan2024dissecting}.

In the context of attribution, early-stage grouping is desirable because it strengthens class-related evidence by integrating dependencies between adjacent patches. In contrast, late-stage contextualization can reduce spatial specificity, as class-relevant evidence becomes entangled with global scene structure and background information.
Therefore, a reasonable choice of $\mathcal{L}$ is to include the early and intermediate layers, up to the point where background or other non-class-related features begin to mix into the patch representations. We next provide further insight into grouping and mixing mechanisms through the self-attention operation.

Consider a Vision Transformer with $N$ patch tokens. Let $z_i^l \in \mathbb{R}^d$ denote the representation of patch $i$ at layer $l$. In a single attention head, the attention update for patch $i$ can be written as
\begin{equation}
\label{eq:attn_update}
\tilde{z}_i^{l+1}
= \sum_{j=1}^N \alpha_{ij}^l v_j^l,
\end{equation}
where $v_j^l$ are value vectors derived from the token representations and $\alpha_{ij}^l$ are attention weights defined by
\begin{equation}
\label{eq:attn_softmax}
\alpha_{ij}^l
= \frac{\exp\left( \langle q_i^l, k_j^l \rangle / \sqrt{d} \right)}
{\sum_{m=1}^N \exp\left( \langle q_i^l, k_m^l \rangle / \sqrt{d} \right)}.
\end{equation}
Let $\mathcal{O}$ denote the set of patches belonging to the same object as patch $i$. The attention update can be decomposed as
\begin{equation}
\tilde{z}_i^{l+1}
=
\sum_{j \in \mathcal{O}} \alpha_{ij}^l v_j^l
+
\sum_{j \notin \mathcal{O}} \alpha_{ij}^l v_j^l.
\end{equation}

We empirically show that in the early layers, where intra-object grouping occurs, the quantity 
$\mathbb{E}_{j \in \mathcal{O}}[\alpha_{ij}^l] - \mathbb{E}_{j \notin \mathcal{O}}[\alpha_{ij}^l]$ 
increases, while after the middle layers it decreases.
This enrichment of intra-object grouping supports the motivation for layer-wise activation patching rather than input-level insertion. However, it also suggests that patching up to the final layer is not desirable, due to increasing extra-object contextualization.
In practice, we select the intervened layers from the beginning up to $\lceil\frac{2}{3} L\rceil$, where patch representations contain strong object-level information while preserving spatial specificity.

\vspace{-1.0mm}
\subsection{Efficient Implementation}
\label{sec:efficient}

In transformers, hidden states at each layer are linearly projected to queries, keys, and values. To avoid recomputing these vectors from hidden states during activation patching, we reuse cached key and value projections associated with the intervened hidden states, rather than recomputing them from residual streams. Only the query vectors of the additional $\mathrm{CLS}$ tokens are computed dynamically in the patched context. We next propose two practical techniques to further accelerate the intervention process: parallel computation of $\mathrm{CLS}$ tokens and precomputation of blank-context attention.

The patch-wise formulation requires one forward pass per patch, which scales linearly with the number of tokens in the image. Since the intervention operates on internal activations, we can compute all patch attribution scores in parallel within a single forward pass while preserving the same causal intervention. This construction exploits the structure of activation patching and is not applicable to input-space masking methods, which require independent forward passes for each modified input.

For the blank image $x_0$, we perform one forward pass and cache its activations $\{h^l(x_0)\}_{l \in \mathcal{L}}$, reused across samples. For each source image $x$, we run a standard forward pass and cache $\{h^l(x)\}_{l \in \mathcal{L}}$. We then build a patched computation containing both activation sets. For each patch $p_i$, we introduce a dedicated classification token $\mathrm{CLS}_i$ with the same positional encoding as the original $\mathrm{CLS}$. The attention mask ensures that $\mathrm{CLS}_i$ attends to source activations at $\mathcal{S}(p_i)$ and to target activations at $\neg \mathcal{S}(p_i)$. No attention is allowed between different classification tokens, and only the $\mathrm{CLS}$ tokens are propagated through the remaining layers. The resulting class score from $\mathrm{CLS}_i$ for class $y$ equals the attribution score in Eq.~\ref{eq:attr} from patch-wise activation replacement for $p_i$.

This parallel construction of $\mathrm{CLS}$ tokens yields the same attribution map (no approximation) while computing all patch scores in parallel. The total cost per sample is one forward pass for the source image and one forward pass with multiple $\mathrm{CLS}$ tokens. In addition, we introduce an extra efficiency method in Appendix~\ref{app:efficiency}, based on the fact that a large fraction of the computation comes from attention between each $\mathrm{CLS}$ token and blank-context tokens. We exploit the near-invariance of these interactions to precompute their contribution and reuse it during attribution. This approximation reduces computation without notable effect on performance (details and results are in Appendix~\ref{app:efficiency}).

While the efficient implementation described above relies on
CLS tokens, CAAP is not limited to CLS-based architectures.
For models with other readout mechanisms, we can apply
activation patching and use the model's original readout
to obtain attribution scores, as demonstrated in
Appendix~\ref{app:cr_readouts}.

\section{Experiments}
\vspace{-0.1mm}
\label{sec:experiments}
In this section, we present qualitative and quantitative evaluations of CAAP. We begin by defining the evaluation metrics used throughout the experiments, followed by a detailed description of the ViT models, datasets, and baseline methods. We then present qualitative and quantitative results, along with ablation studies analyzing the contribution of each individual component of CAAP.

\subsection{Evaluation Metrics}
\label{sec:metrics}

We evaluate attribution maps using six metrics that assess two complementary aspects of quality: faithfulness and localization.
Deletion AUC and Insertion AUC~\cite{Petsiuk2018RISE} are faithfulness measures; they test whether pixels ranked as important by the attribution are indeed the ones the model relies on.
To obtain a single faithfulness metric, we also report Insertion$-$Deletion, defined as the difference between Insertion AUC and Deletion AUC.
AUPR$_1$, AUPR$_0$~\cite{Margolin2014Foreground}, and Pointing-Game (PG)~\cite{bearman2016whatspointsemanticsegmentation} are localization measures; they quantify how well attribution mass concentrates on annotated foreground regions and avoids background regions. See Appendix~\ref{app:compactness} for additional
metrics that are less common in the literature.

\paragraph{Faithfulness metrics. }
Following the evaluation protocol of~\cite{Petsiuk2018RISE}, given an attribution map $A$ for an image $x$ and corresponding class $y$, we sort pixels (patches) in descending attribution order $A$ and perturb $x$ accordingly.
For Deletion, we progressively replace the top-ranked pixels with a constant reference value and track the target class score $f(\cdot)_y$; the area under this curve is Deletion AUC.
For Insertion, we start from a blurred version of $x$ and progressively reveal (unblur) the top-ranked pixels, again tracking $f(\cdot)_y$; the area under this curve is Insertion AUC.
More faithful attributions yield lower Deletion AUC, higher Insertion AUC and higher Insertion$-$Deletion.

\paragraph{Localization metrics. }
Given a normalized attribution map $A\in[0, 1]^{H\times W}$ and a ground-truth binary mask $M\in\{0,1\}^{H\times W}$, AUPR treats $A$ as a pixel-wise confidence map and sweeps a threshold to form a precision--recall curve.
AUPR$_1$ defines positives as foreground pixels, while AUPR$_0$ defines positives as background pixels by evaluating background confidence $1-A$ against the inverted mask $1-M$.
PG~\cite{bearman2016whatspointsemanticsegmentation} checks whether the maximum-attribution pixel $(i^*,j^*)=\arg\max_{i,j}A_{i,j}$ falls inside the annotated foreground region.
The PG score is defined as the fraction of hits over all images.
Higher AUPR$_1$, AUPR$_0$, and PG indicate better spatial alignment with the ground-truth regions.

\subsection{Experiment Settings}
\label{sec:exp_settings}
\vspace{-1.0mm}
\paragraph{Models and datasets.}
We conduct our main experiments on the ImageNet-1K (ILSVRC2012) validation set~\cite{russakovsky2015imagenetlargescalevisual}. 
We evaluate four ViT-L models on ImageNet-1K: CLIP-L/14~\cite{clip}, DINOv2-L/14~\cite{dino2}, ViT-L/16~\cite{vit}, and DeiT3-L/16~\cite{deit}. 
For localization evaluation on ImageNet-1K, we use ImageNet-S~\cite{imagenet_s} annotations and report AUPR$_1$, AUPR$_0$, and PG. 
Results for additional architectures are provided in Appendix~\ref{app:experiments}.
To further validate CAAP, we additionally evaluate on ImageNet-Hard~\cite{imagenet_hard}, Oxford-IIIT-Pet~\cite{oxford_pet}, and Food-101~\cite{food101}. 
For these datasets, we use a ViT-B/16 classifier initialized from pretrained weights.

\vspace{-2.0mm}
\paragraph{Baselines.}
We compare CAAP against a strong set of widely-used attribution baselines drawn from various attribution categories: gradient-based methods, including Grad-CAM~\cite{grad_cam}, IxG+~\cite{mehri-skipplus-cvpr24}, and Libra IxG+~\cite{Mehri_2025_CVPR}; attention-based methods, including Attention Rollout~\cite{attn_r}, Transformer Attribution (T-Attr)~\cite{chefer2021transformer}, TAM~\cite{tam}, and MUTEX~\cite{mutex}; and perturbation-based methods, including RISE~\cite{Petsiuk2018RISE}, ViT-CX~\cite{vit_cx}, and MDA~\cite{mda}.
We evaluate all methods using the same protocol and metrics.

\vspace{-2.0mm}

\subsection{Quantitative Results}
\vspace{-2.0mm}
\paragraph{Faithfulness.}
Table~\ref{tab:faithfulness_comparison_results_main} reports faithfulness metrics across four ViT models. Table~\ref{tab:faithfulness_comparison_other_datasets} further presents the same faithfulness metrics on ImageNet-Hard~\cite{imagenet_hard} and Oxford-IIIT-Pet~\cite{oxford_pet}.
CAAP consistently achieves the lowest Deletion AUC and the highest or competitive Insertion AUC on all backbones.
In particular, CAAP improves the Insertion$-$Deletion with a notable margin compared to prior methods, indicating that the regions ranked as important by our attribution maps more accurately reflect the model’s true decision-making process.

\input{tables/main_imagenet}

\input{tables/other_datasets}

\vspace{-2.0mm}
\paragraph{Localization.}
Localization results are reported in Table~\ref{tab:localization_comparison_results_main}.
CAAP achieves the best overall performance across most backbones. In particular, the higher PG and AUPR$_1$ values indicate that the regions identified as most relevant are consistently located within the annotated foreground areas.
Moreover, strong AUPR$_0$ scores demonstrate effective suppression of background activations.
These results confirm that CAAP produces spatially precise attribution maps that better align with ground-truth object regions. Further quantitative results are provided in Appendix~\ref{app:more_quantitative}, and evaluation with Infidelity and compactness metrics
is reported in Appendix~\ref{app:compactness}.

\begin{figure}[t]
  \centering
  \includegraphics[width=0.99\linewidth]{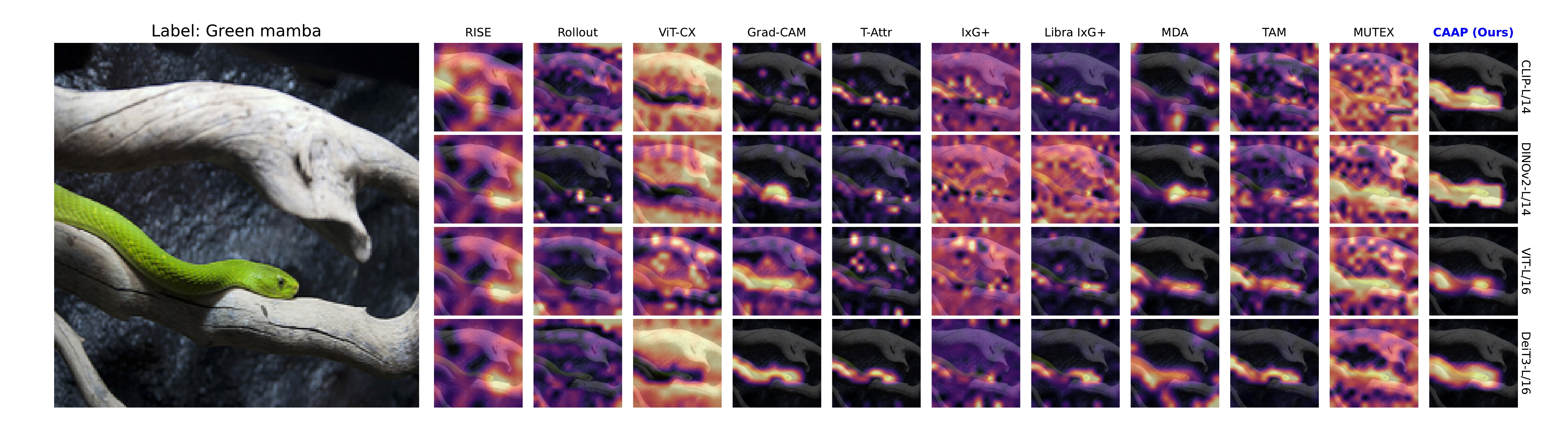}
  \vspace{-1.0mm}
  \caption{
   Qualitative attribution comparison between different methods for a representative ImageNet sample. CAAP produces more compact and well-localized attributions for the target class than the baselines. More single-object qualitative examples are provided in the Appendix~\ref{app:more_qualitative} (Fig.~\ref{fig:vis_11}--~\ref{fig:vis_381}).
    }
  \label{fig:vis_snake}
    \vspace{-3.0mm}
\end{figure}

\begin{figure}[h]
  \centering
  \includegraphics[width=0.99\linewidth]{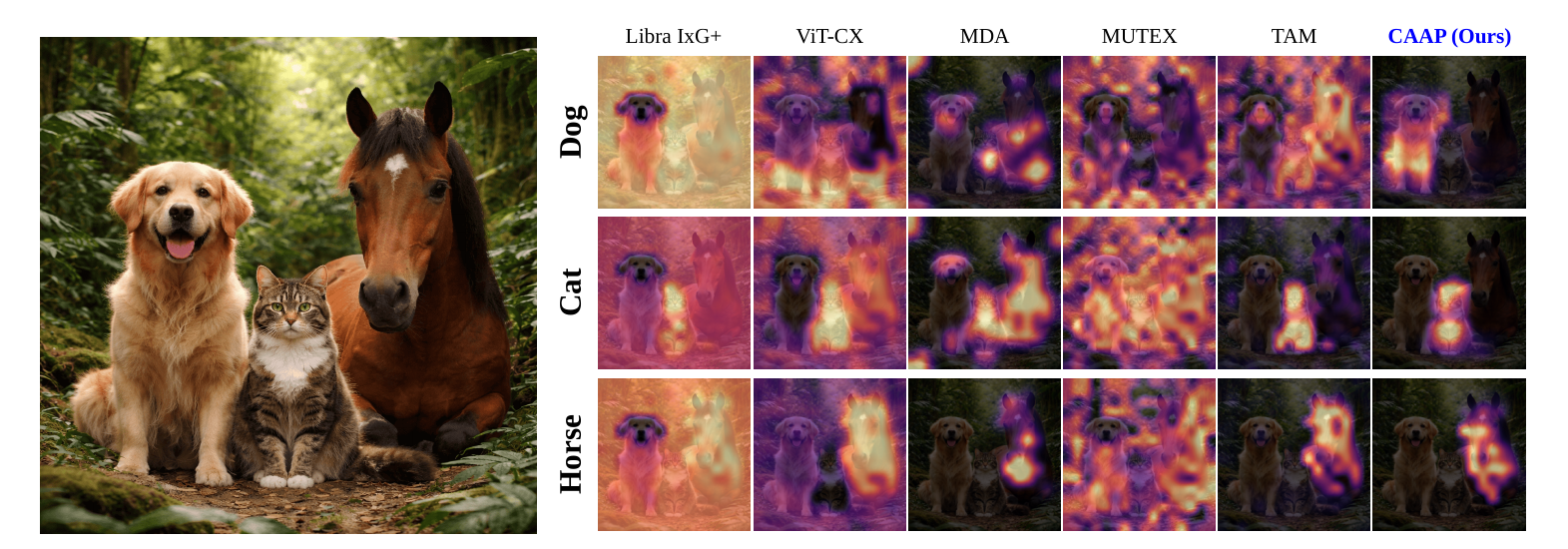}
  \vspace{-1.0mm}
  \caption{
    Qualitative attribution comparison between different methods for an image containing multiple objects. CAAP produces compact and class-specific attributions aligned with the queried object class. More multi-object qualitative examples are provided in Appendix~\ref{app:more_qualitative} (Fig.~\ref{fig:dolphin_shark_1}--\ref{fig:totoro_1}).
    }
  \label{fig:vis_multi}
    \vspace{-3.0mm}
\end{figure}

\vspace{-1.0mm}
\subsection{Qualitative Results}
Qualitative results are provided in Figs.~\ref{fig:vis_snake} and~\ref{fig:vis_multi}. As illustrated, baseline methods often produce diffuse or fragmented attribution maps that extend beyond the target object into the background, or highlight only a small subset of discriminative patches. In contrast, CAAP tends to generate more compact, coherent, and semantically meaningful attributions that align closely with the target object.
Specifically, CAAP's attributions better capture object boundaries and maintain consistency across patches, resulting in fewer activations on irrelevant background regions. The improved localization quality observed in the visualizations is consistent with our superior AUPR$_1$, AUPR$_0$ and PG scores reported in Table~\ref{tab:localization_comparison_results_main}.
Additional qualitative results are provided in Appendix~\ref{app:more_qualitative}.

\vspace{-1.0mm}
\setcounter{topnumber}{3}
\subsection{Ablation Studies}
\label{sec:ablation}
We ablate four aspects of CAAP on ImageNet: (i) the choice of target context, (ii) the type of the selection operator, (iii) the choice of intervened layers, and (iv) comparison with input-based interventions. 
Results are shown for CLIP-L/14, while results on other ViT backbones (ViT-L/16, DINOv2-L/14, DeiT3-L/16) are provided in Appendix~\ref{app:models_ablation}.
Additional results comparing CAAP with input-based
approaches on convolutional architectures are provided
in Appendix~\ref{app:cnn}, further demonstrating the
effectiveness of CAAP beyond ViTs.

\begin{figure}[t]
  \centering
  \begin{subfigure}[t]{0.49\linewidth}
    \centering
    \includegraphics[width=\linewidth,height=3.6cm]{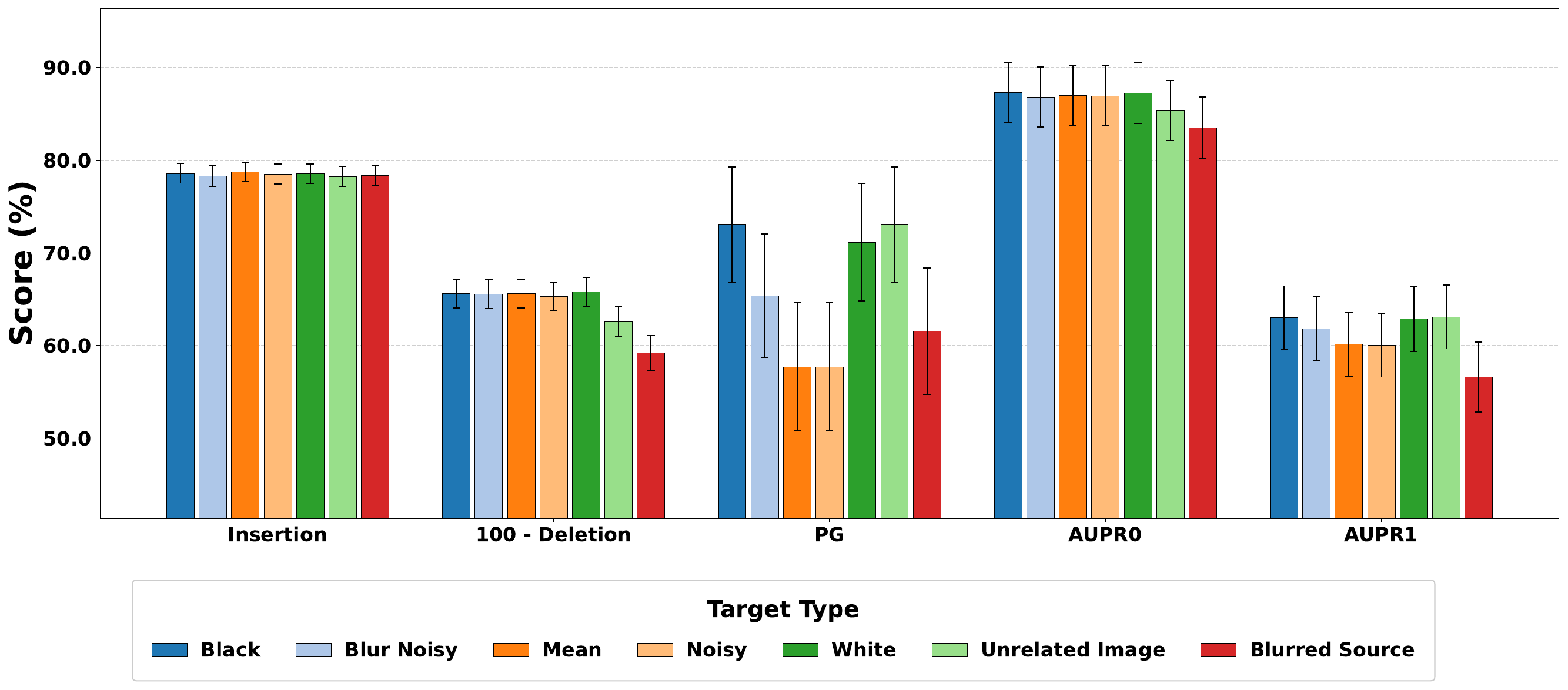}
    \caption{Target context type.}
    \label{fig:clip_ablation_target}
  \end{subfigure}
  \hfill
  \begin{subfigure}[t]{0.49\linewidth}
    \centering
    \includegraphics[width=\linewidth,height=3.6cm]{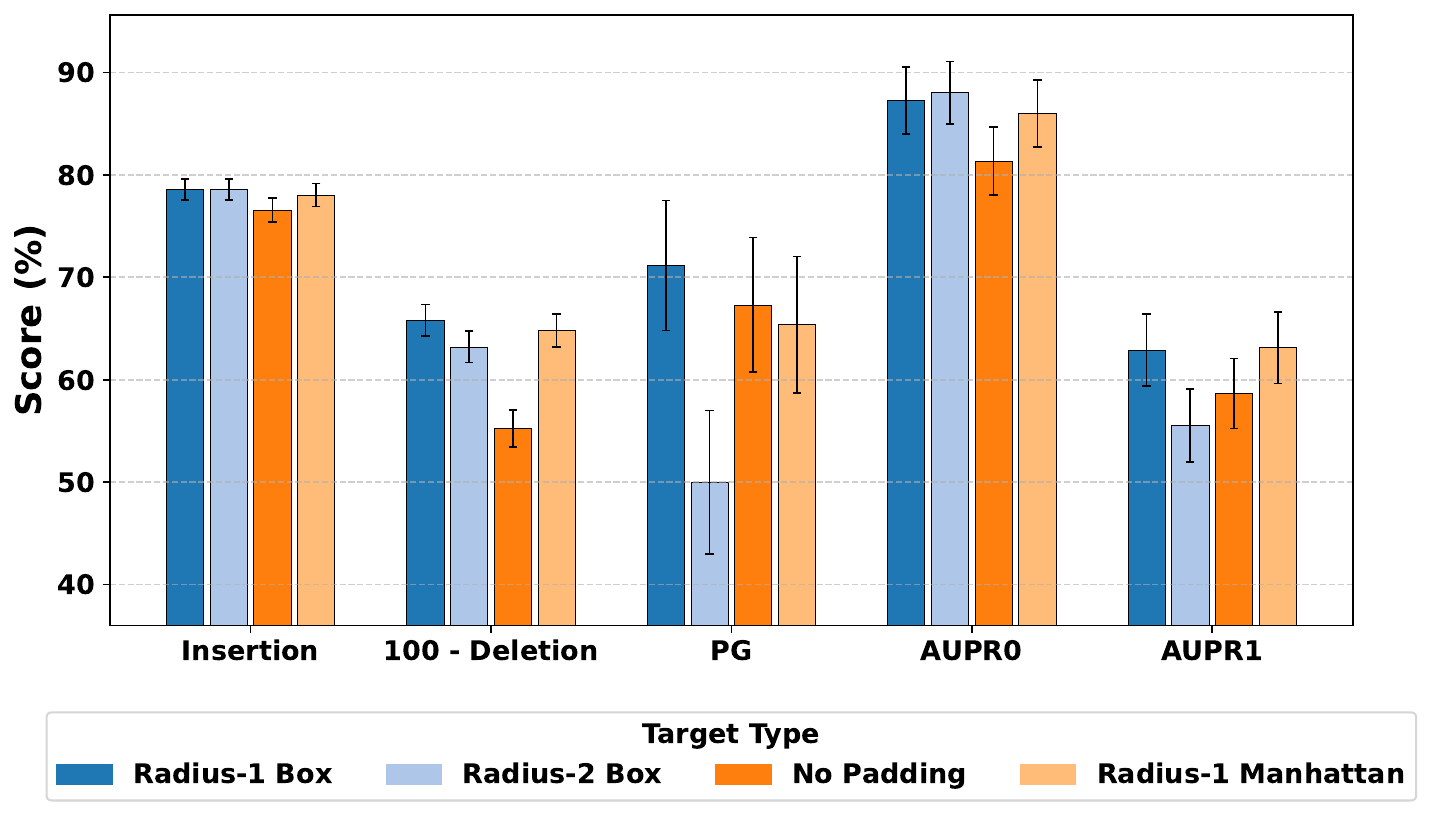}
    \caption{Selection operator (padding).}
    \label{fig:clip_ablation_pad}
  \end{subfigure}
  \vspace{-1mm}
    \caption{
    Ablation studies of activation patching context
    characteristics on ImageNet using CLIP-L/14.
    (a) Effect of target context type (Black, Blur Noisy,
    Mean, Noisy, White, Unrelated Image, Blurred Source).
    (b) Effect of spatial support in the selection operator
    (No Padding, Radius-1 Box, Radius-2 Box,
    Radius-1 Manhattan).
    Insertion ($\uparrow$), $100-\mathrm{Deletion}$
    ($\uparrow$), PG ($\uparrow$), AUPR$_0$ ($\uparrow$),
    and AUPR$_1$ ($\uparrow$) are reported.
    Results on other ViT backbones are provided in
    Appendix~\ref{app:models_ablation}.
    }
  \label{fig:clip_ablation_main}
  \vspace{-3mm}
\end{figure}

\paragraph{Target context type.}
We first ablate the type of target context used when
constructing the patched target, varying the content
among Black, White, Mean, Noisy, Blur Noisy, Unrelated
Image, and Blurred Source. Here, Black and White use
all-black and all-white blank patches, respectively,
Mean fills the blanks with the dataset per-channel
mean, Noisy fills them with i.i.d.\ Gaussian noise,
and Blur Noisy adds i.i.d.\ Gaussian noise to
the target and then applies a blur filter. Unrelated
Image uses a different real image as the target, while
Blurred Source uses a blurred version of the source
image. Fig.~\ref{fig:clip_ablation_target} shows that
faithfulness is largely insensitive to the choice of
target context, with Blurred Source showing slightly
weaker performance. AUPR$_0$ and AUPR$_1$ vary only
marginally across most target types, while PG shows
more noticeable differences, particularly for Blurred
Source. The weaker localization performance of Blurred
Source is likely due to residual information from other
patches leaking into the patched context.
We use White as the default setting in all
experiments.

\paragraph{Patch selection operator.}
We next ablate the spatial support of the intervention, motivated by the fact that ViT patch tokens absorb information from nearby patches via self-attention.
Specifically, for a candidate patch at grid location $(i,j)$,
we consider the following four variants:
\begin{itemize}
    \setlength{\itemsep}{0pt}
    \setlength{\topsep}{2pt}
    \setlength{\parsep}{0pt}
    \item \textbf{No Padding:} Only the candidate patch $(i,j)$.
    \item \textbf{Radius-1 Box:} A $3\times3$ neighborhood,
    $\{(u,v): |u-i|\leq1,\ |v-j|\leq1\}$.
    \item \textbf{Radius-2 Box:} A $5\times5$ neighborhood,
    $\{(u,v): |u-i|\leq2,\ |v-j|\leq2\}$.
    \item \textbf{Radius-1 Manhattan:} The candidate patch and
    its four direct neighbors,
    $\{(u,v): |u-i|+|v-j|\leq1\}$.
\end{itemize}
All selections are restricted to valid patch-grid indices.
Fig.~\ref{fig:clip_ablation_pad} compares four neighborhood variants. The no-padding variant performs the worst, while all padded alternatives improve both faithfulness and localization. This aligns with our selection operator design (Section~\ref{sec:selection_op}), indicating that patching only a single token into a large number of blank patches often does not provide sufficient signal, while spatial padding strengthens the informative signal and improves attribution quality. We use the Radius-1 Box setting, which corresponds to a $3\times3$ region, as the default in our experiments.

\begin{figure}[t]
  \centering
  \begin{subfigure}[t]{0.49\linewidth}
    \centering
    \includegraphics[width=\linewidth]{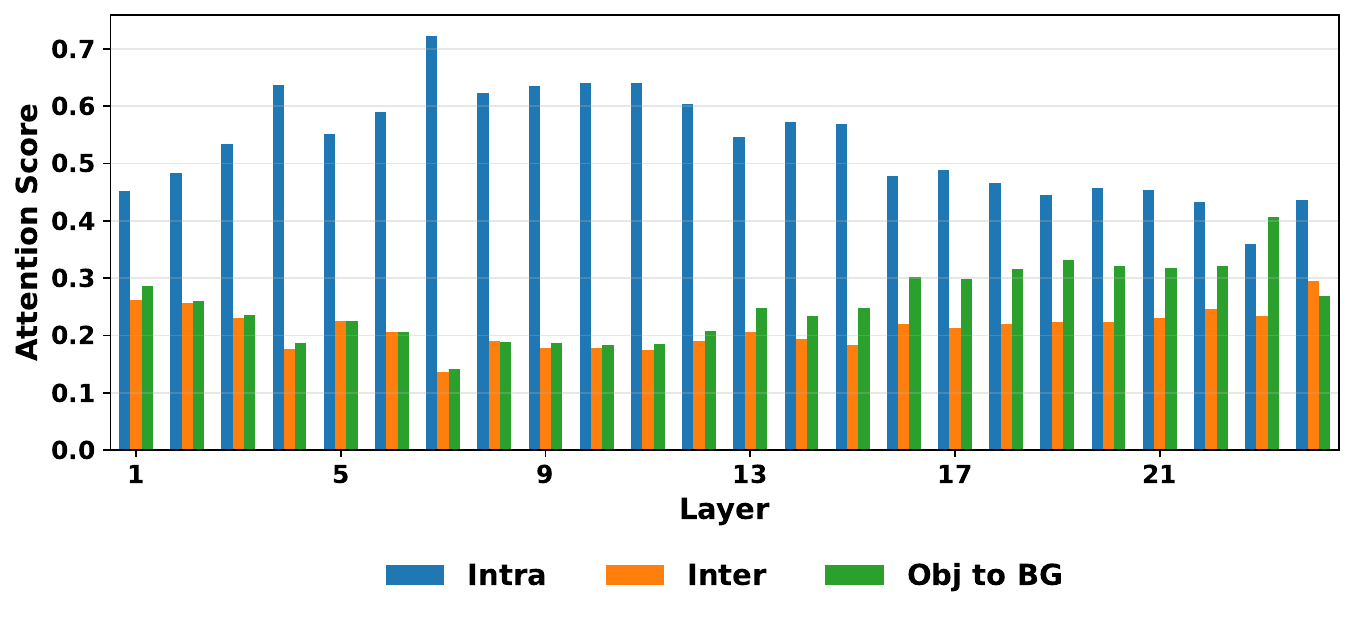}
    \caption{Attention statistics across layers.}
    \label{fig:clip_attn}
  \end{subfigure}
  \hfill
  \begin{subfigure}[t]{0.49\linewidth}
    \centering
    \includegraphics[width=\linewidth]{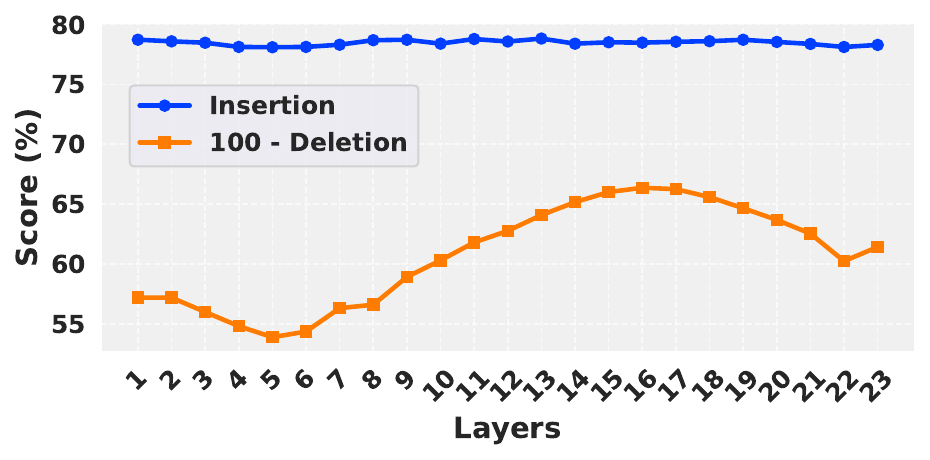}
    \caption{Layer-wise intervention sweep.}
    \label{fig:clip_layers}
  \end{subfigure}
  \vspace{-1mm}
  \caption{
  Layer analysis on ImageNet using CLIP-L/14.
  (a) Mean intra-object, inter-object, and object-to-background attention across layers.
  In early layers, intra-object attention increases and dominates, indicating strong object-level grouping.
  Around the middle layers, this gap reaches its peak.
  In later layers, intra-object attention decreases while object-to-background attention increases, reflecting growing extra-object contextualization and reduced spatial specificity.
  (b) Cumulative intervention depth sweep showing faithfulness metrics.
  Intermediate layers provide the best trade-off between object-level grouping and global contextual mixing.
  }
  \label{fig:clip_layers_combined}
  \vspace{-3mm}
\end{figure}

\paragraph{Choice of intervened layers.}
To analyze the effect of layer depth, we first examine attention patterns that govern information flow (Section~\ref{sec:layer_choice}). Fig.~\ref{fig:clip_attn} shows that 
$\mathbb{E}_{i}[\mathbb{E}_{j \in \mathcal{O}}[\alpha_{ij}^l] - \mathbb{E}_{j \notin \mathcal{O}}[\alpha_{ij}^l]]$
peaks in the middle layers.
To identify the optimal patching depth, we execute a cumulative layer sweep (Fig.~\ref{fig:clip_layers}), progressively patching all transformer blocks from the first up to a cutoff layer. Insertion AUC shows less sensitivity to the intervention depth than Deletion AUC. Overall, faithfulness peaks at intermediate layers: interventions on early blocks mainly inject low-level features with limited class evidence, while extending patching into later blocks reduces reliability, as late representations are dominated by global mixing and weaker spatial specificity. Based on this observation, which aligns with the analysis in Section~\ref{sec:layer_choice}, we fix $l_e = \lceil\tfrac{2}{3}L\rceil$ as the default setting for all architectures. This choice balances intra-object grouping and extra-object contextualization and performs near-optimally across models. Importantly, the performance (particularly Insertion AUC) remains stable across a broad range of intermediate layers, indicating that the method is not highly sensitive to the exact cutoff choice (results on other backbones are provided in Appendix~\ref{app:models_ablation}).

\paragraph{Input-based interventions.}
To better understand the role of representation-level interventions, we compare CAAP with two simple input-based alternatives: \emph{Input Insertion} and \emph{Input Deletion}. In Input Insertion, a selected source patch is inserted directly into a blank target image and the resulting class score is used as the attribution signal. In Input Deletion, the selected patch is removed from the input image while all other pixels remain unchanged. These procedures operate only at the input level and do not modify internal representations across layers. All methods use the default padding ($3 \times 3$ patch region). Tables~\ref{tab:faithfulness_simple_caap_comparison} and~\ref{tab:localization_simple_caap_comparison} report the results on ImageNet across four ViT backbones. Overall, CAAP consistently outperforms both input-based baselines. As shown in Table~\ref{tab:faithfulness_simple_caap_comparison}, CAAP achieves substantially lower Deletion AUC and higher Insertion AUC, leading to much larger Insertion$-$Deletion scores across all architectures. Localization metrics in Table~\ref{tab:localization_simple_caap_comparison} show the same pattern. CAAP yields higher AUPR$_1$, AUPR$_0$, and PG on most models, indicating better alignment with object regions. These results suggest that manipulating pixels alone does not reconstruct the internal evidence used by the model. In contrast, intervening on activations across layers preserves contextualized patch representations formed through self-attention, which leads to more faithful and spatially consistent attribution.

\input{tables/ablation_input}

\vspace{-1.0mm}
\section{Discussion and Limitations}
\vspace{-1.0mm}

CAAP reframes visual attribution in Vision Transformers as a causal intervention problem at the level of internal representations. Instead of relying on gradients, attention weights, or input-space perturbation, it measures the effect of replacing patch-specific activations within a controlled target context. This design directly evaluates how contextualized token embeddings influence the class score. By intervening at intermediate layers, CAAP captures class-relevant evidence after early object grouping while avoiding late-stage global mixing that weakens spatial specificity. In this sense, the method connects mechanistic activation patching with spatial attribution and provides a principled alternative to proxy-based explanations.

Empirically, CAAP improves faithfulness across multiple backbones, including CLIP, DINOv2, ViT, and DeiT3, by achieving lower Deletion AUC and higher Insertion AUC. Localization metrics such as AUPR$_1$, AUPR$_0$, and Pointing Game also improve, indicating that patch-level interventions better align with object regions rather than highlighting only a few highly discriminative patches. These results suggest that activation patching is not only a well-founded causal probe, but also an effective and scalable tool for identifying spatial evidence in ViTs.
However, this method also has limitations, as it is largely fitted to transformer architectures. Although activation patching preserves higher-order interactions between patches from the source context and avoids OOD behavior in its internals, the patched context does not follow the standard activation distribution. While this does not degrade model performance in practice, and ViTs appear robust to this kind of OOD activation, it should still be considered in mechanistic interpretability studies. Extending the approach beyond this setting, specifically to multimodal decoder architectures or generative models, remains an open direction.

\clearpage
{
\small
\bibliographystyle{unsrtnat}
\bibliography{main}
}


\clearpage

\newpage

\input{appendix}




\end{document}

%% file: tables/main_imagenet.tex
\begin{table*}[t]
\caption{Faithfulness quantitative results on ImageNet for ViT-L/16~\cite{vit}, CLIP-L/14~\cite{clip}, DINOv2-L/14~\cite{dino2}, and DeiT3-L/16~\cite{deit}. Deletion ($\downarrow$), Insertion ($\uparrow$), and Ins$-$Del ($\uparrow$) are reported.  Best is \textbf{bold}, second best is \underline{underlined} in each column. }
\vspace{-1.5mm}
\label{tab:faithfulness_comparison_results_main}
\centering
\small
\setlength{\tabcolsep}{4pt}
\renewcommand{\arraystretch}{1.15}
\resizebox{\textwidth}{!}{%
\begin{tabular}{lcccccccccccc}
\toprule
& \multicolumn{3}{c}{ViT-L/16~\cite{vit}}
& \multicolumn{3}{c}{CLIP-L/14~\cite{clip}}
& \multicolumn{3}{c}{DINOv2-L/14~\cite{dino2}}
& \multicolumn{3}{c}{DeiT3-L/16~\cite{deit}} \\
\cmidrule(lr){2-4}\cmidrule(lr){5-7}\cmidrule(lr){8-10}\cmidrule(lr){11-13}
Method
& Del$\downarrow$ & Ins$\uparrow$ & Ins$-$Del$\uparrow$
& Del$\downarrow$ & Ins$\uparrow$ & Ins$-$Del$\uparrow$
& Del$\downarrow$ & Ins$\uparrow$ & Ins$-$Del$\uparrow$
& Del$\downarrow$ & Ins$\uparrow$ & Ins$-$Del$\uparrow$ \\
\midrule

RISE~\cite{Petsiuk2018RISE}
& 55.53$_{\textcolor{gray}{\pm 2.36}}$ & \underline{71.83}$_{\textcolor{gray}{\pm 1.53}}$ & 16.30$_{\textcolor{gray}{\pm 2.09}}$
& 52.57$_{\textcolor{gray}{\pm 2.05}}$ & \underline{75.83}$_{\textcolor{gray}{\pm 1.12}}$ & 23.26$_{\textcolor{gray}{\pm 1.79}}$
& 59.59$_{\textcolor{gray}{\pm 2.20}}$ & \underline{78.03}$_{\textcolor{gray}{\pm 1.66}}$ & 18.45$_{\textcolor{gray}{\pm 1.48}}$
& 50.28$_{\textcolor{gray}{\pm 2.00}}$ & 63.05$_{\textcolor{gray}{\pm 1.39}}$ & 12.77$_{\textcolor{gray}{\pm 1.48}}$ \\

Grad-CAM~\cite{grad_cam}
& 48.99$_{\textcolor{gray}{\pm 1.22}}$ & 70.40$_{\textcolor{gray}{\pm 1.08}}$ & 21.40$_{\textcolor{gray}{\pm 1.17}}$
& 46.74$_{\textcolor{gray}{\pm 1.14}}$ & 73.19$_{\textcolor{gray}{\pm 0.85}}$ & 26.46$_{\textcolor{gray}{\pm 1.07}}$
& 49.36$_{\textcolor{gray}{\pm 1.20}}$ & 76.01$_{\textcolor{gray}{\pm 1.11}}$ & 26.65$_{\textcolor{gray}{\pm 1.00}}$
& \underline{31.81}$_{\textcolor{gray}{\pm 0.89}}$ & 62.18$_{\textcolor{gray}{\pm 0.93}}$ & 30.37$_{\textcolor{gray}{\pm 0.80}}$ \\

Rollout~\cite{attn_r}
& 67.69$_{\textcolor{gray}{\pm 1.06}}$ & 56.11$_{\textcolor{gray}{\pm 1.24}}$ & -11.58$_{\textcolor{gray}{\pm 1.03}}$
& 60.08$_{\textcolor{gray}{\pm 1.05}}$ & 64.90$_{\textcolor{gray}{\pm 1.00}}$ & 4.82$_{\textcolor{gray}{\pm 0.99}}$
& 63.10$_{\textcolor{gray}{\pm 1.16}}$ & 73.97$_{\textcolor{gray}{\pm 1.20}}$ & 10.86$_{\textcolor{gray}{\pm 0.81}}$
& 58.80$_{\textcolor{gray}{\pm 1.04}}$ & 53.27$_{\textcolor{gray}{\pm 1.04}}$ & -5.54$_{\textcolor{gray}{\pm 0.86}}$ \\

T-Attr~\cite{chefer2021transformer}
& 53.99$_{\textcolor{gray}{\pm 1.21}}$ & 68.75$_{\textcolor{gray}{\pm 1.08}}$ & 14.77$_{\textcolor{gray}{\pm 1.10}}$
& 50.20$_{\textcolor{gray}{\pm 1.14}}$ & 72.35$_{\textcolor{gray}{\pm 0.90}}$ & 22.15$_{\textcolor{gray}{\pm 1.02}}$
& 63.23$_{\textcolor{gray}{\pm 1.14}}$ & 73.82$_{\textcolor{gray}{\pm 1.20}}$ & 10.59$_{\textcolor{gray}{\pm 0.86}}$
& 37.36$_{\textcolor{gray}{\pm 1.00}}$ & 66.48$_{\textcolor{gray}{\pm 0.96}}$ & 29.12$_{\textcolor{gray}{\pm 0.92}}$ \\

ViT-CX~\cite{vit_cx}
& \underline{48.57}$_{\textcolor{gray}{\pm 1.27}}$ & 71.36$_{\textcolor{gray}{\pm 1.10}}$ & \underline{22.79}$_{\textcolor{gray}{\pm 1.25}}$
& 50.27$_{\textcolor{gray}{\pm 1.26}}$ & 69.65$_{\textcolor{gray}{\pm 0.96}}$ & 19.38$_{\textcolor{gray}{\pm 1.40}}$
& 49.50$_{\textcolor{gray}{\pm 1.26}}$ & 78.02$_{\textcolor{gray}{\pm 1.12}}$ & \underline{28.51}$_{\textcolor{gray}{\pm 1.17}}$
& 38.56$_{\textcolor{gray}{\pm 1.13}}$ & 58.99$_{\textcolor{gray}{\pm 0.96}}$ & 20.43$_{\textcolor{gray}{\pm 1.22}}$ \\

TAM~\cite{tam}
& 49.86$_{\textcolor{gray}{\pm 1.26}}$ & 69.52$_{\textcolor{gray}{\pm 1.11}}$ & 19.71$_{\textcolor{gray}{\pm 1.25}}$
& 46.25$_{\textcolor{gray}{\pm 1.17}}$ & 73.69$_{\textcolor{gray}{\pm 0.88}}$ & 27.44$_{\textcolor{gray}{\pm 1.12}}$
& 54.83$_{\textcolor{gray}{\pm 2.08}}$ & 77.52$_{\textcolor{gray}{\pm 1.59}}$ & 22.68$_{\textcolor{gray}{\pm 1.55}}$
& 35.88$_{\textcolor{gray}{\pm 1.04}}$ & \underline{67.78}$_{\textcolor{gray}{\pm 0.95}}$ & 31.90$_{\textcolor{gray}{\pm 0.94}}$ \\

IxG+~\cite{mehri-skipplus-cvpr24}
& 64.75$_{\textcolor{gray}{\pm 1.27}}$ & 59.21$_{\textcolor{gray}{\pm 1.19}}$ & -5.54$_{\textcolor{gray}{\pm 1.01}}$
& 61.87$_{\textcolor{gray}{\pm 1.19}}$ & 70.56$_{\textcolor{gray}{\pm 0.96}}$ & 8.69$_{\textcolor{gray}{\pm 0.92}}$
& 68.12$_{\textcolor{gray}{\pm 1.24}}$ & 73.45$_{\textcolor{gray}{\pm 1.34}}$ & 5.33$_{\textcolor{gray}{\pm 1.82}}$
& 52.16$_{\textcolor{gray}{\pm 1.07}}$ & 65.08$_{\textcolor{gray}{\pm 1.06}}$ & 12.92$_{\textcolor{gray}{\pm 0.98}}$ \\

Libra IxG+~\cite{Mehri_2025_CVPR}
& 51.92$_{\textcolor{gray}{\pm 1.30}}$ & 71.14$_{\textcolor{gray}{\pm 1.11}}$ & 19.22$_{\textcolor{gray}{\pm 1.21}}$
& \underline{44.27}$_{\textcolor{gray}{\pm 1.16}}$ & 75.24$_{\textcolor{gray}{\pm 0.83}}$ & \underline{30.97}$_{\textcolor{gray}{\pm 1.08}}$
& 60.80$_{\textcolor{gray}{\pm 1.22}}$ & 74.99$_{\textcolor{gray}{\pm 1.21}}$ & 14.18$_{\textcolor{gray}{\pm 1.71}}$
& 33.72$_{\textcolor{gray}{\pm 1.00}}$ & \textbf{69.49}$_{\textcolor{gray}{\pm 0.90}}$ & \textbf{35.77}$_{\textcolor{gray}{\pm 0.93}}$ \\

MDA~\cite{mda}
& 57.02$_{\textcolor{gray}{\pm 1.16}}$ & 62.12$_{\textcolor{gray}{\pm 1.21}}$ & 5.10$_{\textcolor{gray}{\pm 1.28}}$
& 46.56$_{\textcolor{gray}{\pm 1.14}}$ & 70.31$_{\textcolor{gray}{\pm 0.94}}$ & 23.74$_{\textcolor{gray}{\pm 1.21}}$
& \underline{49.20}$_{\textcolor{gray}{\pm 1.21}}$ & 76.95$_{\textcolor{gray}{\pm 1.16}}$ & 27.75$_{\textcolor{gray}{\pm 1.07}}$
& 47.49$_{\textcolor{gray}{\pm 1.03}}$ & 54.52$_{\textcolor{gray}{\pm 1.01}}$ & 7.02$_{\textcolor{gray}{\pm 1.06}}$ \\

MUTEX~\cite{mutex}
& 50.84$_{\textcolor{gray}{\pm 1.20}}$ & 68.60$_{\textcolor{gray}{\pm 1.08}}$ & 17.76$_{\textcolor{gray}{\pm 1.11}}$
& 48.75$_{\textcolor{gray}{\pm 1.16}}$ & 70.85$_{\textcolor{gray}{\pm 0.89}}$ & 22.11$_{\textcolor{gray}{\pm 1.10}}$
& 53.06$_{\textcolor{gray}{\pm 1.10}}$ & 76.32$_{\textcolor{gray}{\pm 1.15}}$ & 23.26$_{\textcolor{gray}{\pm 0.84}}$
& 39.14$_{\textcolor{gray}{\pm 1.05}}$ & 64.77$_{\textcolor{gray}{\pm 0.93}}$ & 25.62$_{\textcolor{gray}{\pm 1.04}}$ \\

\rowcolor{blue!10}
CAAP (Ours)
& \textbf{42.67}$_{\textcolor{gray}{\pm 1.16}}$ & \textbf{73.42}$_{\textcolor{gray}{\pm 1.07}}$ & \textbf{30.75}$_{\textcolor{gray}{\pm 1.18}}$
& \textbf{34.35}$_{\textcolor{gray}{\pm 0.99}}$ & \textbf{75.88}$_{\textcolor{gray}{\pm 0.85}}$ & \textbf{41.52}$_{\textcolor{gray}{\pm 1.02}}$
& \textbf{43.11}$_{\textcolor{gray}{\pm 1.14}}$ & \textbf{80.09}$_{\textcolor{gray}{\pm 1.08}}$ & \textbf{36.98}$_{\textcolor{gray}{\pm 1.00}}$
& \textbf{31.56}$_{\textcolor{gray}{\pm 0.91}}$ & 64.84$_{\textcolor{gray}{\pm 0.93}}$ & \underline{33.28}$_{\textcolor{gray}{\pm 0.85}}$ \\

\bottomrule
\end{tabular}
}
\end{table*}

\begin{table*}[t]
\caption{Localization quantitative results on ImageNet for ViT-L/16~\cite{vit}, CLIP-L/14~\cite{clip}, DINOv2-L/14~\cite{dino2}, and DeiT3-L/16~\cite{deit}. AUPR$_1$ ($\uparrow$), AUPR$_0$ ($\uparrow$) and PG ($\uparrow$) are reported.  Best is \textbf{bold}, second best is \underline{underlined} in each column.}
\vspace{-1.5mm}
\label{tab:localization_comparison_results_main}
\centering
\small
\setlength{\tabcolsep}{3.8pt}
\renewcommand{\arraystretch}{1.15}
\resizebox{\textwidth}{!}{%
\begin{tabular}{lcccccccccccc}
\toprule
& \multicolumn{3}{c}{ViT-L/16~\cite{vit}}
& \multicolumn{3}{c}{CLIP-L/14~\cite{clip}}
& \multicolumn{3}{c}{DINOv2-L/14~\cite{dino2}}
& \multicolumn{3}{c}{DeiT3-L/16~\cite{deit}} \\
\cmidrule(lr){2-4}\cmidrule(lr){5-7}\cmidrule(lr){8-10}\cmidrule(lr){11-13}
Method
& AUPR$_1$ $\uparrow$ & AUPR$_0$ $\uparrow$ & PG $\uparrow$
& AUPR$_1$ $\uparrow$ & AUPR$_0$ $\uparrow$ & PG $\uparrow$
& AUPR$_1$ $\uparrow$ & AUPR$_0$ $\uparrow$ & PG $\uparrow$
& AUPR$_1$ $\uparrow$ & AUPR$_0$ $\uparrow$ & PG $\uparrow$ \\
\midrule

RISE~\cite{Petsiuk2018RISE}
& 41.92$_{\textcolor{gray}{\pm 2.15}}$ & 71.04$_{\textcolor{gray}{\pm 2.24}}$ & 40.62$_{\textcolor{gray}{\pm 4.34}}$
& 43.12$_{\textcolor{gray}{\pm 2.13}}$ & 72.75$_{\textcolor{gray}{\pm 2.20}}$ & 39.84$_{\textcolor{gray}{\pm 4.33}}$
& 37.32$_{\textcolor{gray}{\pm 2.25}}$ & 69.62$_{\textcolor{gray}{\pm 2.19}}$ & 30.47$_{\textcolor{gray}{\pm 4.07}}$
& 40.28$_{\textcolor{gray}{\pm 2.19}}$ & 69.79$_{\textcolor{gray}{\pm 2.28}}$ & 35.94$_{\textcolor{gray}{\pm 4.24}}$ \\

Grad-CAM~\cite{grad_cam}
& 52.75$_{\textcolor{gray}{\pm 2.12}}$ & 78.83$_{\textcolor{gray}{\pm 2.10}}$ & 50.00$_{\textcolor{gray}{\pm 4.42}}$
& 53.34$_{\textcolor{gray}{\pm 2.05}}$ & 75.29$_{\textcolor{gray}{\pm 2.27}}$ & \underline{56.25}$_{\textcolor{gray}{\pm 4.38}}$
& \underline{50.10}$_{\textcolor{gray}{\pm 2.11}}$ & \underline{77.52}$_{\textcolor{gray}{\pm 2.24}}$ & \underline{43.75}$_{\textcolor{gray}{\pm 4.38}}$
& \underline{67.25}$_{\textcolor{gray}{\pm 1.93}}$ & \underline{85.64}$_{\textcolor{gray}{\pm 1.86}}$ & \underline{70.31}$_{\textcolor{gray}{\pm 4.04}}$ \\

Rollout~\cite{attn_r}
& 27.70$_{\textcolor{gray}{\pm 1.81}}$ & 61.81$_{\textcolor{gray}{\pm 2.58}}$ & 14.84$_{\textcolor{gray}{\pm 3.14}}$
& 37.75$_{\textcolor{gray}{\pm 2.11}}$ & 68.25$_{\textcolor{gray}{\pm 2.57}}$ & 32.81$_{\textcolor{gray}{\pm 4.15}}$
& 34.31$_{\textcolor{gray}{\pm 1.94}}$ & 66.23$_{\textcolor{gray}{\pm 2.40}}$ & 33.59$_{\textcolor{gray}{\pm 4.17}}$
& 29.40$_{\textcolor{gray}{\pm 1.90}}$ & 64.95$_{\textcolor{gray}{\pm 2.46}}$ & 3.12$_{\textcolor{gray}{\pm 1.54}}$ \\

T-Attr~\cite{chefer2021transformer}
& 38.74$_{\textcolor{gray}{\pm 1.91}}$ & 72.64$_{\textcolor{gray}{\pm 2.41}}$ & 20.31$_{\textcolor{gray}{\pm 3.56}}$
& 51.00$_{\textcolor{gray}{\pm 1.70}}$ & 75.27$_{\textcolor{gray}{\pm 2.35}}$ & 40.62$_{\textcolor{gray}{\pm 4.34}}$
& 40.23$_{\textcolor{gray}{\pm 2.32}}$ & 72.22$_{\textcolor{gray}{\pm 2.16}}$ & 36.72$_{\textcolor{gray}{\pm 4.26}}$
& 60.90$_{\textcolor{gray}{\pm 1.71}}$ & 81.13$_{\textcolor{gray}{\pm 1.99}}$ & 34.38$_{\textcolor{gray}{\pm 4.20}}$ \\

ViT-CX~\cite{vit_cx}
& \underline{56.35}$_{\textcolor{gray}{\pm 2.05}}$ & \underline{78.97}$_{\textcolor{gray}{\pm 2.22}}$ & \underline{53.12}$_{\textcolor{gray}{\pm 4.41}}$
& 49.89$_{\textcolor{gray}{\pm 2.42}}$ & 73.12$_{\textcolor{gray}{\pm 2.37}}$ & 50.78$_{\textcolor{gray}{\pm 4.42}}$
& 47.76$_{\textcolor{gray}{\pm 2.10}}$ & 72.29$_{\textcolor{gray}{\pm 2.50}}$ & \textbf{45.31}$_{\textcolor{gray}{\pm 4.40}}$
& 57.88$_{\textcolor{gray}{\pm 2.30}}$ & 77.61$_{\textcolor{gray}{\pm 2.48}}$ & 52.34$_{\textcolor{gray}{\pm 4.41}}$ \\

TAM~\cite{tam}
& 49.52$_{\textcolor{gray}{\pm 1.71}}$ & 74.59$_{\textcolor{gray}{\pm 2.48}}$ & 41.41$_{\textcolor{gray}{\pm 4.35}}$
& \underline{57.01}$_{\textcolor{gray}{\pm 1.94}}$ & 77.66$_{\textcolor{gray}{\pm 2.30}}$ & 54.69$_{\textcolor{gray}{\pm 4.40}}$
& 43.62$_{\textcolor{gray}{\pm 1.89}}$ & 72.91$_{\textcolor{gray}{\pm 2.45}}$ & 34.38$_{\textcolor{gray}{\pm 4.20}}$
& 62.34$_{\textcolor{gray}{\pm 1.86}}$ & 82.35$_{\textcolor{gray}{\pm 2.00}}$ & 57.03$_{\textcolor{gray}{\pm 4.38}}$ \\

IxG+~\cite{mehri-skipplus-cvpr24}
& 34.54$_{\textcolor{gray}{\pm 2.10}}$ & 67.30$_{\textcolor{gray}{\pm 2.09}}$ & 31.25$_{\textcolor{gray}{\pm 4.10}}$
& 43.18$_{\textcolor{gray}{\pm 1.92}}$ & 69.07$_{\textcolor{gray}{\pm 2.20}}$ & 49.22$_{\textcolor{gray}{\pm 4.42}}$
& 37.91$_{\textcolor{gray}{\pm 1.95}}$ & 66.24$_{\textcolor{gray}{\pm 2.11}}$ & 39.84$_{\textcolor{gray}{\pm 4.33}}$
& 38.10$_{\textcolor{gray}{\pm 1.91}}$ & 72.81$_{\textcolor{gray}{\pm 2.13}}$ & 10.94$_{\textcolor{gray}{\pm 2.76}}$ \\

Libra IxG+~\cite{Mehri_2025_CVPR}
& 49.04$_{\textcolor{gray}{\pm 1.90}}$ & 74.42$_{\textcolor{gray}{\pm 2.18}}$ & 50.78$_{\textcolor{gray}{\pm 4.42}}$
& 51.09$_{\textcolor{gray}{\pm 1.80}}$ & \underline{77.90}$_{\textcolor{gray}{\pm 2.15}}$ & 36.72$_{\textcolor{gray}{\pm 4.26}}$
& 38.73$_{\textcolor{gray}{\pm 2.03}}$ & 71.19$_{\textcolor{gray}{\pm 2.17}}$ & 35.94$_{\textcolor{gray}{\pm 4.24}}$
& 56.53$_{\textcolor{gray}{\pm 1.68}}$ & 80.70$_{\textcolor{gray}{\pm 2.09}}$ & 39.06$_{\textcolor{gray}{\pm 4.31}}$ \\

MDA~\cite{mda}
& 30.63$_{\textcolor{gray}{\pm 1.76}}$ & 64.73$_{\textcolor{gray}{\pm 2.69}}$ & 14.17$_{\textcolor{gray}{\pm 3.09}}$
& 43.82$_{\textcolor{gray}{\pm 2.11}}$ & 72.31$_{\textcolor{gray}{\pm 2.60}}$ & 32.28$_{\textcolor{gray}{\pm 4.15}}$
& 41.33$_{\textcolor{gray}{\pm 1.87}}$ & 69.30$_{\textcolor{gray}{\pm 2.70}}$ & 26.98$_{\textcolor{gray}{\pm 3.95}}$
& 34.03$_{\textcolor{gray}{\pm 1.94}}$ & 67.16$_{\textcolor{gray}{\pm 2.59}}$ & 13.28$_{\textcolor{gray}{\pm 3.00}}$ \\

MUTEX~\cite{mutex}
& 46.61$_{\textcolor{gray}{\pm 1.87}}$ & 73.48$_{\textcolor{gray}{\pm 2.51}}$ & 35.16$_{\textcolor{gray}{\pm 4.22}}$
& 46.81$_{\textcolor{gray}{\pm 2.02}}$ & 75.21$_{\textcolor{gray}{\pm 2.39}}$ & 39.84$_{\textcolor{gray}{\pm 4.33}}$
& 43.96$_{\textcolor{gray}{\pm 2.14}}$ & \textbf{77.74}$_{\textcolor{gray}{\pm 2.16}}$ & 35.16$_{\textcolor{gray}{\pm 4.22}}$
& 51.46$_{\textcolor{gray}{\pm 2.04}}$ & 77.87$_{\textcolor{gray}{\pm 2.28}}$ & 48.44$_{\textcolor{gray}{\pm 4.42}}$ \\

\rowcolor{blue!10}
CAAP (Ours)
& \textbf{63.40}$_{\textcolor{gray}{\pm 2.01}}$ & \textbf{83.40}$_{\textcolor{gray}{\pm 1.99}}$ & \textbf{73.44}$_{\textcolor{gray}{\pm 3.90}}$
& \textbf{66.42}$_{\textcolor{gray}{\pm 1.78}}$ & \textbf{84.86}$_{\textcolor{gray}{\pm 1.89}}$ & \textbf{74.22}$_{\textcolor{gray}{\pm 3.87}}$
& \textbf{52.76}$_{\textcolor{gray}{\pm 1.86}}$ & 76.68$_{\textcolor{gray}{\pm 2.17}}$ & 41.41$_{\textcolor{gray}{\pm 4.35}}$
& \textbf{71.21}$_{\textcolor{gray}{\pm 1.86}}$ & \textbf{86.93}$_{\textcolor{gray}{\pm 1.85}}$ & \textbf{82.81}$_{\textcolor{gray}{\pm 3.33}}$ \\

\bottomrule
\end{tabular}
}
\vspace{-4.0mm}
\end{table*}

%% file: tables/other_datasets.tex
\begin{table*}[t]
\caption{Faithfulness quantitative results on Oxford-IIIT-Pet~\cite{oxford_pet}, ImageNet-Hard~\cite{imagenet_hard}, and Food-101~\cite{food101} for ViT-B/16. Deletion ($\downarrow$), Insertion ($\uparrow$), and Ins$-$Del ($\uparrow$) are reported. Best is \textbf{bold} and second best is \underline{underlined} in each column.}
\label{tab:faithfulness_comparison_other_datasets}
\centering
\small
\setlength{\tabcolsep}{4pt}
\renewcommand{\arraystretch}{1.15}
\resizebox{\textwidth}{!}{%
\begin{tabular}{lccccccccc}
\toprule
& \multicolumn{3}{c}{Oxford-IIIT-Pet~\cite{oxford_pet}}
& \multicolumn{3}{c}{ImageNet-Hard\cite{imagenet_hard}}
& \multicolumn{3}{c}{Food-101~\cite{food101}} \\
\cmidrule(lr){2-4}\cmidrule(lr){5-7}\cmidrule(lr){8-10}
Method
& Del$\downarrow$ & Ins$\uparrow$ & Ins$-$Del$\uparrow$
& Del$\downarrow$ & Ins$\uparrow$ & Ins$-$Del$\uparrow$
& Del$\downarrow$ & Ins$\uparrow$ & Ins$-$Del$\uparrow$ \\
\midrule

RISE~\cite{Petsiuk2018RISE}
& 62.15$_{\textcolor{gray}{\pm 1.25}}$ & 85.08$_{\textcolor{gray}{\pm 0.57}}$ & 22.92$_{\textcolor{gray}{\pm 1.23}}$
& 11.69$_{\textcolor{gray}{\pm 0.78}}$ & \textbf{34.82}$_{\textcolor{gray}{\pm 1.29}}$ & \textbf{23.13}$_{\textcolor{gray}{\pm 1.15}}$
& 46.82$_{\textcolor{gray}{\pm 1.35}}$ & \underline{70.80}$_{\textcolor{gray}{\pm 0.82}}$ & 23.98$_{\textcolor{gray}{\pm 1.16}}$ \\

Grad-CAM~\cite{grad_cam}
& 61.95$_{\textcolor{gray}{\pm 1.08}}$ & 78.41$_{\textcolor{gray}{\pm 0.77}}$ & 16.46$_{\textcolor{gray}{\pm 1.23}}$
& 14.66$_{\textcolor{gray}{\pm 0.89}}$ & 26.04$_{\textcolor{gray}{\pm 1.15}}$ & 11.38$_{\textcolor{gray}{\pm 1.11}}$
& 51.84$_{\textcolor{gray}{\pm 1.26}}$ & 62.08$_{\textcolor{gray}{\pm 1.00}}$ & 10.25$_{\textcolor{gray}{\pm 1.22}}$ \\

Rollout~\cite{attn_r}
& 73.84$_{\textcolor{gray}{\pm 0.90}}$ & 72.15$_{\textcolor{gray}{\pm 0.83}}$ & -1.69$_{\textcolor{gray}{\pm 1.03}}$
& 22.61$_{\textcolor{gray}{\pm 1.07}}$ & 17.91$_{\textcolor{gray}{\pm 0.91}}$ & -4.70$_{\textcolor{gray}{\pm 0.97}}$
& 60.70$_{\textcolor{gray}{\pm 1.07}}$ & 54.77$_{\textcolor{gray}{\pm 1.02}}$ & -5.93$_{\textcolor{gray}{\pm 0.92}}$ \\

T-Attr~\cite{chefer2021transformer}
& 53.37$_{\textcolor{gray}{\pm 0.99}}$ & 84.32$_{\textcolor{gray}{\pm 0.61}}$ & 30.95$_{\textcolor{gray}{\pm 0.99}}$
& 12.86$_{\textcolor{gray}{\pm 0.79}}$ & 27.41$_{\textcolor{gray}{\pm 1.14}}$ & 14.55$_{\textcolor{gray}{\pm 1.01}}$
& 41.23$_{\textcolor{gray}{\pm 1.07}}$ & 65.52$_{\textcolor{gray}{\pm 1.00}}$ & 24.30$_{\textcolor{gray}{\pm 1.00}}$ \\

ViT-CX~\cite{vit_cx}
& \underline{47.79}$_{\textcolor{gray}{\pm 0.93}}$ & 85.26$_{\textcolor{gray}{\pm 0.71}}$ & \underline{37.47}$_{\textcolor{gray}{\pm 1.03}}$
& 14.14$_{\textcolor{gray}{\pm 0.78}}$ & 25.76$_{\textcolor{gray}{\pm 1.19}}$ & 11.62$_{\textcolor{gray}{\pm 1.12}}$
& \textbf{37.62}$_{\textcolor{gray}{\pm 0.96}}$ & 66.34$_{\textcolor{gray}{\pm 1.07}}$ & 28.71$_{\textcolor{gray}{\pm 1.11}}$ \\

TAM~\cite{tam}
& 58.79$_{\textcolor{gray}{\pm 1.08}}$ & 85.14$_{\textcolor{gray}{\pm 0.63}}$ & 26.34$_{\textcolor{gray}{\pm 1.09}}$
& 14.17$_{\textcolor{gray}{\pm 0.85}}$ & 26.27$_{\textcolor{gray}{\pm 1.15}}$ & 12.10$_{\textcolor{gray}{\pm 1.06}}$
& 45.59$_{\textcolor{gray}{\pm 1.17}}$ & 64.74$_{\textcolor{gray}{\pm 1.02}}$ & 19.15$_{\textcolor{gray}{\pm 1.08}}$ \\

IxG+~\cite{mehri-skipplus-cvpr24}
& 76.15$_{\textcolor{gray}{\pm 1.05}}$ & 79.06$_{\textcolor{gray}{\pm 0.79}}$ & 2.92$_{\textcolor{gray}{\pm 1.08}}$
& 15.77$_{\textcolor{gray}{\pm 0.98}}$ & 23.89$_{\textcolor{gray}{\pm 1.16}}$ & 8.12$_{\textcolor{gray}{\pm 1.14}}$
& 50.27$_{\textcolor{gray}{\pm 1.32}}$ & 63.05$_{\textcolor{gray}{\pm 1.01}}$ & 12.78$_{\textcolor{gray}{\pm 1.13}}$ \\

Libra IxG+~\cite{Mehri_2025_CVPR}
& 54.72$_{\textcolor{gray}{\pm 1.07}}$ & \underline{86.99}$_{\textcolor{gray}{\pm 0.59}}$ & 32.27$_{\textcolor{gray}{\pm 1.04}}$
& \underline{10.90}$_{\textcolor{gray}{\pm 0.73}}$ & \underline{31.64}$_{\textcolor{gray}{\pm 1.29}}$ & \underline{20.74}$_{\textcolor{gray}{\pm 1.21}}$
& \underline{38.10}$_{\textcolor{gray}{\pm 1.09}}$ & 69.47$_{\textcolor{gray}{\pm 1.00}}$ & \underline{31.37}$_{\textcolor{gray}{\pm 1.08}}$ \\

MDA~\cite{mda}
& 61.82$_{\textcolor{gray}{\pm 1.22}}$ & 75.18$_{\textcolor{gray}{\pm 0.98}}$ & 13.36$_{\textcolor{gray}{\pm 1.51}}$
& 19.69$_{\textcolor{gray}{\pm 0.98}}$ & 21.30$_{\textcolor{gray}{\pm 1.03}}$ & 1.61$_{\textcolor{gray}{\pm 1.03}}$
& 49.54$_{\textcolor{gray}{\pm 1.22}}$ & 60.99$_{\textcolor{gray}{\pm 1.18}}$ & 11.45$_{\textcolor{gray}{\pm 1.19}}$ \\

MUTEX~\cite{mutex}
& 56.12$_{\textcolor{gray}{\pm 1.03}}$ & 83.59$_{\textcolor{gray}{\pm 0.62}}$ & 27.47$_{\textcolor{gray}{\pm 1.05}}$
& 13.57$_{\textcolor{gray}{\pm 0.79}}$ & 27.48$_{\textcolor{gray}{\pm 1.16}}$ & 13.90$_{\textcolor{gray}{\pm 1.01}}$
& 40.08$_{\textcolor{gray}{\pm 1.04}}$ & 66.41$_{\textcolor{gray}{\pm 0.96}}$ & 26.32$_{\textcolor{gray}{\pm 0.95}}$ \\

\rowcolor{blue!10}
CAAP (Ours)
& \textbf{46.77}$_{\textcolor{gray}{\pm 0.95}}$
& \textbf{87.79}$_{\textcolor{gray}{\pm 0.62}}$
& \textbf{41.02}$_{\textcolor{gray}{\pm 0.95}}$
& \textbf{10.37}$_{\textcolor{gray}{\pm 0.66}}$
& 30.70$_{\textcolor{gray}{\pm 1.24}}$
& 20.33$_{\textcolor{gray}{\pm 1.09}}$
& 39.11$_{\textcolor{gray}{\pm 1.04}}$
& \textbf{71.50}$_{\textcolor{gray}{\pm 1.01}}$
& \textbf{32.39}$_{\textcolor{gray}{\pm 1.01}}$ \\

\bottomrule
\end{tabular}
}
\end{table*}

%% file: tables/ablation_input.tex
\begin{table*}[t]
\caption{Faithfulness quantitative results for Input Insertion, Input Deletion, and CAAP on ImageNet. Deletion ($\downarrow$), Insertion ($\uparrow$), and Ins$-$Del ($\uparrow$) are reported. Best is \textbf{bold}, second best is \underline{underlined} in each column.}
\vspace{-1.5mm}
\label{tab:faithfulness_simple_caap_comparison}
\centering
\small
\setlength{\tabcolsep}{4pt}
\renewcommand{\arraystretch}{1.15}
\resizebox{\textwidth}{!}{%
\begin{tabular}{lcccccccccccc}
\toprule
& \multicolumn{3}{c}{ViT-L/16~\cite{vit}}
& \multicolumn{3}{c}{CLIP-L/14~\cite{clip}}
& \multicolumn{3}{c}{DINOv2-L/14~\cite{dino2}}
& \multicolumn{3}{c}{DeiT3-L/16~\cite{deit}} \\
\cmidrule(lr){2-4}\cmidrule(lr){5-7}\cmidrule(lr){8-10}\cmidrule(lr){11-13}
Method
& Del$\downarrow$ & Ins$\uparrow$ & Ins$-$Del$\uparrow$
& Del$\downarrow$ & Ins$\uparrow$ & Ins$-$Del$\uparrow$
& Del$\downarrow$ & Ins$\uparrow$ & Ins$-$Del$\uparrow$
& Del$\downarrow$ & Ins$\uparrow$ & Ins$-$Del$\uparrow$ \\
\midrule

Input Insertion
& \underline{55.63}$_{\textcolor{gray}{\pm 1.24}}$ & 69.51$_{\textcolor{gray}{\pm 1.19}}$ & 10.63$_{\textcolor{gray}{\pm 1.01}}$
& \underline{43.86}$_{\textcolor{gray}{\pm 1.10}}$ & \underline{75.33}$_{\textcolor{gray}{\pm 0.87}}$ & \underline{14.93}$_{\textcolor{gray}{\pm 0.83}}$
& \underline{49.79}$_{\textcolor{gray}{\pm 1.20}}$ & 79.41$_{\textcolor{gray}{\pm 1.11}}$ & 19.59$_{\textcolor{gray}{\pm 0.79}}$
& \underline{33.33}$_{\textcolor{gray}{\pm 0.92}}$ & \underline{64.50}$_{\textcolor{gray}{\pm 0.97}}$ & \underline{22.16}$_{\textcolor{gray}{\pm 0.75}}$ \\

Input Deletion
& 58.89$_{\textcolor{gray}{\pm 1.37}}$ & \underline{69.75}$_{\textcolor{gray}{\pm 1.07}}$ & \underline{10.86}$_{\textcolor{gray}{\pm 1.20}}$
& 60.40$_{\textcolor{gray}{\pm 1.27}}$ & 72.38$_{\textcolor{gray}{\pm 0.89}}$ & 11.98$_{\textcolor{gray}{\pm 1.14}}$
& 59.82$_{\textcolor{gray}{\pm 1.33}}$ & \textbf{80.18}$_{\textcolor{gray}{\pm 1.05}}$ & \underline{20.36}$_{\textcolor{gray}{\pm 0.93}}$
& 42.34$_{\textcolor{gray}{\pm 0.90}}$ & 61.13$_{\textcolor{gray}{\pm 0.89}}$ & 18.79$_{\textcolor{gray}{\pm 0.88}}$ \\

\rowcolor{blue!10}
CAAP (Ours)
& \textbf{42.67}$_{\textcolor{gray}{\pm 1.16}}$ & \textbf{73.42}$_{\textcolor{gray}{\pm 1.07}}$ & \textbf{30.75}$_{\textcolor{gray}{\pm 1.18}}$
& \textbf{34.35}$_{\textcolor{gray}{\pm 0.99}}$ & \textbf{75.88}$_{\textcolor{gray}{\pm 0.85}}$ & \textbf{41.52}$_{\textcolor{gray}{\pm 1.02}}$
& \textbf{43.11}$_{\textcolor{gray}{\pm 1.14}}$ & \underline{80.09}$_{\textcolor{gray}{\pm 1.08}}$ & \textbf{36.98}$_{\textcolor{gray}{\pm 1.00}}$
& \textbf{31.56}$_{\textcolor{gray}{\pm 0.91}}$ & \textbf{64.84}$_{\textcolor{gray}{\pm 0.93}}$ & \textbf{33.28}$_{\textcolor{gray}{\pm 0.85}}$ \\

\bottomrule
\end{tabular}
}
\end{table*}

\begin{table*}[t]
\caption{Localization quantitative results for Input Insertion, Input Deletion, and CAAP on ImageNet. AUPR$_1$ ($\uparrow$), AUPR$_0$ ($\uparrow$) and PG ($\uparrow$) are reported. Best is \textbf{bold}, second best is \underline{underlined} in each column.}
\vspace{-1.5mm}
\label{tab:localization_simple_caap_comparison}
\centering
\small
\setlength{\tabcolsep}{3.8pt}
\renewcommand{\arraystretch}{1.15}
\resizebox{\textwidth}{!}{%
\begin{tabular}{lcccccccccccc}
\toprule
& \multicolumn{3}{c}{ViT-L/16~\cite{vit}}
& \multicolumn{3}{c}{CLIP-L/14~\cite{clip}}
& \multicolumn{3}{c}{DINOv2-L/14~\cite{dino2}}
& \multicolumn{3}{c}{DeiT3-L/16~\cite{deit}} \\
\cmidrule(lr){2-4}\cmidrule(lr){5-7}\cmidrule(lr){8-10}\cmidrule(lr){11-13}
Method
& AUPR$_1$ $\uparrow$ & AUPR$_0$ $\uparrow$ & PG $\uparrow$
& AUPR$_1$ $\uparrow$ & AUPR$_0$ $\uparrow$ & PG $\uparrow$
& AUPR$_1$ $\uparrow$ & AUPR$_0$ $\uparrow$ & PG $\uparrow$
& AUPR$_1$ $\uparrow$ & AUPR$_0$ $\uparrow$ & PG $\uparrow$ \\
\midrule

Input Insertion
& \underline{49.28}$_{\textcolor{gray}{\pm 1.99}}$ & \underline{71.10}$_{\textcolor{gray}{\pm 2.18}}$ & \underline{61.72}$_{\textcolor{gray}{\pm 4.30}}$
& \underline{55.32}$_{\textcolor{gray}{\pm 1.96}}$ & \underline{75.55}$_{\textcolor{gray}{\pm 2.09}}$ & \underline{59.38}$_{\textcolor{gray}{\pm 4.34}}$
& \underline{48.13}$_{\textcolor{gray}{\pm 2.02}}$ & \underline{72.81}$_{\textcolor{gray}{\pm 2.16}}$ & \textbf{52.34}$_{\textcolor{gray}{\pm 4.41}}$
& \underline{55.59}$_{\textcolor{gray}{\pm 1.97}}$ & 79.02$_{\textcolor{gray}{\pm 1.99}}$ & \underline{62.50}$_{\textcolor{gray}{\pm 4.28}}$ \\

Input Deletion
& 40.82$_{\textcolor{gray}{\pm 2.07}}$ & 67.53$_{\textcolor{gray}{\pm 2.23}}$ & 46.09$_{\textcolor{gray}{\pm 4.41}}$
& 40.44$_{\textcolor{gray}{\pm 2.10}}$ & 66.01$_{\textcolor{gray}{\pm 2.18}}$ & 58.59$_{\textcolor{gray}{\pm 4.35}}$
& 39.05$_{\textcolor{gray}{\pm 2.09}}$ & 67.45$_{\textcolor{gray}{\pm 2.27}}$ & \underline{45.31}$_{\textcolor{gray}{\pm 4.40}}$
& 47.51$_{\textcolor{gray}{\pm 2.50}}$ & \underline{80.74}$_{\textcolor{gray}{\pm 1.69}}$ & 57.81$_{\textcolor{gray}{\pm 4.37}}$ \\

\rowcolor{blue!10}
CAAP (Ours)
& \textbf{63.40}$_{\textcolor{gray}{\pm 2.01}}$ & \textbf{83.40}$_{\textcolor{gray}{\pm 1.99}}$ & \textbf{73.44}$_{\textcolor{gray}{\pm 3.90}}$
& \textbf{66.42}$_{\textcolor{gray}{\pm 1.78}}$ & \textbf{84.86}$_{\textcolor{gray}{\pm 1.89}}$ & \textbf{74.22}$_{\textcolor{gray}{\pm 3.87}}$
& \textbf{52.76}$_{\textcolor{gray}{\pm 1.86}}$ & \textbf{76.68}$_{\textcolor{gray}{\pm 2.17}}$ & 41.41$_{\textcolor{gray}{\pm 4.35}}$
& \textbf{71.21}$_{\textcolor{gray}{\pm 1.86}}$ & \textbf{86.93}$_{\textcolor{gray}{\pm 1.85}}$ & \textbf{82.81}$_{\textcolor{gray}{\pm 3.33}}$ \\

\bottomrule
\end{tabular}
}
\end{table*}

%% file: appendix.tex
\appendix

\section{Related Work}
\vspace{-1.0mm}

We review prior work on explaining Vision Transformers (ViTs) through three commonly used sources of explanatory signal: gradients, attention, and input-level perturbations.

\vspace{-1.0mm}
\paragraph{Gradient-based attribution.}
Gradient-based approaches estimate relevance using local sensitivity of the class score with respect to inputs or intermediate features.
Representative methods include activation-difference propagation and path-integrated gradients, such as DeepLIFT and Integrated Gradients~\cite{shrikumar2016not,sundararajan2017axiomatic}, with noise-reduction variants such as SmoothGrad~\cite{smilkov2017smoothgrad}.
CAM-style localization derives class-discriminative maps from gradient-weighted or score-weighted features, including Grad-CAM, Grad-CAM++, and Score-CAM~\cite{grad_cam,chattopadhyay2017gradcampp,wang2020scorecam}.
For ViTs, LibraGrad shows that nonlinear layers can make gradient-based attribution unfaithful, and proposes post-hoc gradient corrections that approximately mitigate this issue for common nonlinear operations in Transformers~\cite{Mehri_2025_CVPR}.
Layer-aggregation variants such as SkipPLUS further strengthen explanations by reducing the effect of noisy early layers when aggregating gradient-based signals~\cite{mehri-skipplus-cvpr24}.
In contrast to gradient- and relevance-propagation methods, which trace local sensitivity or backward relevance through the computation graph, CAAP directly intervenes on the forward activations of contextualized patch tokens.

\vspace{-1.0mm}
\paragraph{Attention-based attribution.}
Because ViTs rely on self-attention, many explanations derive saliency from attention weights and their compositions.
Attention Flow and related rollout-style approaches compose attention across layers to propagate token influence~\cite{attn_r}.
However, attention weights alone do not necessarily align with causal importance~\cite{Jain2019Attention,kobayashi2020attention}, motivating approaches that incorporate class information or additional model signals.
Transformer Attribution propagates relevance through attention blocks using gradient-informed updates~\cite{chefer2021transformer}, and AttCAT combines features, gradients, and attention to produce token-level explanations~\cite{qiang2022attcat}.
Transition Attention Maps (TAM) models information flow through attention as a Markov process~\cite{tam}, while Grad-SAM leverages gradient self-attention maps to highlight influential inputs within attention units~\cite{barkan2021gradsam}.
Graph-structured attention explanations also exist; MUTEX constructs a multiplex network over attention patterns across layers to generate patch heatmaps~\cite{mutex}.
AtMan manipulates attention mechanisms in Transformers to produce relevance maps without relying on backpropagation, using a memory-efficient token-search procedure based on neighborhoods in embedding space~\cite{deb2023atman}.
Unlike attention-based methods, which use token-mixing patterns or attention manipulations as explanatory signals, CAAP measures the effect of patch-associated representations on the model output through an explicit activation intervention.

\vspace{-1.0mm}
\paragraph{Perturbation-based attribution and masking in ViTs.}
Perturbation approaches estimate importance by modifying inputs, masks, or early token representations and observing output changes.
RISE aggregates responses over randomized input masks~\cite{Petsiuk2018RISE}, and optimization-based masking seeks minimally sufficient or necessary regions~\cite{fong2017meaningful,fong2019extremal}.
For ViTs, ViT-CX builds causal explanations from patch embeddings and their causal impact on the model output~\cite{vit_cx}, and Transformer Input Sampling (TIS) perturbs the model by sampling input tokens, using the variable-token property of Transformers to avoid introducing replacement artifacts~\cite{Englebert_2023_ICCV}.
Metric-Driven Attributions (MDA) directly optimizes attribution maps to score well under faithfulness metrics such as insertion and deletion~\cite{mda}.
These methods provide important perturbation-based or metric-driven views of ViT attribution, but their interventions are applied to external inputs, early token representations, or the attribution map itself. In contrast, CAAP intervenes on contextualized residual activations inside the ViT.

\vspace{-1.0mm}
\paragraph{Positioning of CAAP.}
CAAP can be viewed as an interventional attribution approach that assigns relevance by replacing internal activations and measuring the resulting target-class score.
It complements masking- and sampling-based perturbation methods by moving the intervention from the input or token level to the representation level, and it complements attention-derived explanations by treating attention as part of the computation that forms the intervened states rather than as the explanatory signal itself.
It is also distinct from metric-driven approaches that optimize attribution maps to fit evaluation criteria: CAAP derives each score from an explicit intervention within the classifier itself, without learning an external explainer, mask generator, or metric-specific surrogate.
The key distinction is that CAAP directly tests the predictive effect of intermediate residual representations associated with image patches.
Methods such as RISE~\cite{Petsiuk2018RISE} estimate importance by aggregating model responses over input-space masks, while ViT-CX~\cite{vit_cx} constructs explanations from patch embeddings and output-level causal scores.
These methods therefore do not directly measure how the contextualized residual activations formed inside the transformer affect the final prediction.
CAAP addresses this gap by inserting source-image residual activations into a controlled target computation and asking whether those internal patch representations support the target-class prediction.
Thus, CAAP shifts the intervention site for ViT attribution from external or early model variables to internal representations that are closer to the model's final decision computation.

\section{Broader Experiments}
\label{app:experiments}
\begin{figure*}[h]
  \centering
  \begin{subfigure}[t]{0.49\textwidth}
    \centering
    \includegraphics[width=\linewidth]{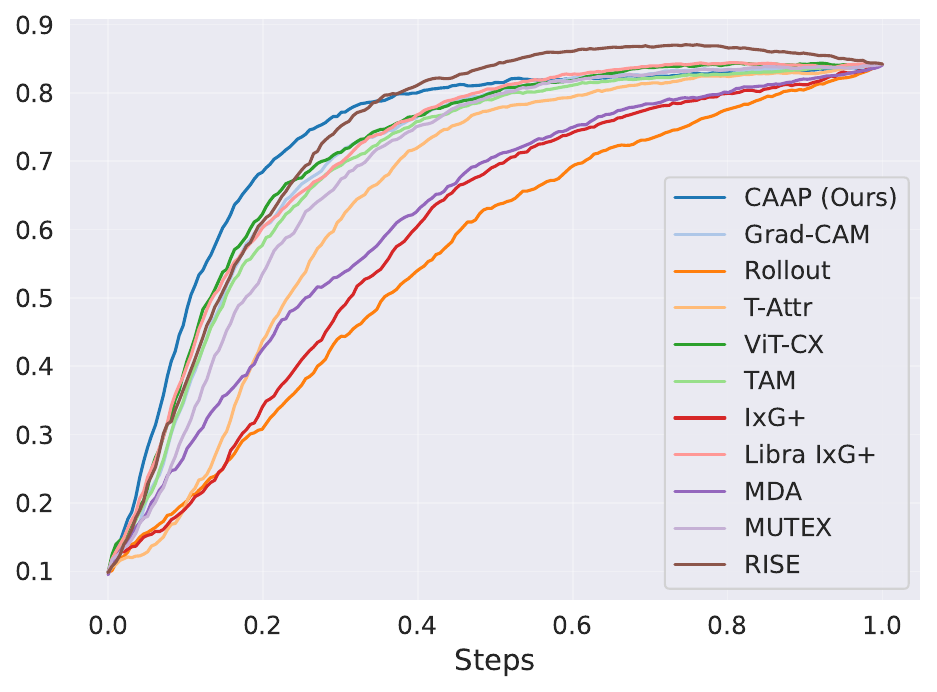}
    \caption{ViT-L/16}
    \label{fig:ins_vit}
  \end{subfigure}\hfill
  \begin{subfigure}[t]{0.49\textwidth}
    \centering
    \includegraphics[width=\linewidth]{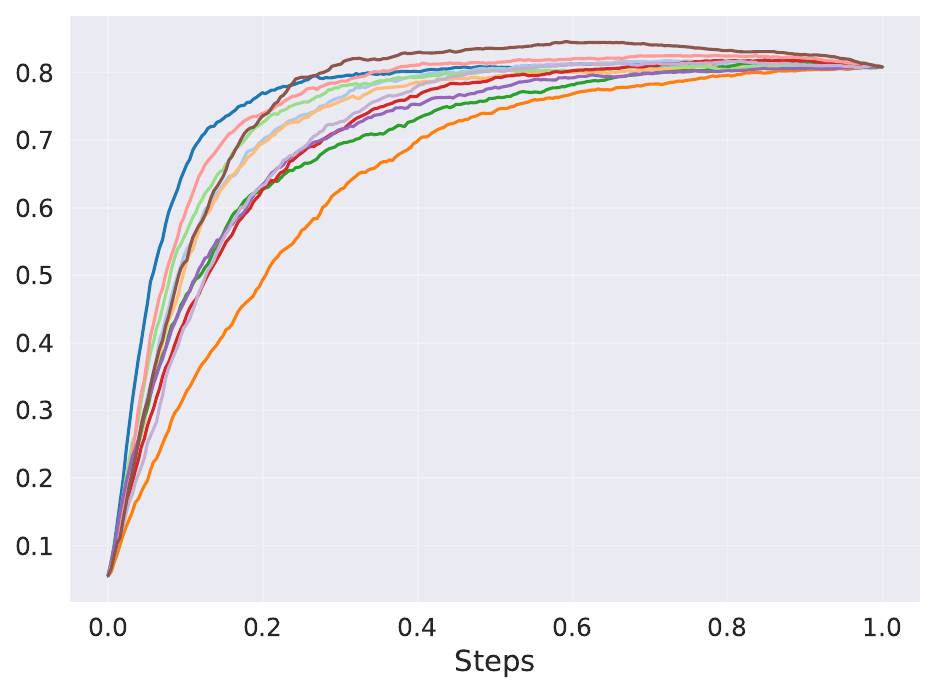}
    \caption{CLIP-L/14}
    \label{fig:ins_clip}
  \end{subfigure}\hfill
  \begin{subfigure}[t]{0.49\textwidth}
    \centering
    \includegraphics[width=\linewidth]{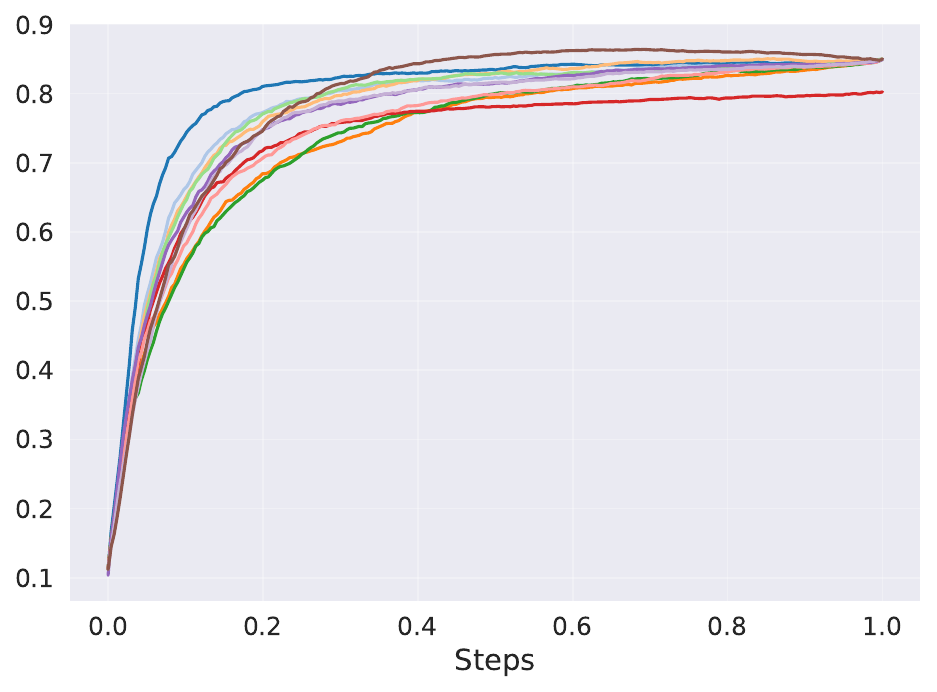}
    \caption{DINOv2-L/14}
    \label{fig:ins_dino}
  \end{subfigure}\hfill
  \begin{subfigure}[t]{0.49\textwidth}
    \centering
    \includegraphics[width=\linewidth]{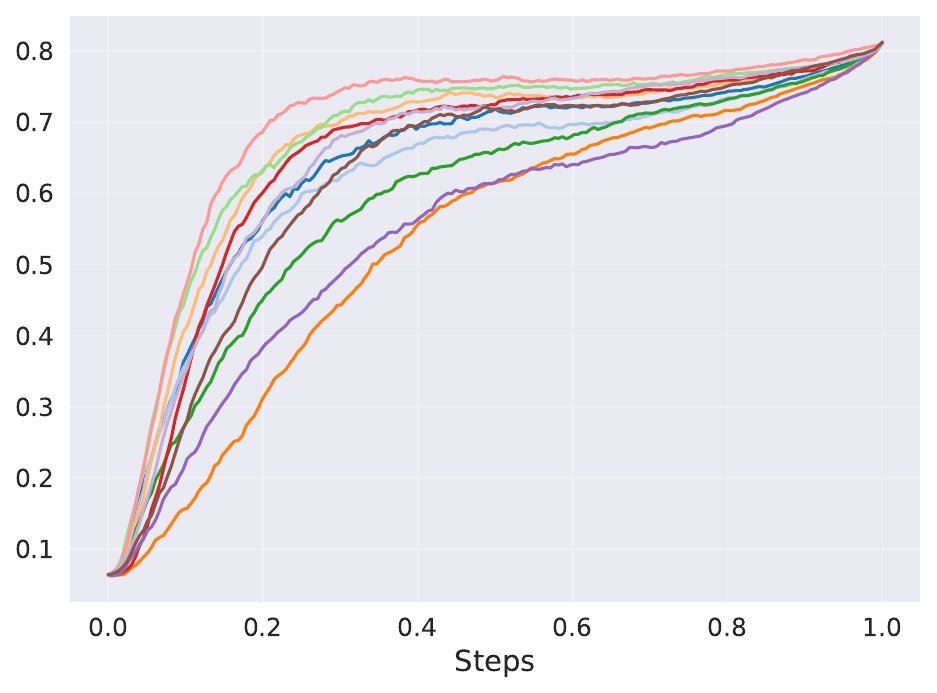}
    \caption{DeiT3-L/16}
    \label{fig:ins_deit}
  \end{subfigure}

  \caption{Representative insertion curves on ImageNet for ViT-L/16, CLIP-L/14, DINOv2-L/14, and DeiT3-L/16. Higher curves indicate that the attribution method more effectively identifies regions that positively support the model prediction.}
  \label{fig:insertion_curves_4}
\end{figure*}

\begin{figure*}[h]
  \centering
  \begin{subfigure}[t]{0.49\textwidth}
    \centering
    \includegraphics[width=\linewidth]{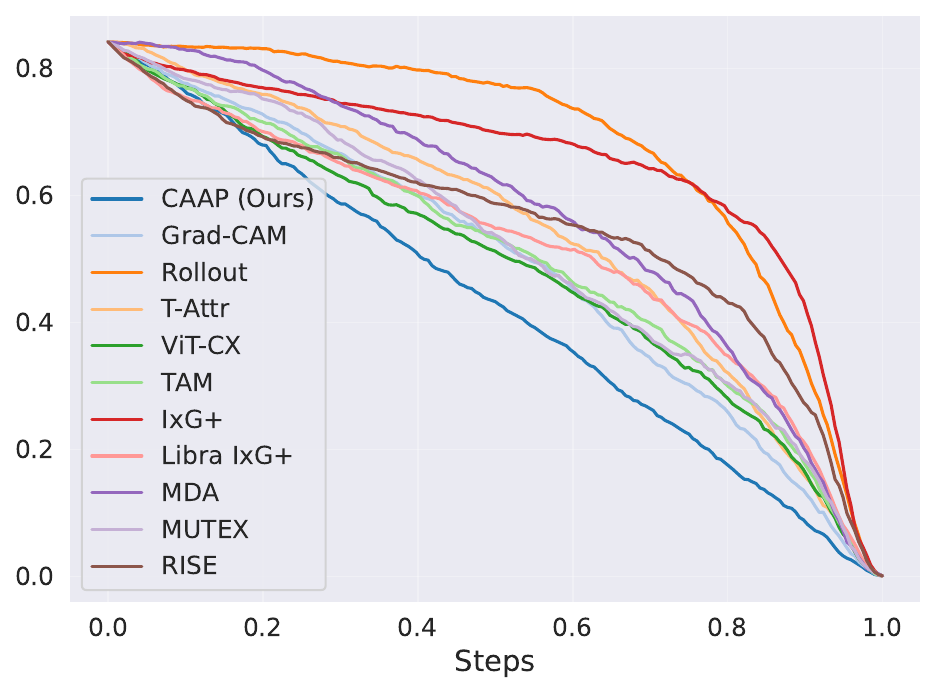}
    \caption{ViT-L/16}
    \label{fig:del_vit}
  \end{subfigure}\hfill
  \begin{subfigure}[t]{0.49\textwidth}
    \centering
    \includegraphics[width=\linewidth]{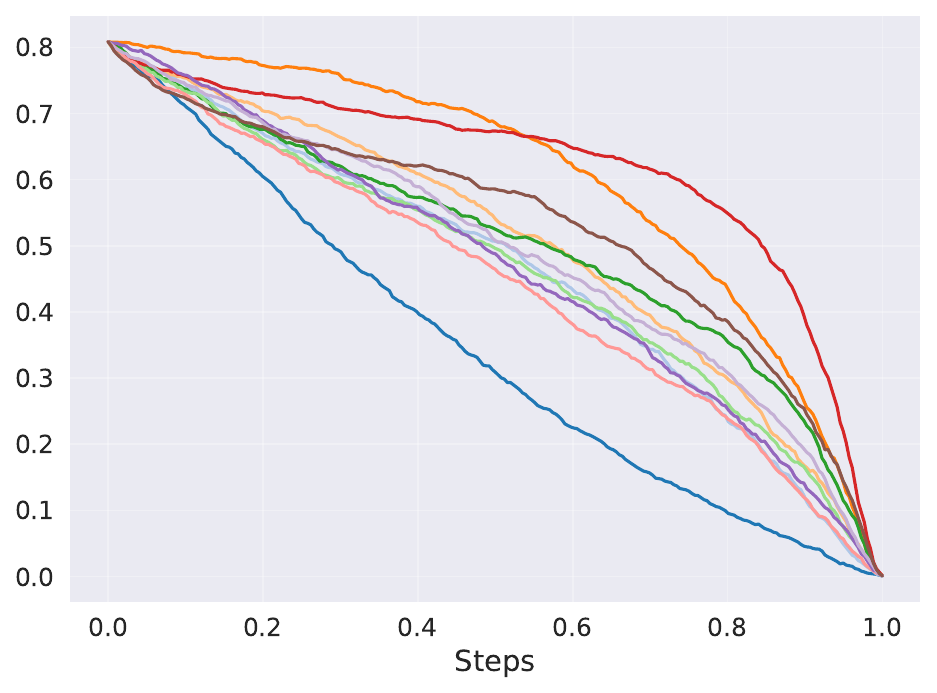}
    \caption{CLIP-L/14}
    \label{fig:del_clip}
  \end{subfigure}\hfill
  \begin{subfigure}[t]{0.49\textwidth}
    \centering
    \includegraphics[width=\linewidth]{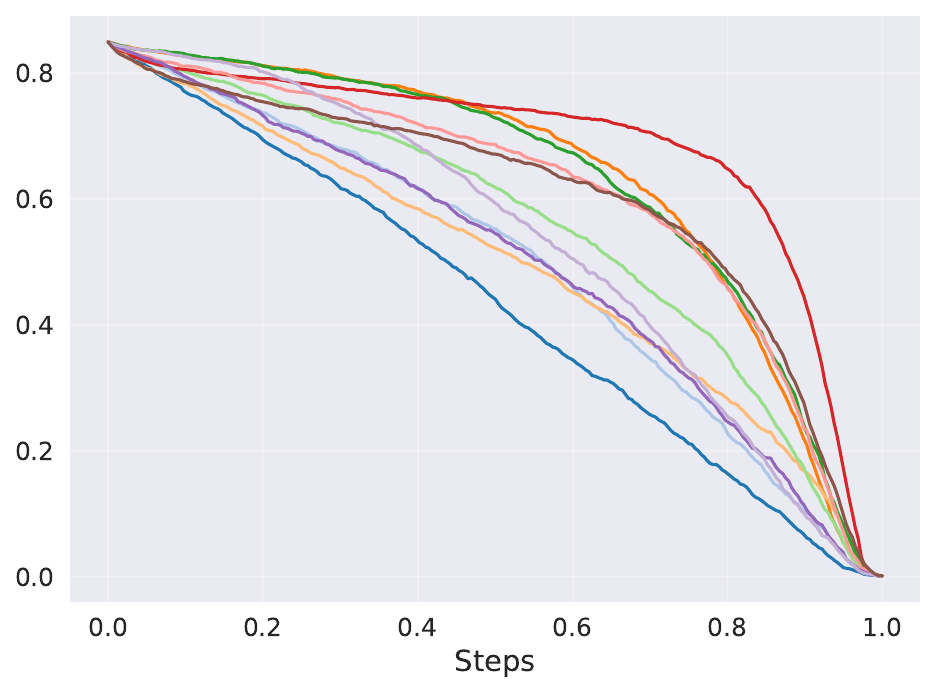}
    \caption{DINOv2-L/14}
    \label{fig:del_dino}
  \end{subfigure}\hfill
  \begin{subfigure}[t]{0.49\textwidth}
    \centering
    \includegraphics[width=\linewidth]{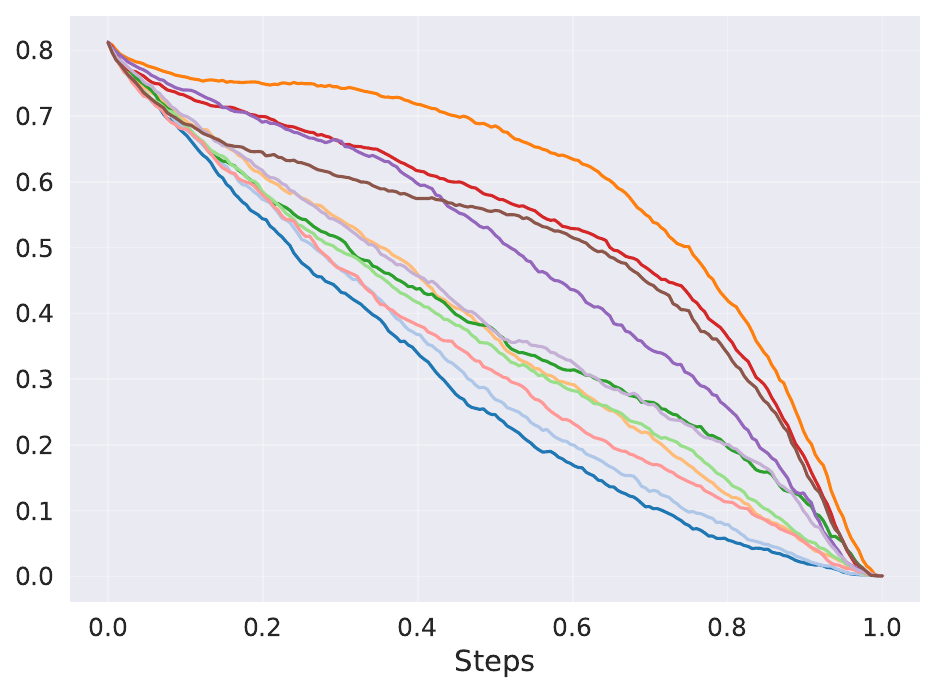}
    \caption{DeiT3-L/16}
    \label{fig:del_deit}
  \end{subfigure}

  \caption{Representative deletion curves on ImageNet for ViT-L/16, CLIP-L/14, DINOv2-L/14, and DeiT3-L/16. Lower curves indicate that the attribution method more accurately identifies regions whose removal causes a stronger drop in model confidence.}
  \label{fig:deletion_curves_4}
\end{figure*}

In this section, we present additional experimental results that further expand the empirical evaluation in the main paper. In particular, we extend the study to other model scales, including base and, where available, small variants, as well as the largest variants of the considered architectures (huge/giant models). Additional quantitative results are reported in Section~\ref{app:more_quantitative}, and further qualitative examples are provided in Section~\ref{app:more_qualitative}. We also provide representative insertion and deletion curves for the large variants of the backbones in Figs~\ref{fig:insertion_curves_4} and~\ref{fig:deletion_curves_4}.

\subsection{Additional Quantitative Results}
\label{app:more_quantitative}
\begin{figure}[t]
    \centering
    \includegraphics[width=1\linewidth]{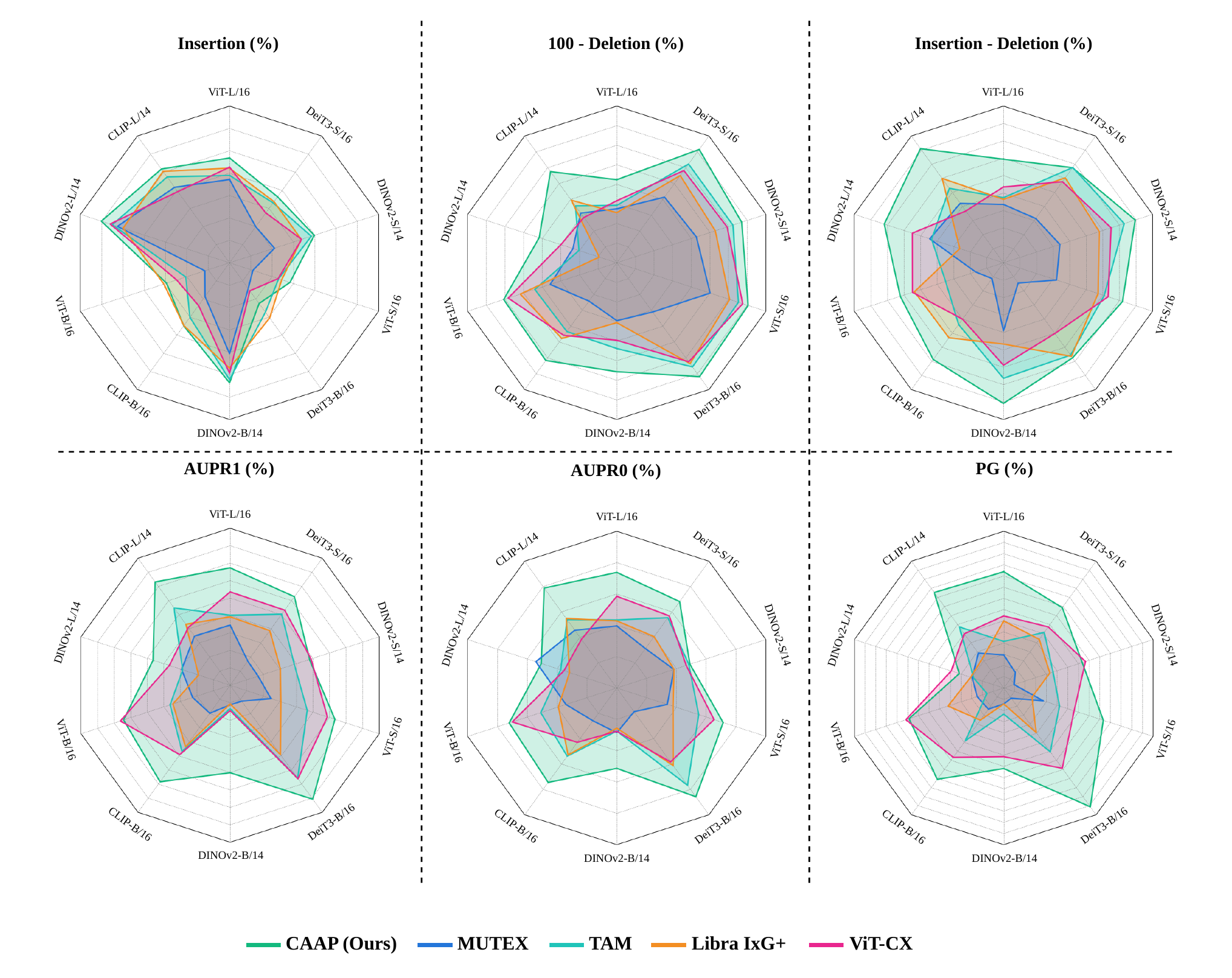}
    \caption{
Radar-plot comparison of attribution performance over 10 representative ViT backbones. CAAP is compared against the four strongest baselines: MUTEX, TAM, Libra IxG+, and ViT-CX, across faithfulness and localization metrics. Higher values indicate better performance. CAAP shows the most consistently strong performance across backbones, supporting its effectiveness as a faithful and spatially precise  method.
}
    \label{fig:radar-comparison}
\end{figure}

We present broader quantitative experiments on a wider range of architecture scales. In particular, we report results for base variants in Tables~\ref{tab:faithfulness_comparison_results_base} and~\ref{tab:localization_comparison_results_base}. We also include, where available, results for small variants in Tables~\ref{tab:faithfulness_comparison_results_small} and~\ref{tab:localization_comparison_results_small}, as well as for the largest available variants in Tables~\ref{tab:faithfulness_comparison_results_largest} and~\ref{tab:localization_comparison_results_largest}. Across all of these settings, the trends observed in the main experiments remain consistent. Specifically, our method continues to achieve lower Deletion scores together with higher Insertion and Insertion$-$Deletion values than the competing baselines, indicating stronger faithfulness of the produced attributions. In addition, CAAP consistently obtains the best or among the best AUPR$_1$, AUPR$_0$, and PG results, demonstrating improved localization quality and a stronger ability to focus attribution mass on foreground object regions while suppressing background responses. This behavior is also reflected in Figure~\ref{fig:radar-comparison}, which summarizes the comparison over 10 representative ViT backbones and shows that CAAP maintains the strongest overall profile across both faithfulness and localization criteria. These results further support the robustness of our approach across different architectures, training paradigms, and model scales, and suggest that our interventional patch-level attribution method generalizes well beyond the backbone configurations considered in the main paper.

\input{tables/base_imagenet}

\input{tables/small_imagenet}

\input{tables/huge_imagenet}

\subsection{Infidelity and Compactness Metrics}
\label{app:compactness}

In addition to our primary faithfulness and localization
metrics, we evaluate attribution maps using
Infidelity~\cite{yeh2019infidelity} and compactness metrics.
Infidelity measures how well the attribution-predicted
change agrees with the actual change in the model's
target-class score under input perturbations.
Given an attribution map $A(x)$ for image $x$,
target class $y$, and a perturbation $I$, it is defined as
\begin{equation}
\mathrm{INFD}(A,x) =
\mathbb{E}_{I}\left[
\left(I^\top A(x) -
\left(f(x)_y - f(x-I)_y\right)\right)^2
\right].
\end{equation}

We also assess the spatial concentration of attributions
using compactness metrics for single-object classification.
Given nonnegative patch scores $a_i$ and normalized scores
$p_i=a_i/\sum_j a_j$, we report normalized Entropy,
$$
H(p)=-\frac{1}{\log N}\sum_{i=1}^N p_i\log p_i,
$$
and the Gini index,
$$
G=\frac{\sum_{i=1}^N (2i-N-1)a_{(i)}}{N\sum_{i=1}^N a_i},
$$
where $a_{(i)}$ are scores sorted in increasing order.

Lower Infidelity indicates better agreement between
attribution scores and the changes in the model's prediction.
Lower Entropy and higher Gini indicate more concentrated
attribution maps.

Table~\ref{tab:compactness_comparison_results_main}
reports Infidelity, Entropy, and Gini across four ViT
backbones. CAAP achieves the lowest Infidelity on
ViT-L/16, CLIP-L/14, and DINOv2-L/14 and generally
produces compact attribution maps, while MUTEX obtains
the lowest Infidelity on DeiT3-L/16.

\input{tables/compactness_scores}

\subsection{Attribution with Non-CLS Readouts}
\label{app:cr_readouts}

To evaluate the applicability of CAAP beyond CLS-token-based
architectures, we conduct additional experiments on BEiT and
SigLIP, which employ mean pooling and attention pooling,
respectively. For these models, we apply activation patching
to the internal patch representations and use each model's
original readout mechanism to compute the attribution scores.
Table~\ref{tab:cr_readouts} presents the faithfulness and
localization results. CAAP achieves consistently strong
performance across both architectures, demonstrating its
effectiveness with different readout mechanisms.

\input{tables/non_cls_readouts}

\subsection{Additional Qualitative Results}
\label{app:more_qualitative}
\begin{figure}[t]
  \centering
  \includegraphics[width=0.99\linewidth]{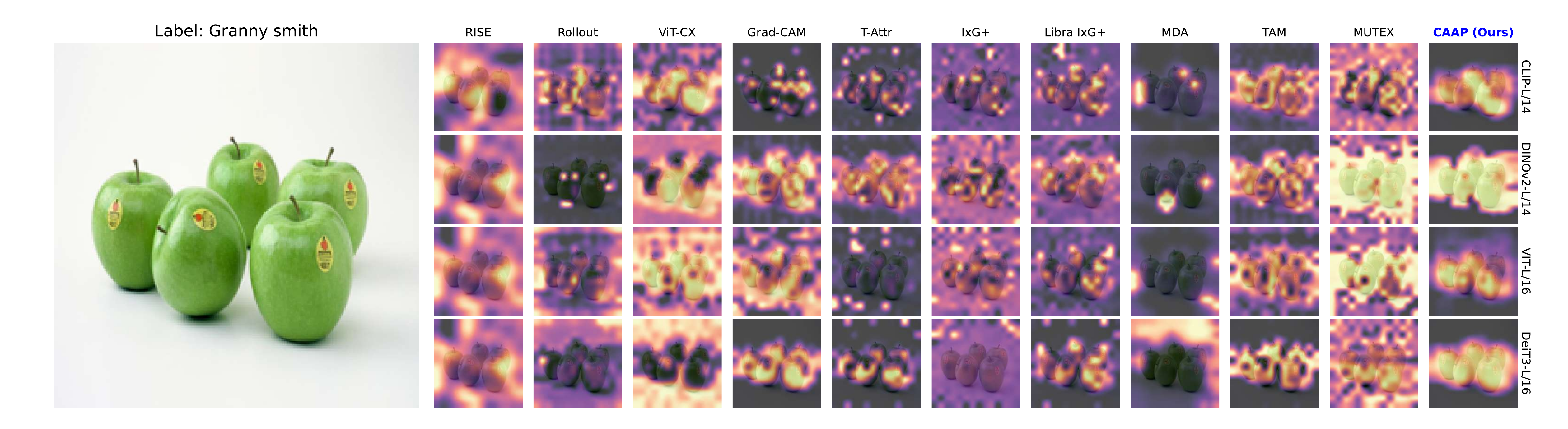}
  \vspace{-1mm}
  \caption{
    Visualization of single-object attribution maps produced by different methods across various models. The target object is an \texttt{Granny smith}.
    }
  \label{fig:vis_11}
  \vspace{-1mm}
\end{figure}

\begin{figure}[t]
  \centering
  \includegraphics[width=0.99\linewidth]{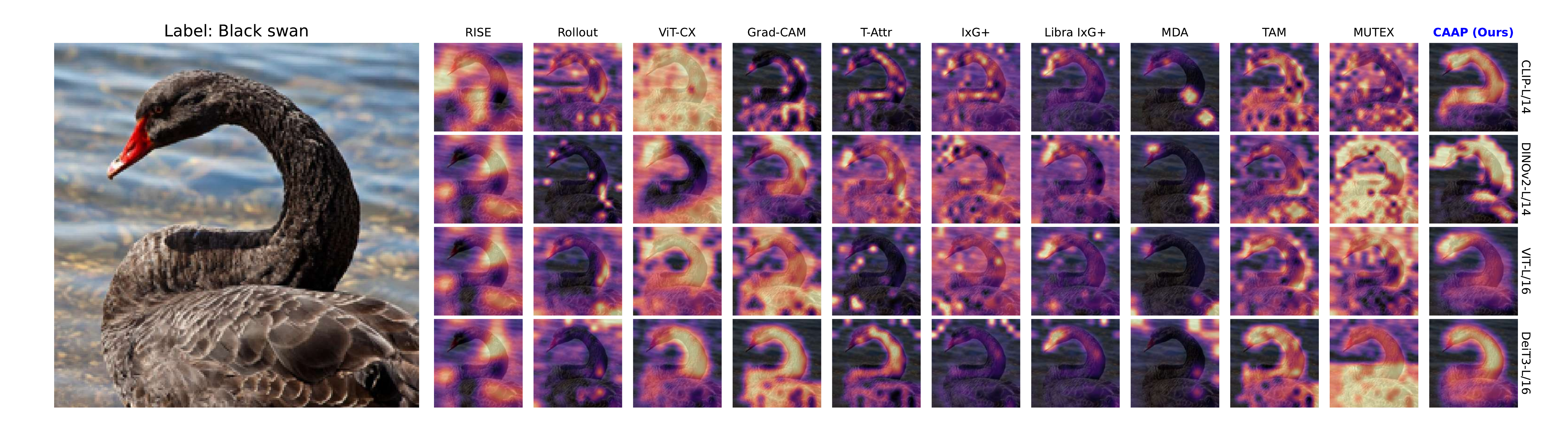}
  \vspace{-1mm}
  \caption{
    Visualization of single-object attribution maps produced by different methods across various models. The target object is a \texttt{Black swan}.
    }
  \label{fig:vis_12}
  \vspace{-1mm}
\end{figure}

\begin{figure}[t]
  \centering
  \includegraphics[width=0.99\linewidth]{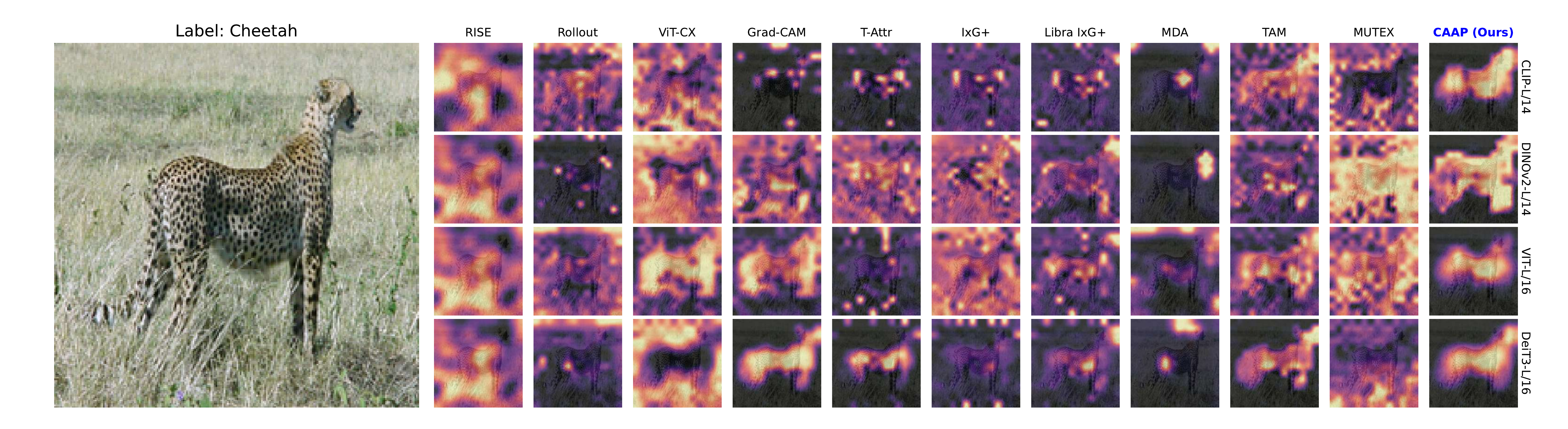}
  \vspace{-1mm}
  \caption{
    Visualization of single-object attribution maps produced by different methods across various models. The target object is a \texttt{Cheetah}.
    }
  \label{fig:vis_29}
  \vspace{-1mm}
\end{figure}

\begin{figure}[t]
  \centering
  \includegraphics[width=0.99\linewidth]{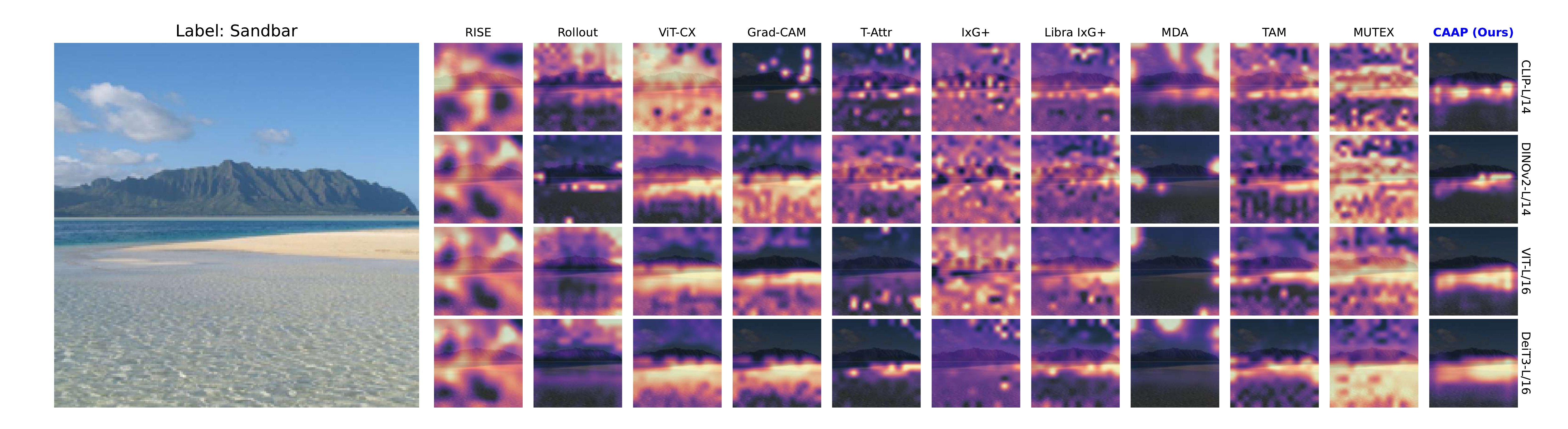}
  \vspace{-1mm}
  \caption{
    Visualization of single-object attribution maps produced by different methods across various models. The target object is a \texttt{Sandbar}.
    }
  \label{fig:vis_38}
  \vspace{-1mm}
\end{figure}

\begin{figure}[t]
  \centering
  \includegraphics[width=0.99\linewidth]{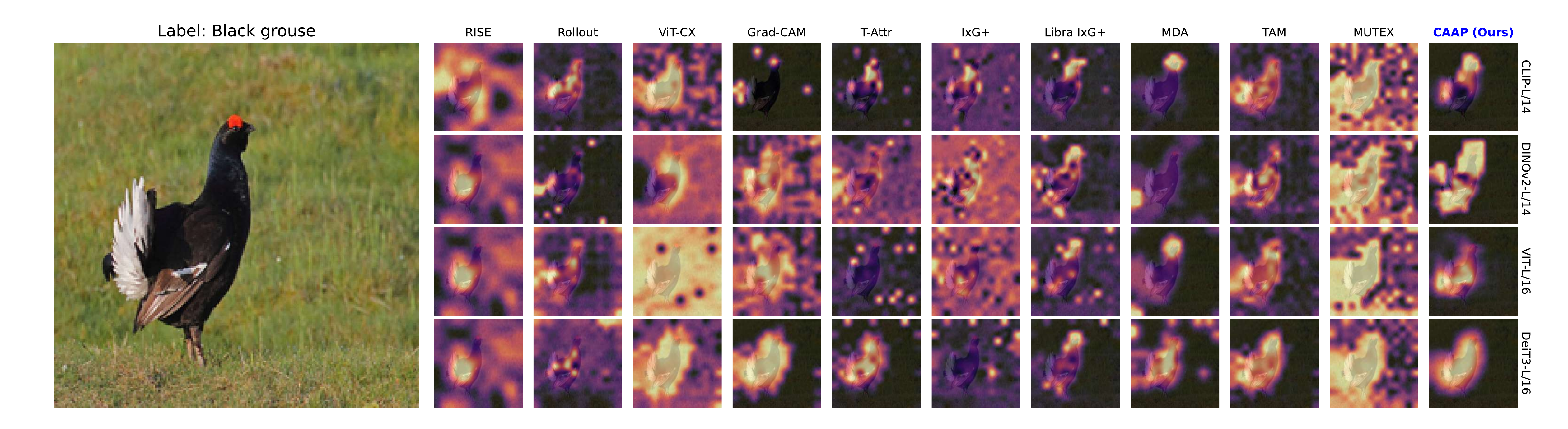}
  \vspace{-1mm}
  \caption{
    Visualization of single-object attribution maps produced by different methods across various models. The target object is a \texttt{Black grouse}.
    }
  \label{fig:vis_51}
  \vspace{-1mm}
\end{figure}

\begin{figure}[t]
  \centering
  \includegraphics[width=0.99\linewidth]{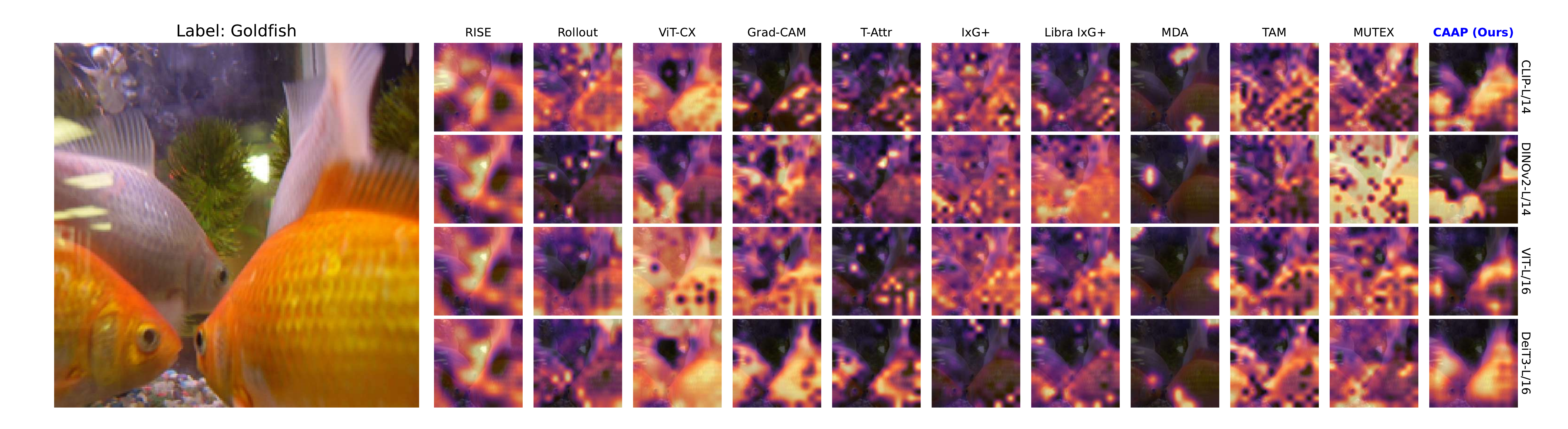}
  \vspace{-1mm}
  \caption{
    Visualization of single-object attribution maps produced by different methods across various models. The target object is a \texttt{Goldfish}.
    }
  \label{fig:vis_77}
  \vspace{-1mm}
\end{figure}

\begin{figure}[t]
  \centering
  \includegraphics[width=0.99\linewidth]{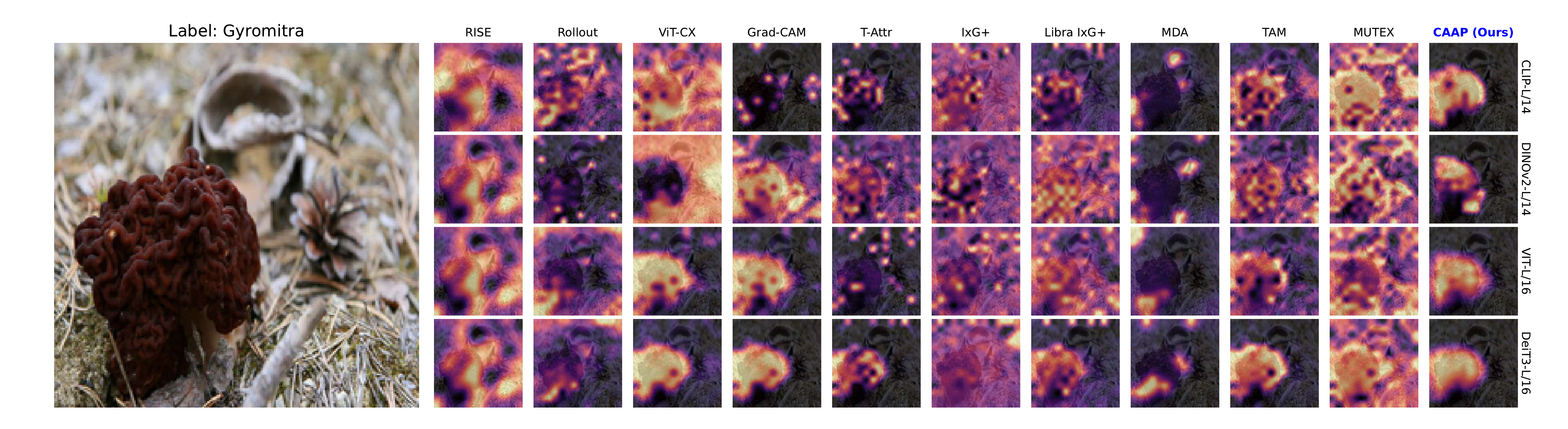}
  \vspace{-1mm}
  \caption{
    Visualization of single-object attribution maps produced by different methods across various models. The target object is a \texttt{Gyromitra}.
    }
  \label{fig:vis_80}
  \vspace{-1mm}
\end{figure}

\begin{figure}[t]
  \centering
  \includegraphics[width=0.99\linewidth]{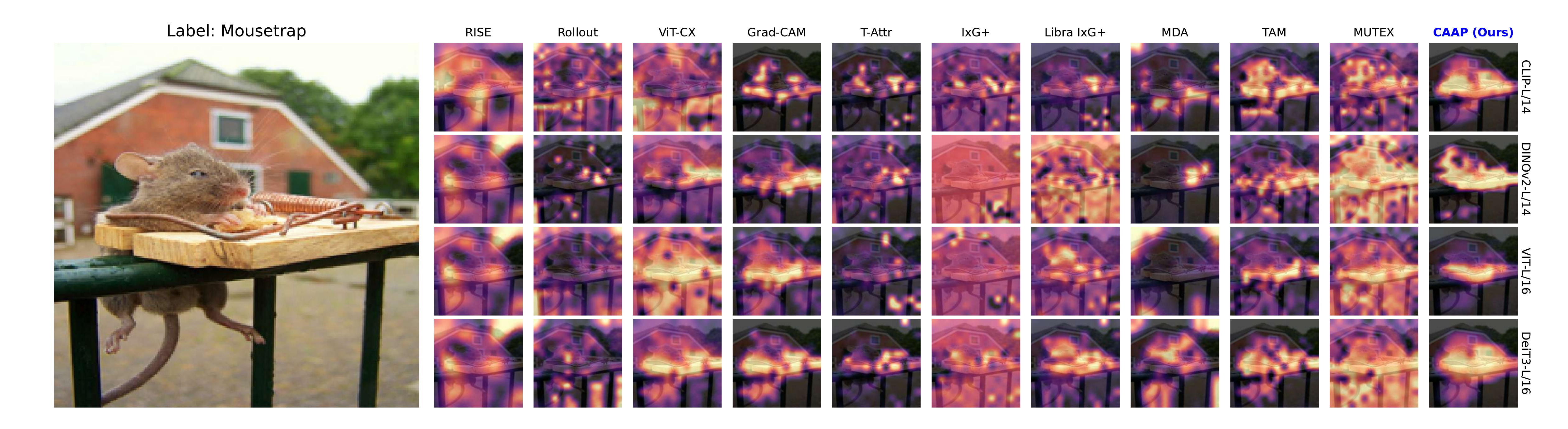}
  \vspace{-1mm}
  \caption{
    Visualization of single-object attribution maps produced by different methods across various models. The target object is a \texttt{Mousetrap}.
    }
  \label{fig:vis_88}
  \vspace{-1mm}
\end{figure}

\begin{figure}[t]
  \centering
  \includegraphics[width=0.99\linewidth]{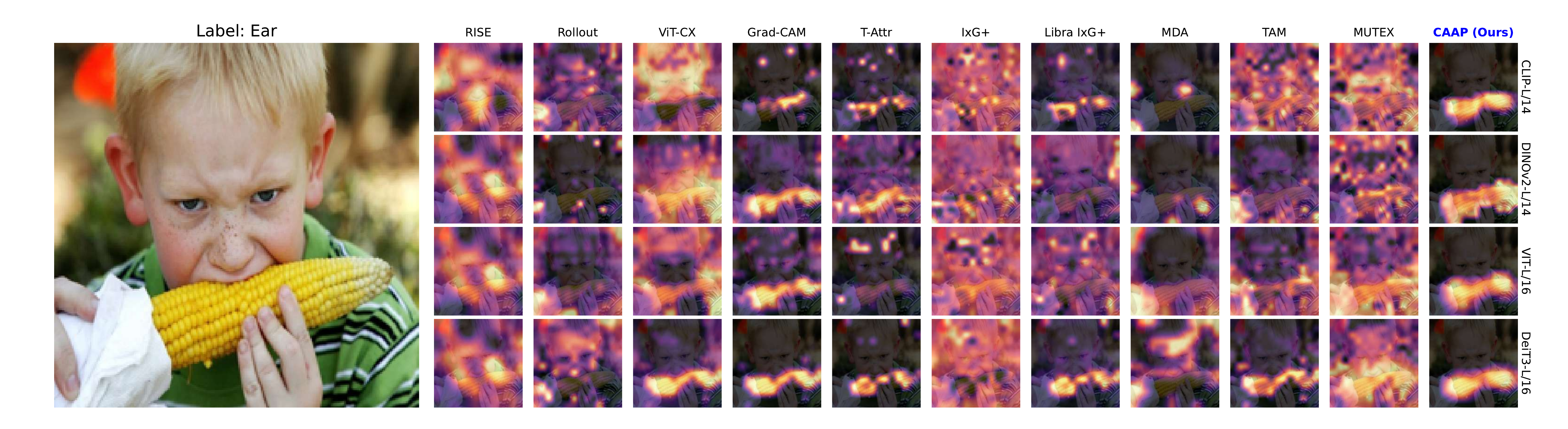}
  \vspace{-1mm}
  \caption{
    Visualization of single-object attribution maps produced by different methods across various models. The target object is an \texttt{Ear}.
    }
  \label{fig:vis_103}
  \vspace{-1mm}
\end{figure}

\begin{figure}[t]
  \centering
  \includegraphics[width=0.99\linewidth]{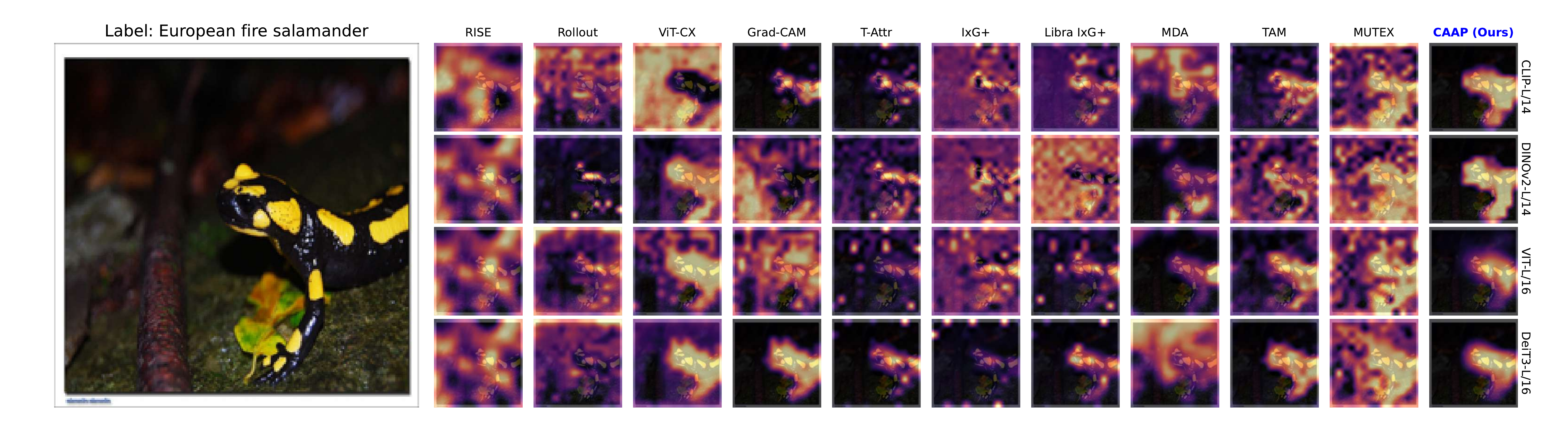}
  \vspace{-1mm}
  \caption{
    Visualization of single-object attribution maps produced by different methods across various models. The target object is a \texttt{European fire salamander}.
    }
  \label{fig:vis_132}
  \vspace{-1mm}
\end{figure}

\begin{figure}[t]
  \centering
  \includegraphics[width=0.99\linewidth]{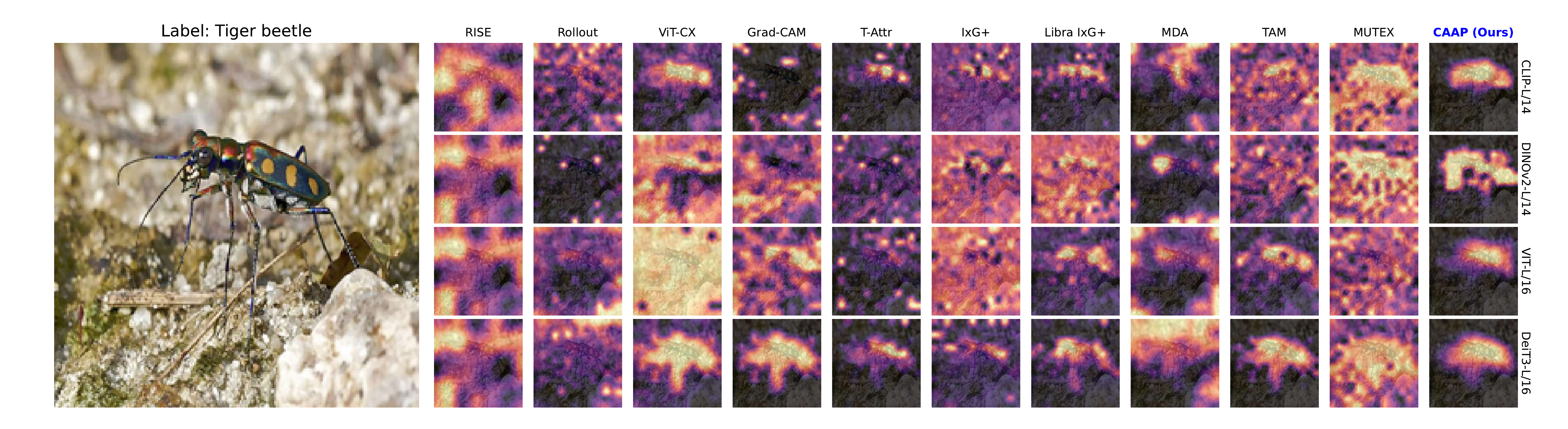}
  \vspace{-1mm}
  \caption{
    Visualization of single-object attribution maps produced by different methods across various models. The target object is a \texttt{Tiger beetle}.
    }
  \label{fig:vis_143}
  \vspace{-1mm}
\end{figure}

\begin{figure}[t]
  \centering
  \includegraphics[width=0.99\linewidth]{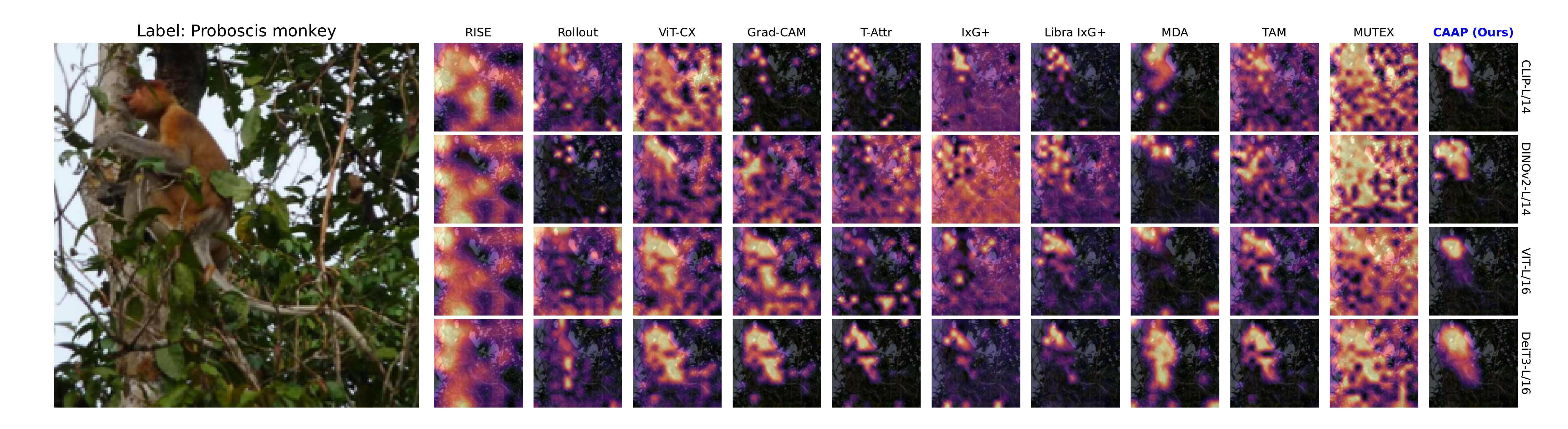}
  \vspace{-1mm}
  \caption{
    Visualization of single-object attribution maps produced by different methods across various models. The target object is a \texttt{Proboscis monkey}.
    }
  \label{fig:vis_156}
  \vspace{-1mm}
\end{figure}

\begin{figure}[t]
  \centering
  \includegraphics[width=0.99\linewidth]{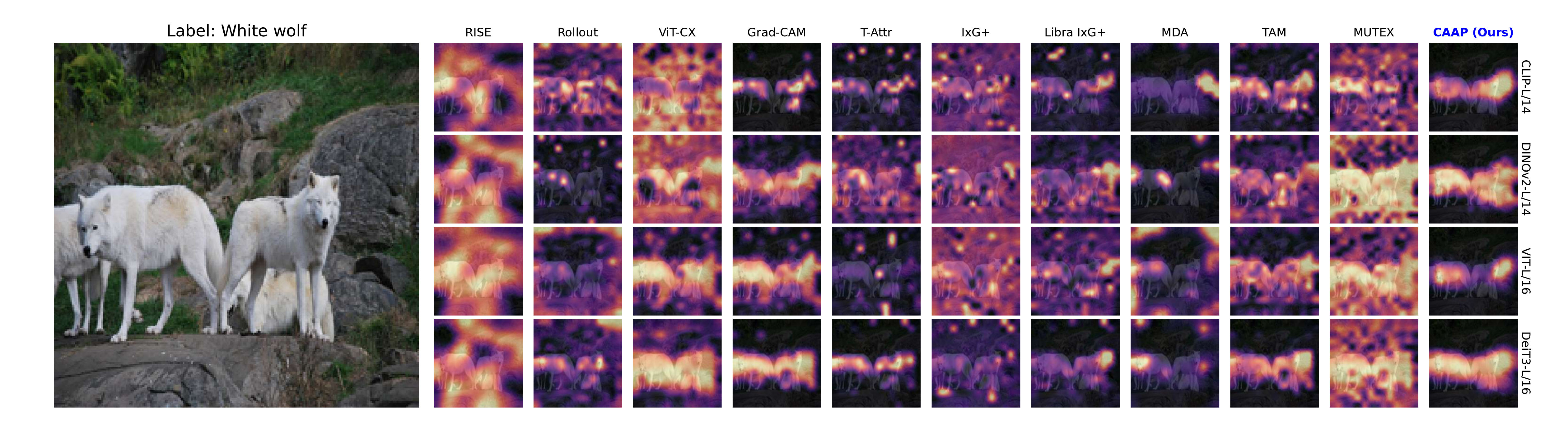}
  \vspace{-1mm}
  \caption{
    Visualization of single-object attribution maps produced by different methods across various models. The target object is a \texttt{White wolf}.
    }
  \label{fig:vis_200}
  \vspace{-1mm}
\end{figure}

\begin{figure}[t]
  \centering
  \includegraphics[width=0.99\linewidth]{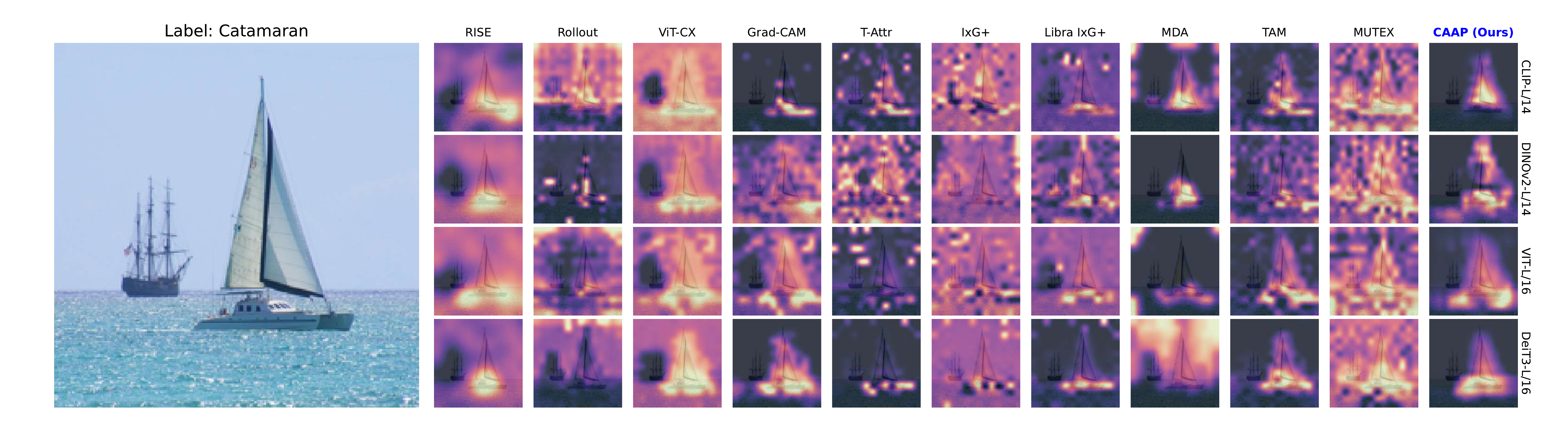}
  \vspace{-1mm}
  \caption{
    Visualization of single-object attribution maps produced by different methods across various models. The target object is a \texttt{Catamaran}.
    }
  \label{fig:vis_212}
  \vspace{-1mm}
\end{figure}

\begin{figure}[t]
  \centering
  \includegraphics[width=0.99\linewidth]{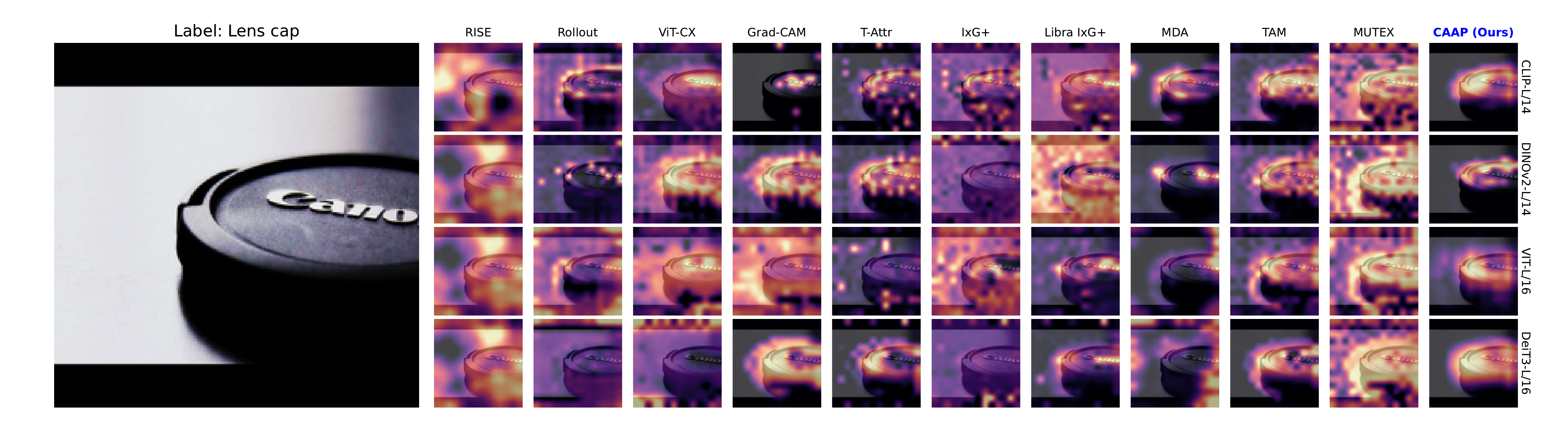}
  \vspace{-1mm}
  \caption{
    Visualization of single-object attribution maps produced by different methods across various models. The target object is a \texttt{Lens cap}.
    }
  \label{fig:vis_234}
  \vspace{-1mm}
\end{figure}

\begin{figure}[t]
  \centering
  \includegraphics[width=0.99\linewidth]{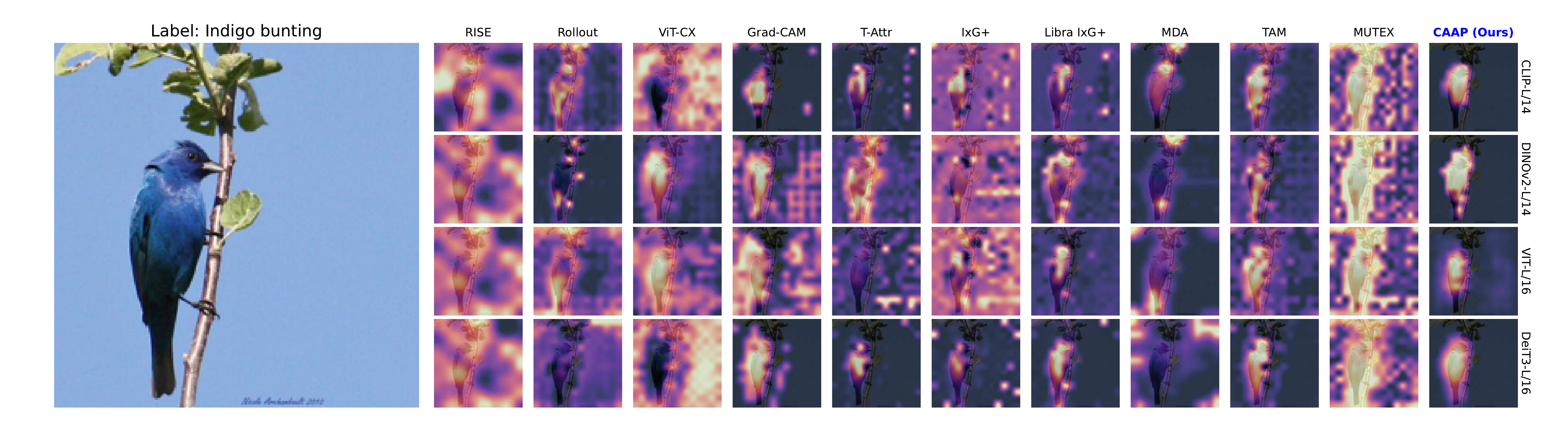}
  \vspace{-1mm}
  \caption{
    Visualization of single-object attribution maps produced by different methods across various models. The target object is an \texttt{Indigo bunting}.
    }
  \label{fig:vis_251}
  \vspace{-1mm}
\end{figure}

\begin{figure}[t]
  \centering
  \includegraphics[width=0.99\linewidth]{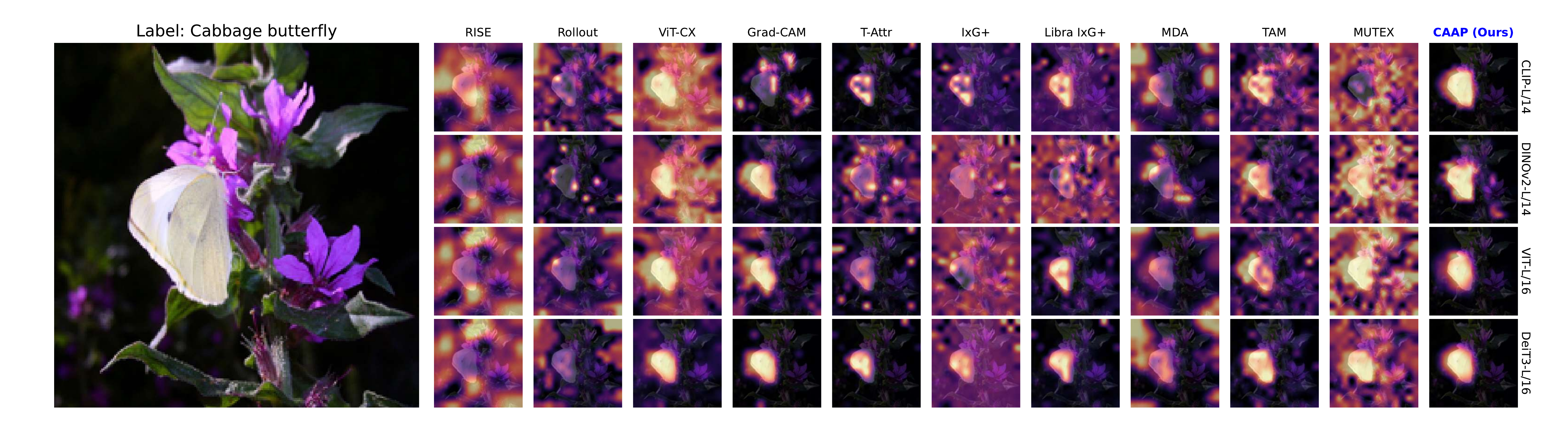}
  \vspace{-1mm}
  \caption{
    Visualization of single-object attribution maps produced by different methods across various models. The target object is a \texttt{Cabbage butterfly}.
    }
  \label{fig:vis_257}
  \vspace{-1mm}
\end{figure}

\begin{figure}[t]
  \centering
  \includegraphics[width=0.99\linewidth]{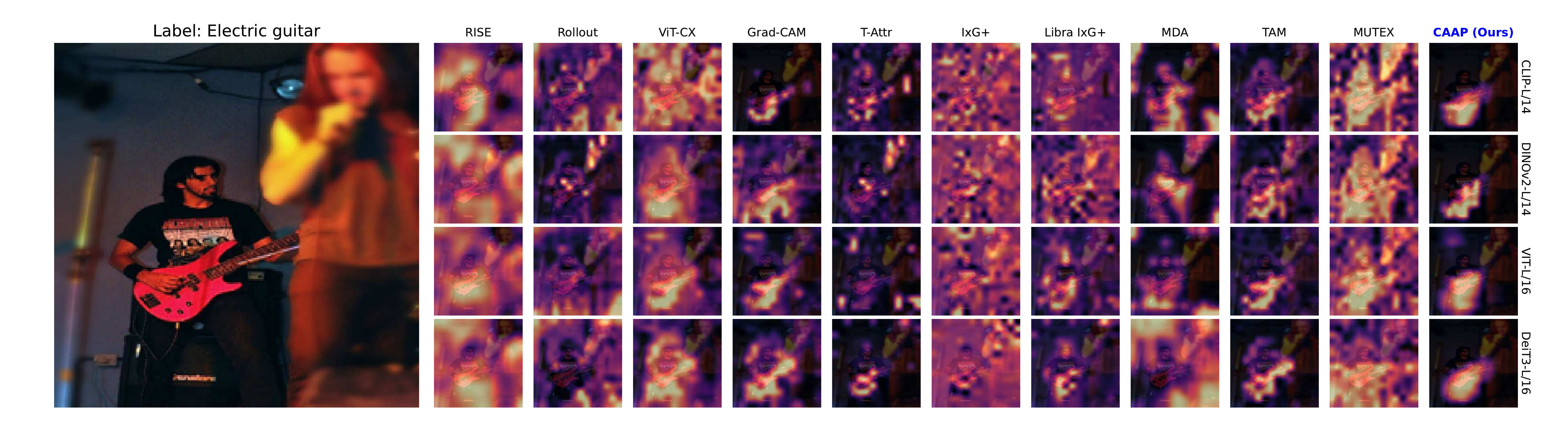}
  \vspace{-1mm}
  \caption{
    Visualization of single-object attribution maps produced by different methods across various models. The target object is an \texttt{Electric guitar}.
    }
  \label{fig:vis_266}
  \vspace{-1mm}
\end{figure}

\begin{figure}[t]
  \centering
  \includegraphics[width=0.99\linewidth]{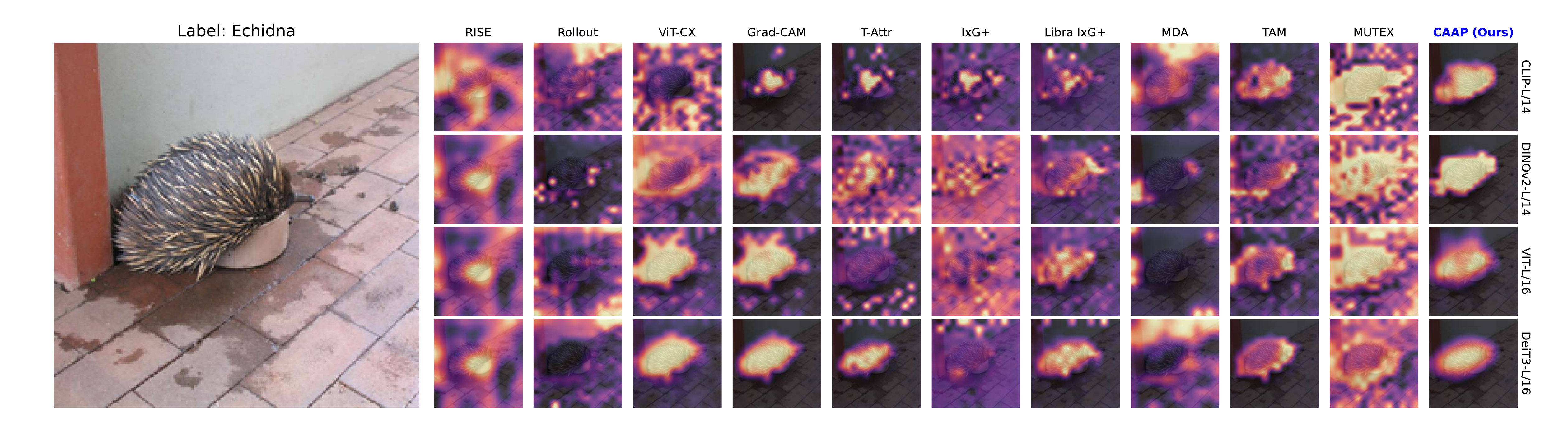}
  \vspace{-1mm}
  \caption{
    Visualization of single-object attribution maps produced by different methods across various models. The target object is an \texttt{Echidna}.
    }
  \label{fig:vis_278}
  \vspace{-1mm}
\end{figure}

\begin{figure}[t]
  \centering
  \includegraphics[width=0.99\linewidth]{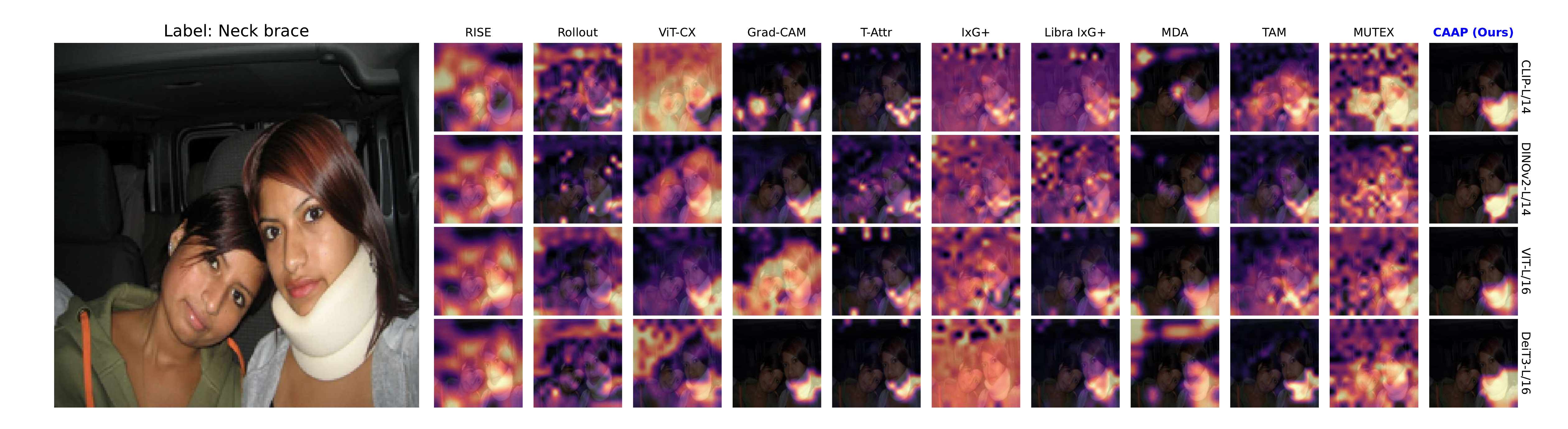}
  \vspace{-1mm}
  \caption{
    Visualization of single-object attribution maps produced by different methods across various models. The target object is a \texttt{Neck brace}.
    }
  \label{fig:vis_291}
  \vspace{-1mm}
\end{figure}

\begin{figure}[t]
  \centering
  \includegraphics[width=0.99\linewidth]{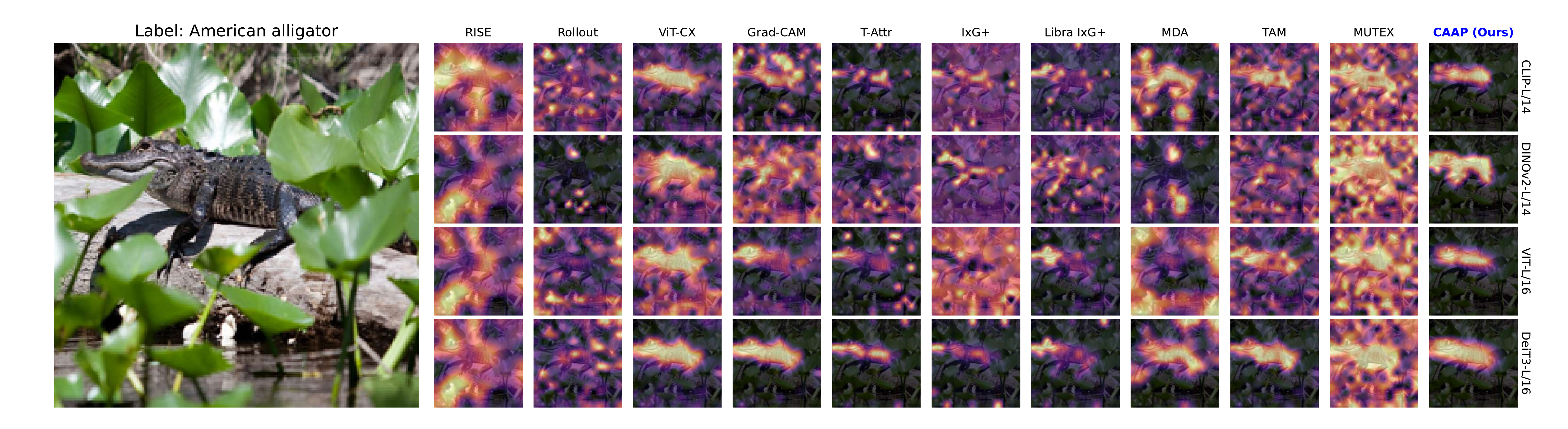}
  \vspace{-1mm}
  \caption{
    Visualization of single-object attribution maps produced by different methods across various models. The target object is an \texttt{American alligator}.
    }
  \label{fig:vis_294}
  \vspace{-1mm}
\end{figure}

\begin{figure}[t]
  \centering
  \includegraphics[width=0.99\linewidth]{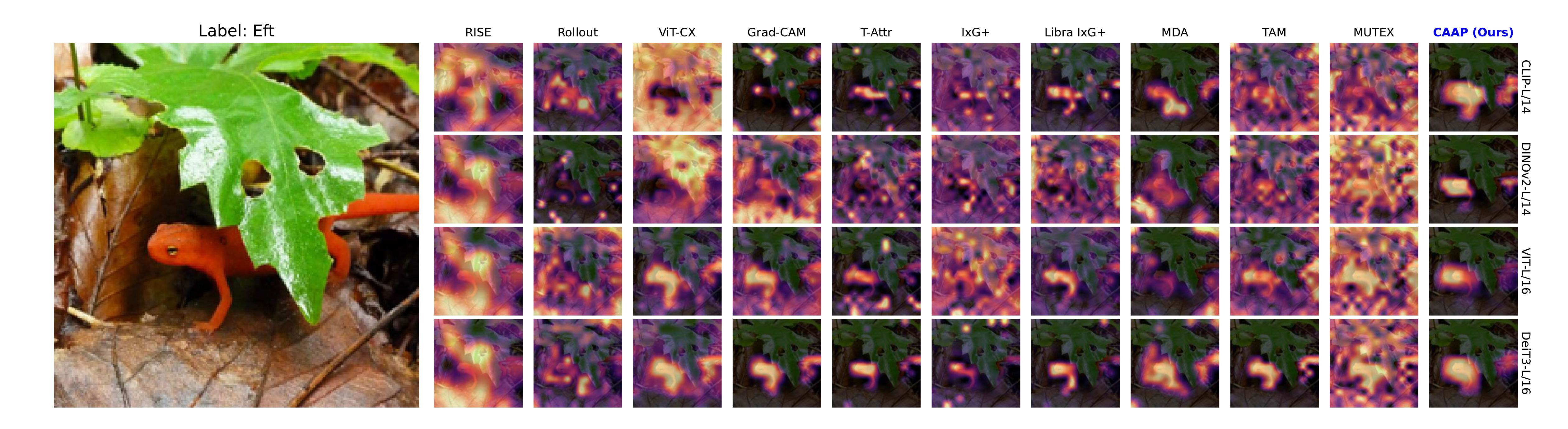}
  \vspace{-1mm}
  \caption{
    Visualization of single-object attribution maps produced by different methods across various models. The target object is an \texttt{Eft}.
    }
  \label{fig:vis_304}
  \vspace{-1mm}
\end{figure}

\begin{figure}[t]
  \centering
  \includegraphics[width=0.99\linewidth]{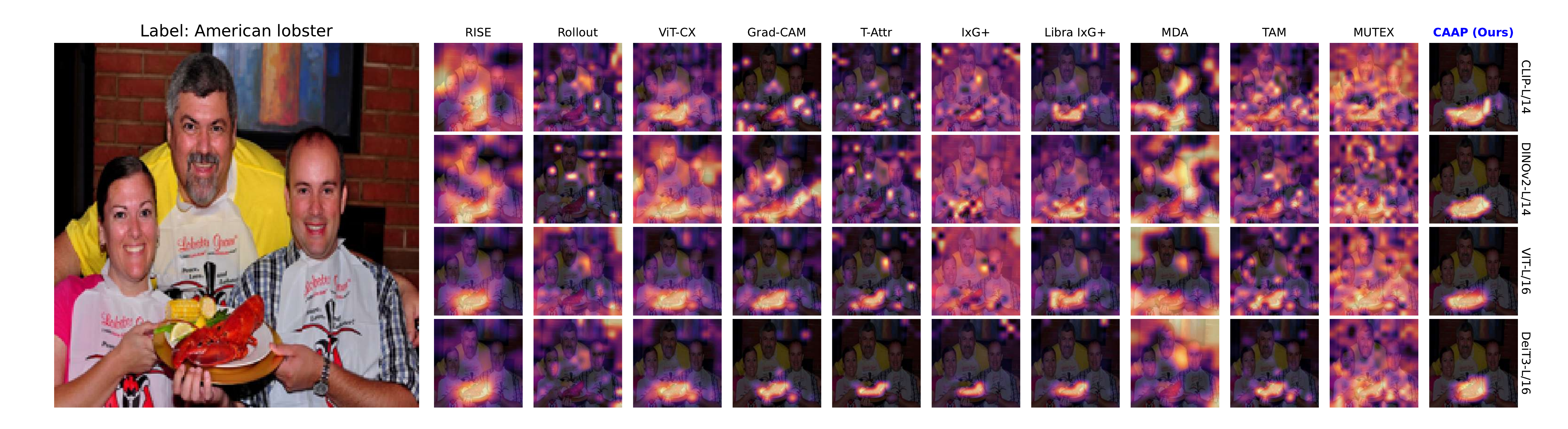}
  \vspace{-1mm}
  \caption{
    Visualization of single-object attribution maps produced by different methods across various models. The target object is an \texttt{American lobster}.
    }
  \label{fig:vis_322}
  \vspace{-1mm}
\end{figure}

\begin{figure}[t]
  \centering
  \includegraphics[width=0.99\linewidth]{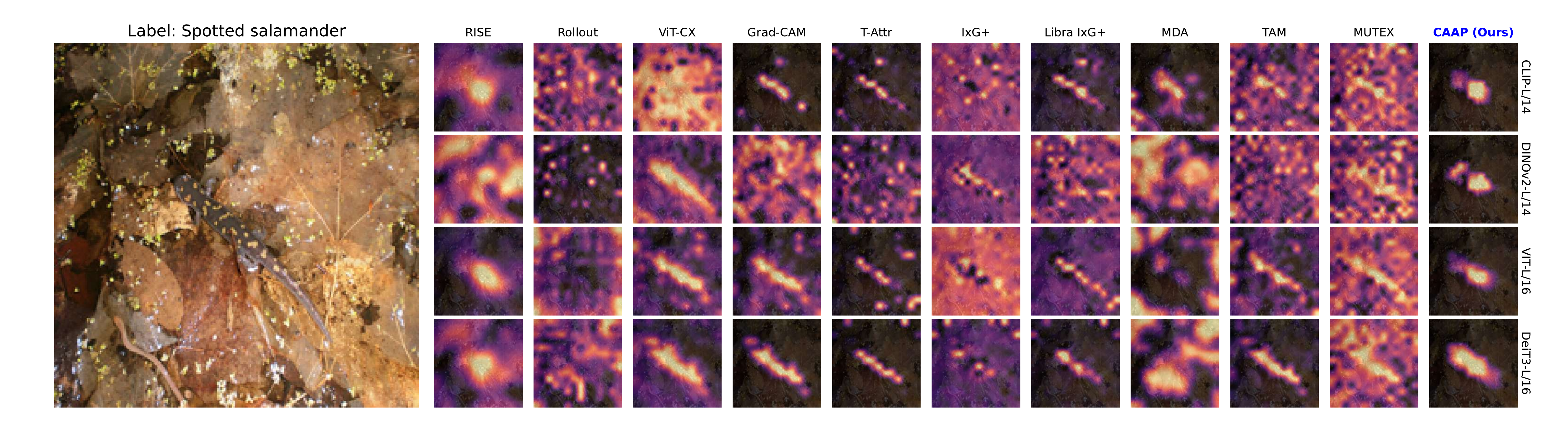}
  \vspace{-1mm}
  \caption{
    Visualization of single-object attribution maps produced by different methods across various models. The target object is a \texttt{Spotted salamander}.
    }
  \label{fig:vis_323}
  \vspace{-1mm}
\end{figure}

\begin{figure}[t]
  \centering
  \includegraphics[width=0.99\linewidth]{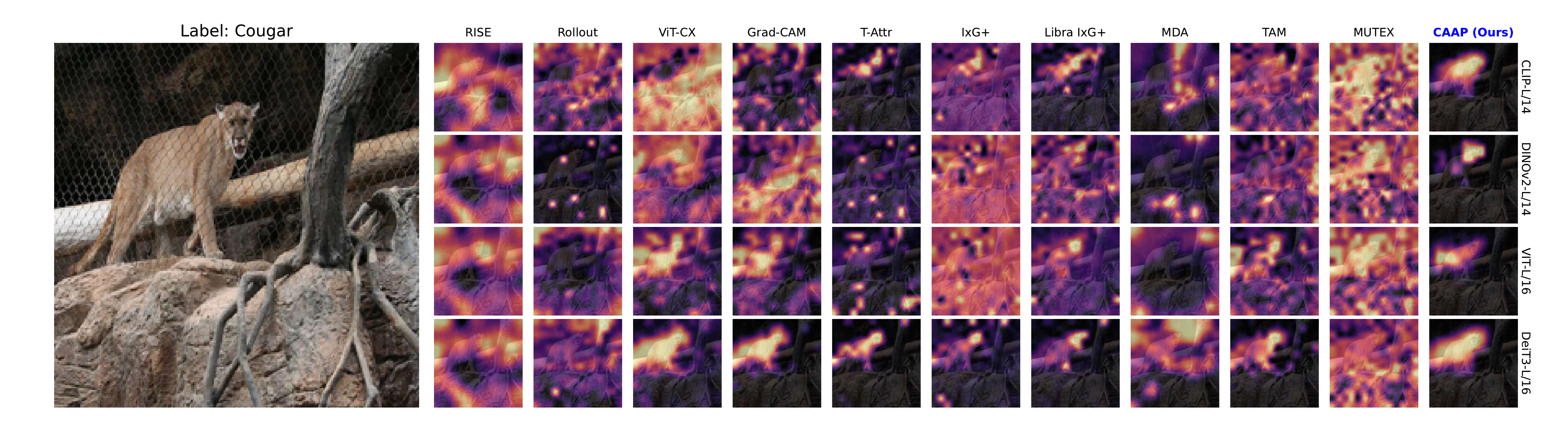}
  \vspace{-1mm}
  \caption{
    Visualization of single-object attribution maps produced by different methods across various models. The target object is a \texttt{Cougar}.
    }
  \label{fig:vis_326}
  \vspace{-1mm}
\end{figure}

\begin{figure}[t]
  \centering
  \includegraphics[width=0.99\linewidth]{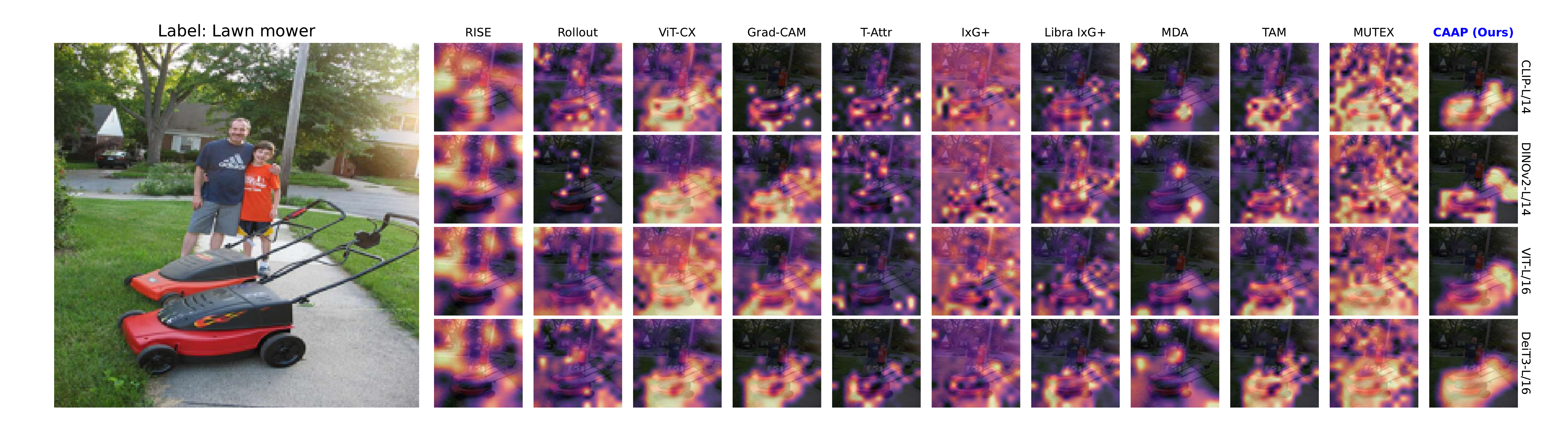}
  \vspace{-1mm}
  \caption{
    Visualization of single-object attribution maps produced by different methods across various models. The target object is a \texttt{Lawn mower}.
    }
  \label{fig:vis_332}
  \vspace{-1mm}
\end{figure}

\begin{figure}[t]
  \centering
  \includegraphics[width=0.99\linewidth]{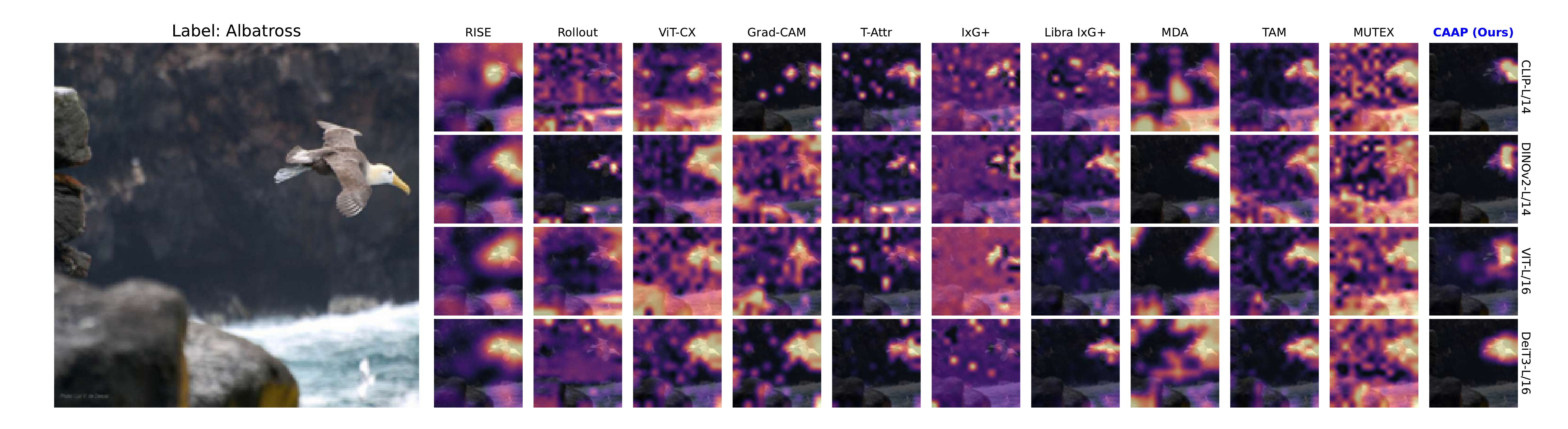}
  \vspace{-1mm}
  \caption{
    Visualization of single-object attribution maps produced by different methods across various models. The target object is an \texttt{Albatross}.
    }
  \label{fig:vis_334}
  \vspace{-1mm}
\end{figure}

\begin{figure}[t]
  \centering
  \includegraphics[width=0.99\linewidth]{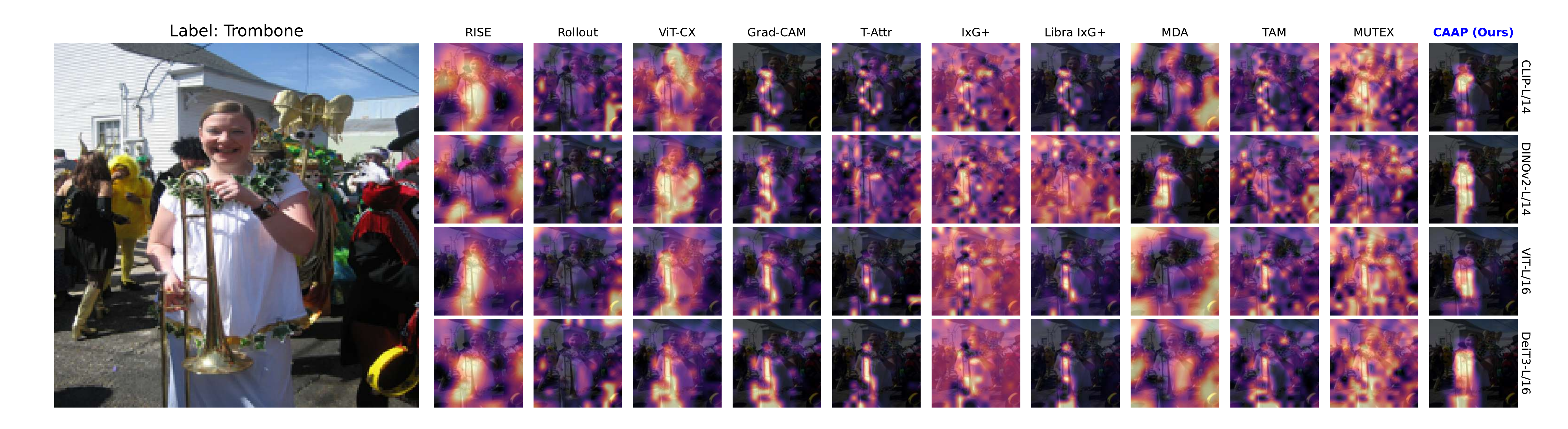}
  \vspace{-1mm}
  \caption{
    Visualization of single-object attribution maps produced by different methods across various models. The target object is a \texttt{Trombone}.
    }
  \label{fig:vis_348}
  \vspace{-1mm}
\end{figure}

\begin{figure}[t]
  \centering
  \includegraphics[width=0.99\linewidth]{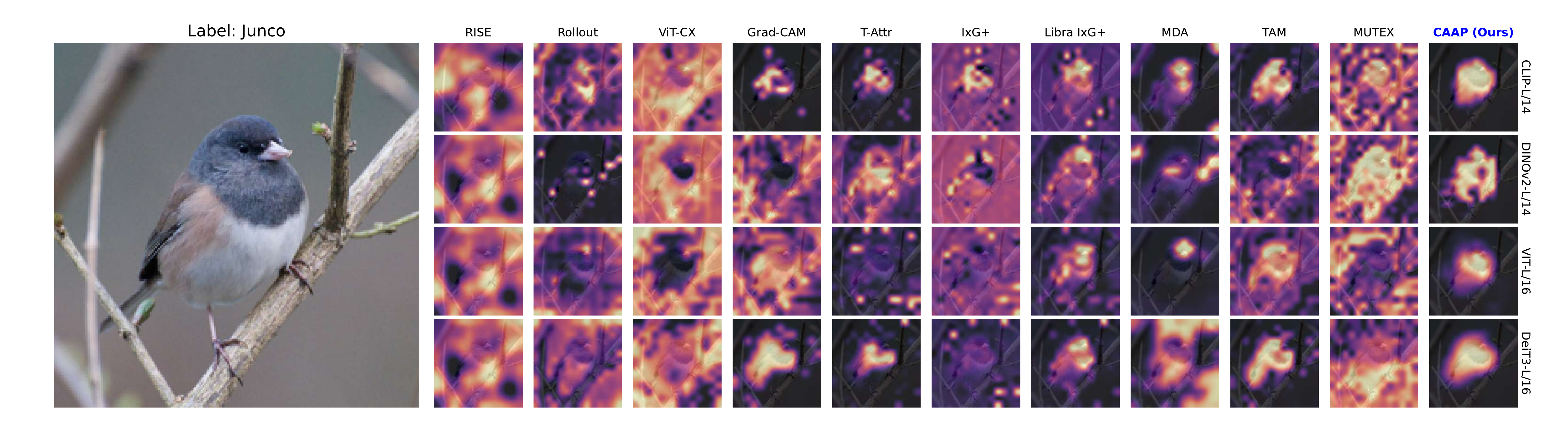}
  \vspace{-1mm}
  \caption{
    Visualization of single-object attribution maps produced by different methods across various models. The target object is a \texttt{Junco}.
    }
  \label{fig:vis_366}
  \vspace{-1mm}
\end{figure}

\begin{figure}[t]
  \centering
  \includegraphics[width=0.99\linewidth]{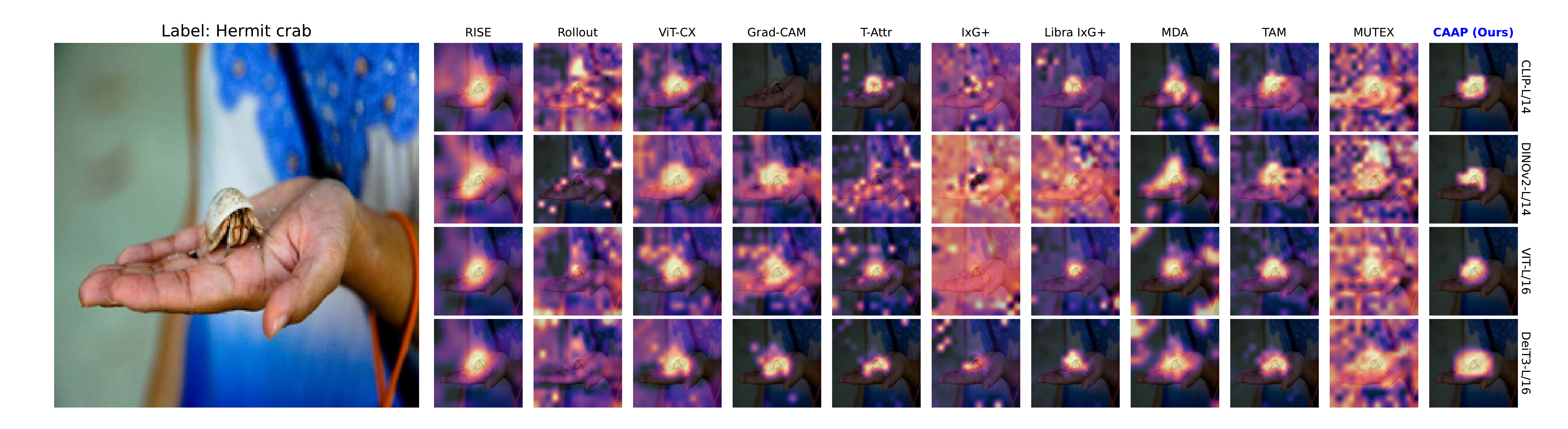}
  \vspace{-1mm}
  \caption{
    Visualization of single-object attribution maps produced by different methods across various models. The target object is a \texttt{Hermit crab}.
    }
  \label{fig:vis_381}
  \vspace{-1mm}
\end{figure}

\begin{figure}[t]
  \centering
  \includegraphics[width=0.99\linewidth]{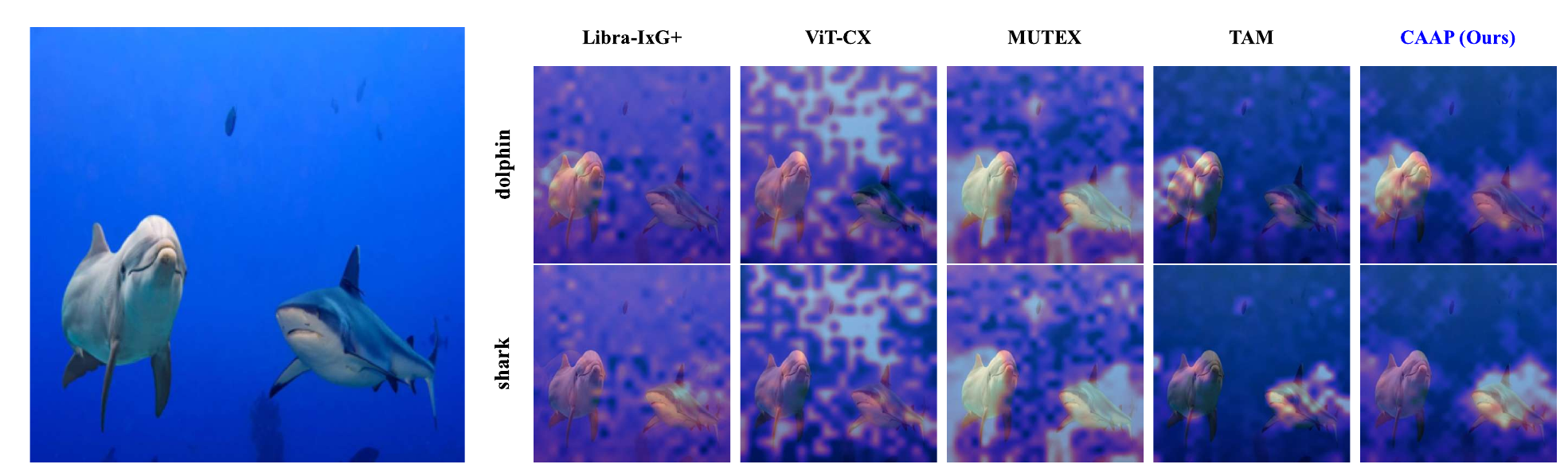}
  \vspace{-1mm}
  \caption{
    Visualization of multi-object attribution maps produced by a selected set of strong methods. The target objects are a \texttt{dolphin} and a \texttt{shark}.
    }
  \label{fig:dolphin_shark_1}
  \vspace{-1mm}
\end{figure}

\begin{figure}[t]
  \centering
  \includegraphics[width=0.99\linewidth]{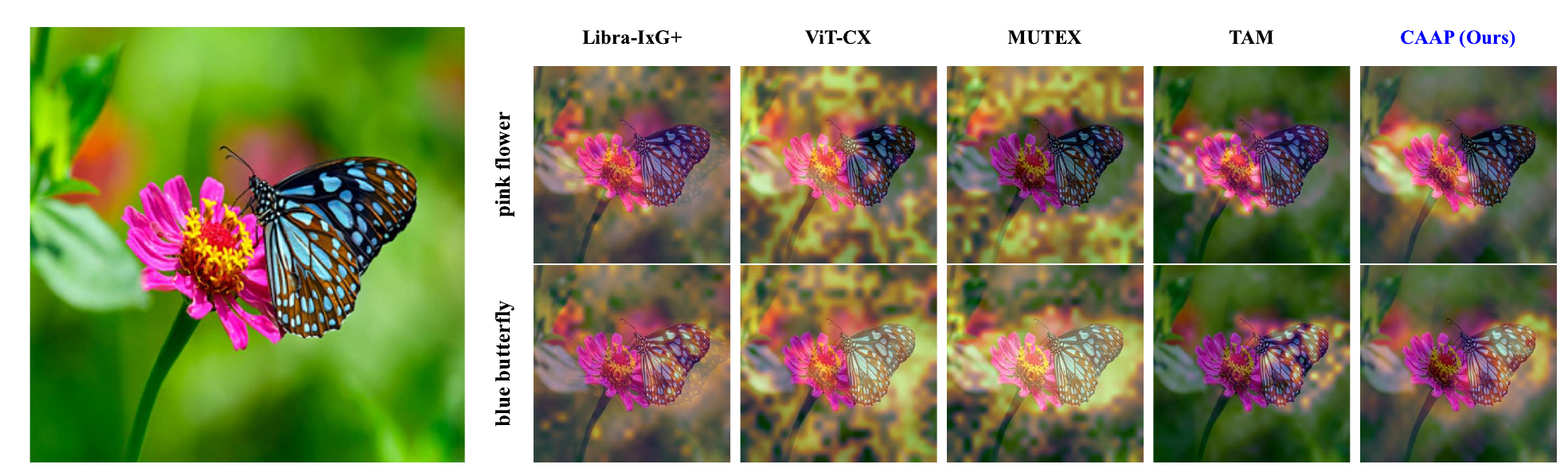}
  \vspace{-1mm}
  \caption{
    Visualization of multi-object attribution maps produced by a selected set of strong methods. The target objects are a \texttt{pink flower} and a \texttt{blue butterfly}.
    }
  \label{fig:flower_butterfly_1}
  \vspace{-1mm}
\end{figure}

\begin{figure}[t]
  \centering
  \includegraphics[width=0.99\linewidth]{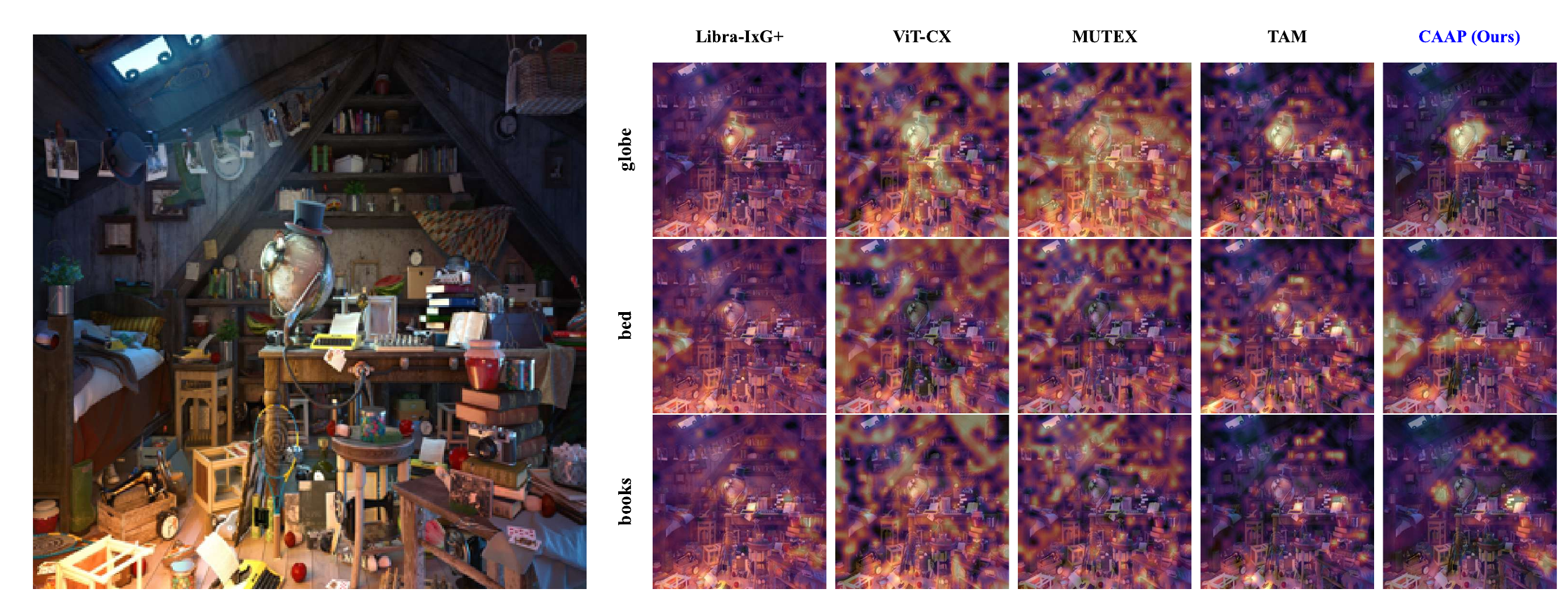}
  \vspace{-1mm}
  \caption{
    Visualization of multi-object attribution maps produced by a selected set of strong methods. The target objects are a \texttt{globe}, a \texttt{bed}, and \texttt{books}.
    }
  \label{fig:hidden_objects_1}
  \vspace{-1mm}
\end{figure}

\begin{figure}[t]
  \centering
  \includegraphics[width=0.99\linewidth]{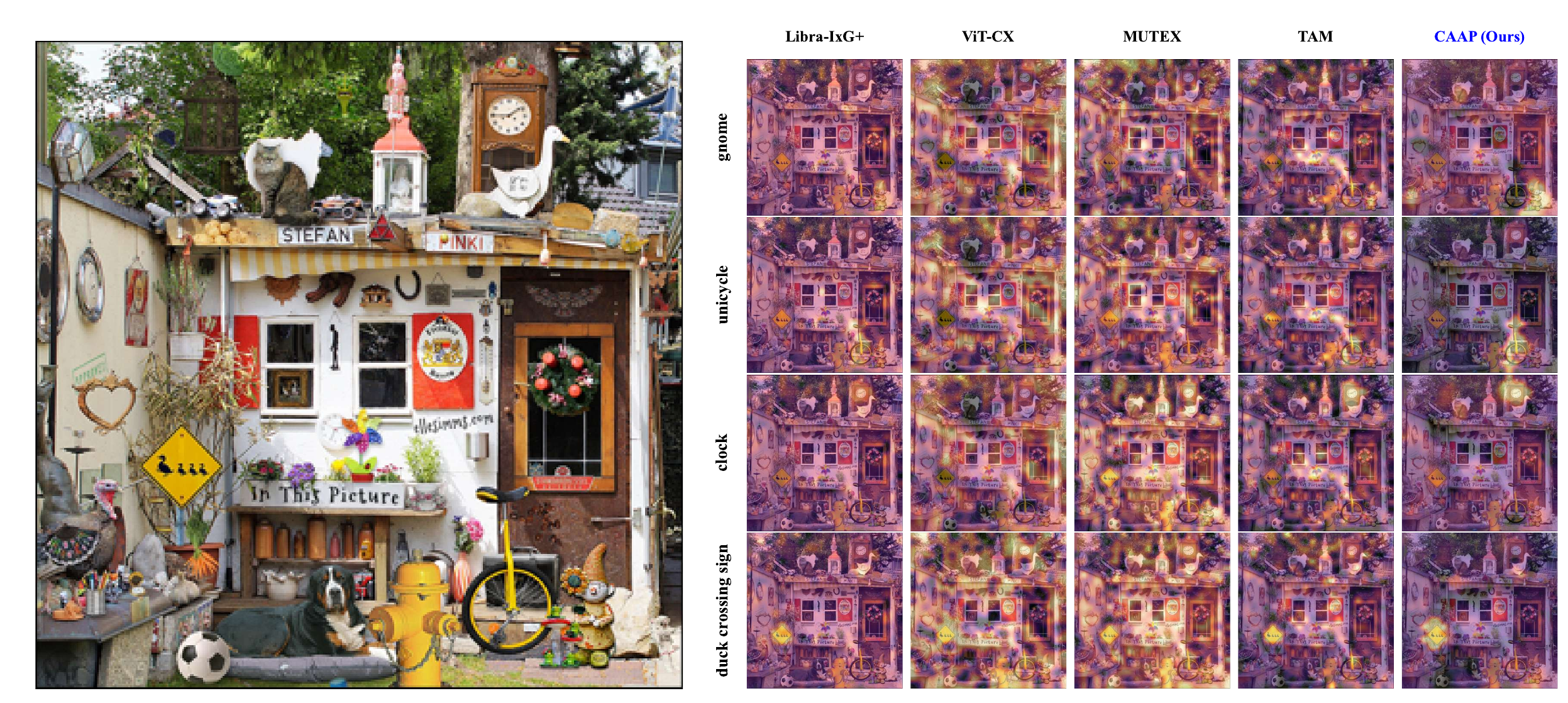}
  \vspace{-1mm}
  \caption{
    Visualization of multi-object attribution maps produced by a selected set of strong methods. The target objects are a \texttt{gnome}, a \texttt{unicycle}, a \texttt{clock}, and a \texttt{duck crossing sign}.
    }
  \label{fig:hidden_objects_2}
  \vspace{-1mm}
\end{figure}

\begin{figure}[t]
  \centering
  \includegraphics[width=0.99\linewidth]{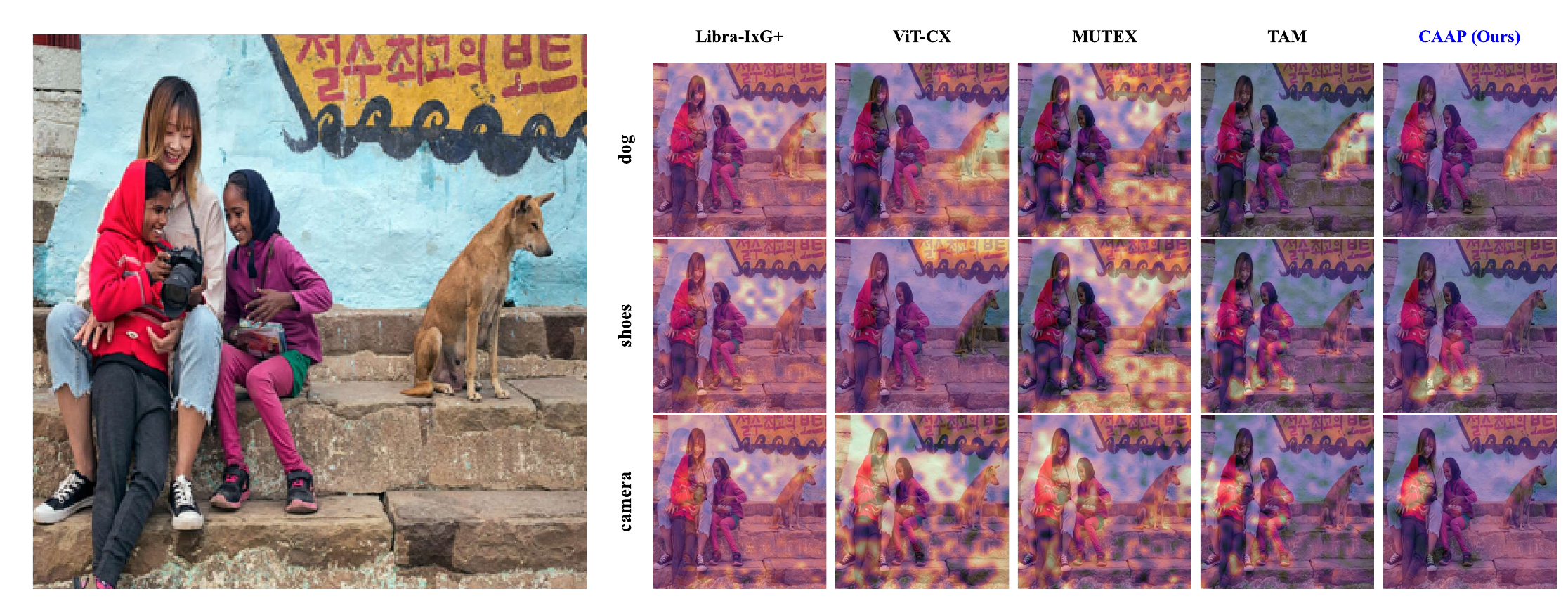}
  \vspace{-1mm}
  \caption{
    Visualization of multi-object attribution maps produced by a selected set of strong methods. The target objects are a \texttt{dog}, \texttt{shoes}, and a \texttt{camera}.
    }
  \label{fig:indians_2}
  \vspace{-1mm}
\end{figure}

\begin{figure}[t]
  \centering
  \includegraphics[width=0.99\linewidth]{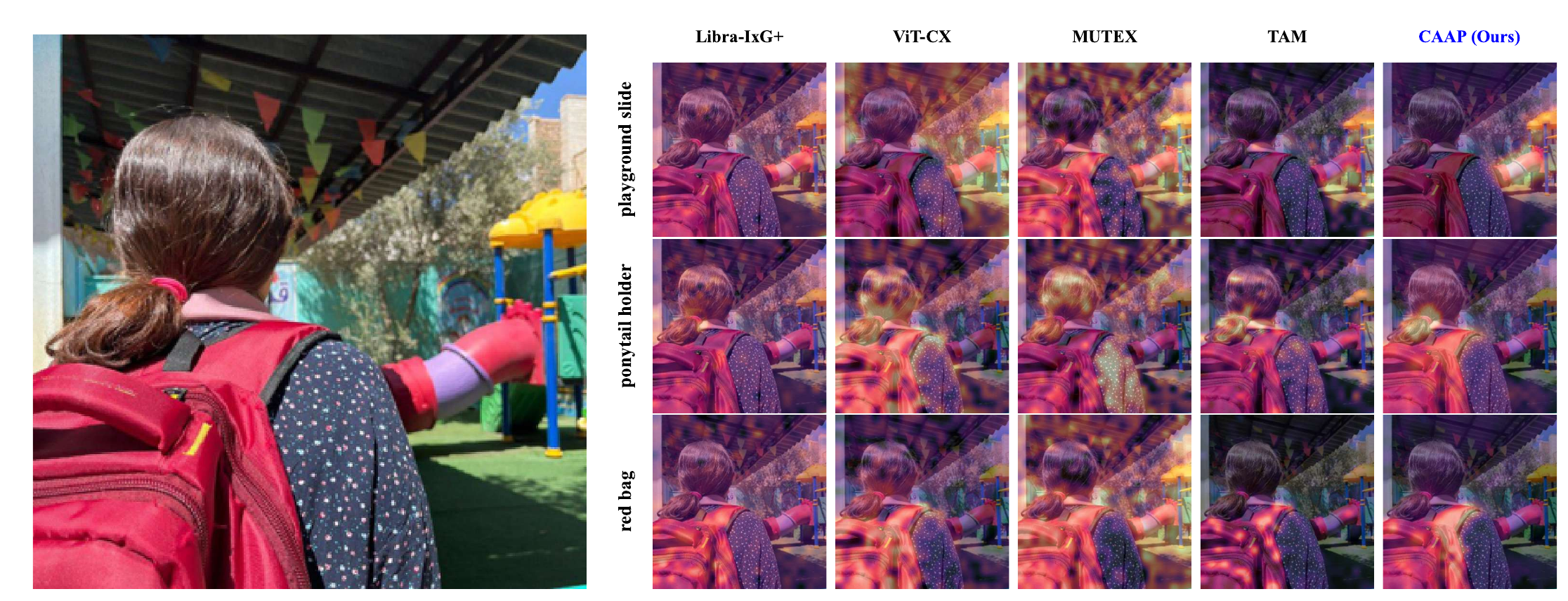}
  \vspace{-1mm}
  \caption{
    Visualization of multi-object attribution maps produced by a selected set of strong methods. The target objects are a \texttt{playground slide}, a \texttt{ponytail holder}, and a \texttt{red bag}.
    }
  \label{fig:mi_1}
  \vspace{-1mm}
\end{figure}

\begin{figure}[t]
  \centering
  \includegraphics[width=0.99\linewidth]{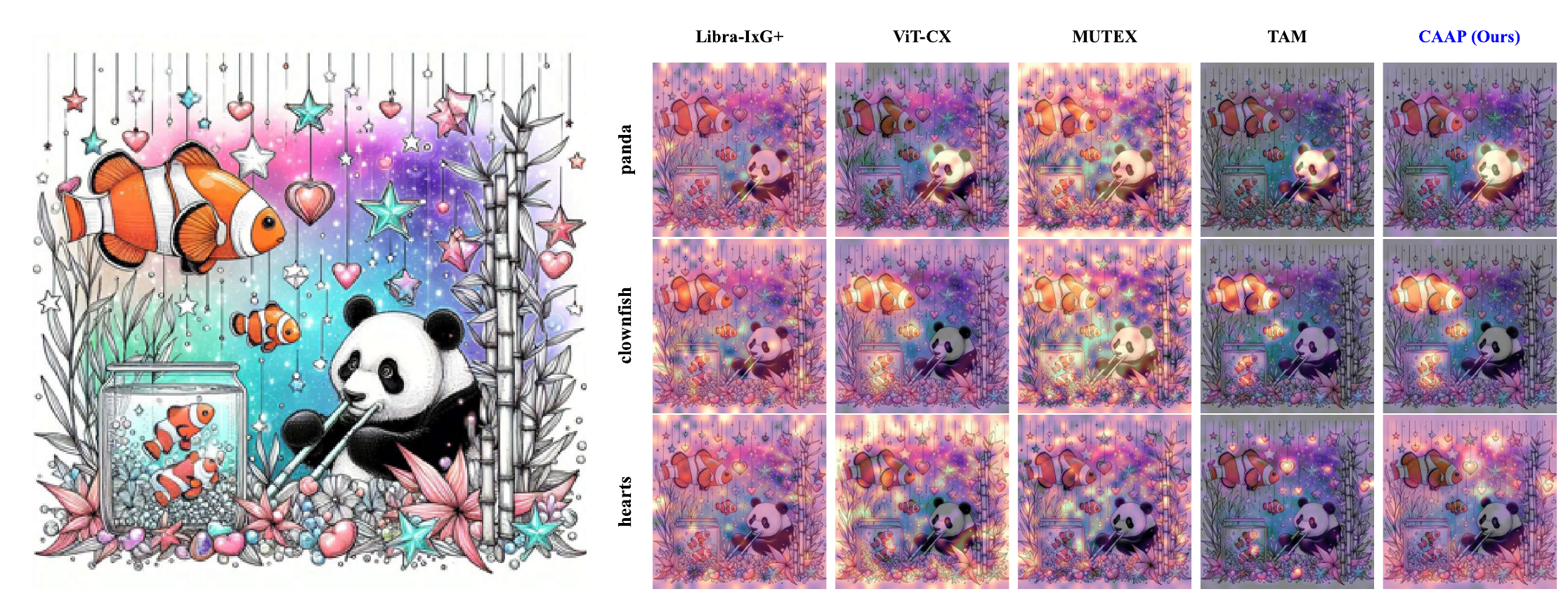}
  \vspace{-1mm}
  \caption{
    Visualization of multi-object attribution maps produced by a selected set of strong methods. The target objects are a \texttt{panda}, a \texttt{clownfish}, and \texttt{hearts}.
    }
  \label{fig:panda_clownfish_1}
  \vspace{-1mm}
\end{figure}

\begin{figure}[t]
  \centering
  \includegraphics[width=0.99\linewidth]{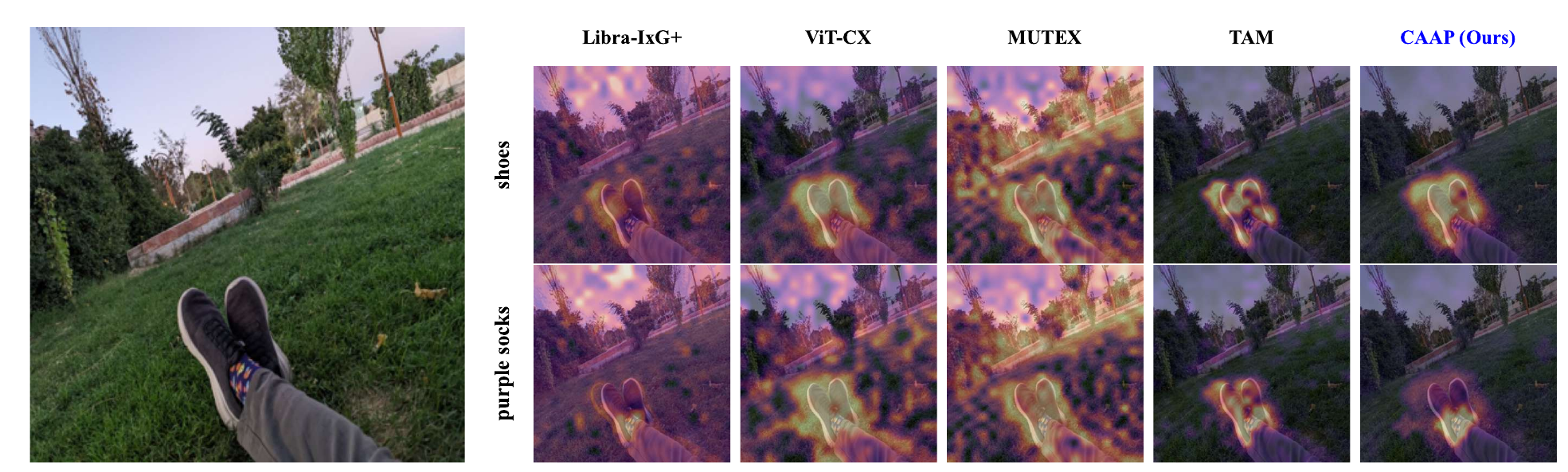}
  \vspace{-1mm}
  \caption{
    Visualization of multi-object attribution maps produced by a selected set of strong methods. The target objects are \texttt{shoes} and \texttt{purple socks}.
    }
  \label{fig:park_salamat_1}
  \vspace{-1mm}
\end{figure}

\begin{figure}[t]
  \centering
  \includegraphics[width=0.99\linewidth]{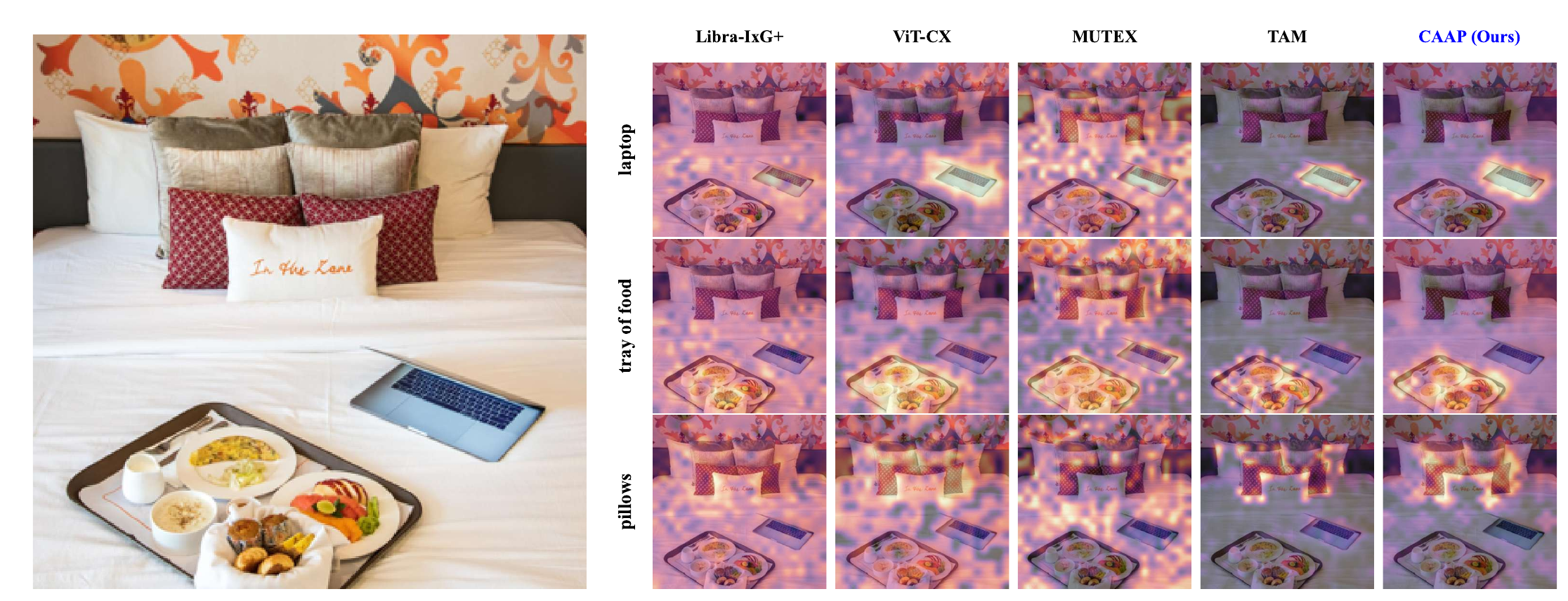}
  \vspace{-1mm}
  \caption{
    Visualization of multi-object attribution maps produced by a selected set of strong methods. The target objects are a \texttt{laptop}, a \texttt{tray of food}, and \texttt{pillows}.
    }
  \label{fig:room_3}
  \vspace{-1mm}
\end{figure}

\begin{figure}[t]
  \centering
  \includegraphics[width=0.99\linewidth]{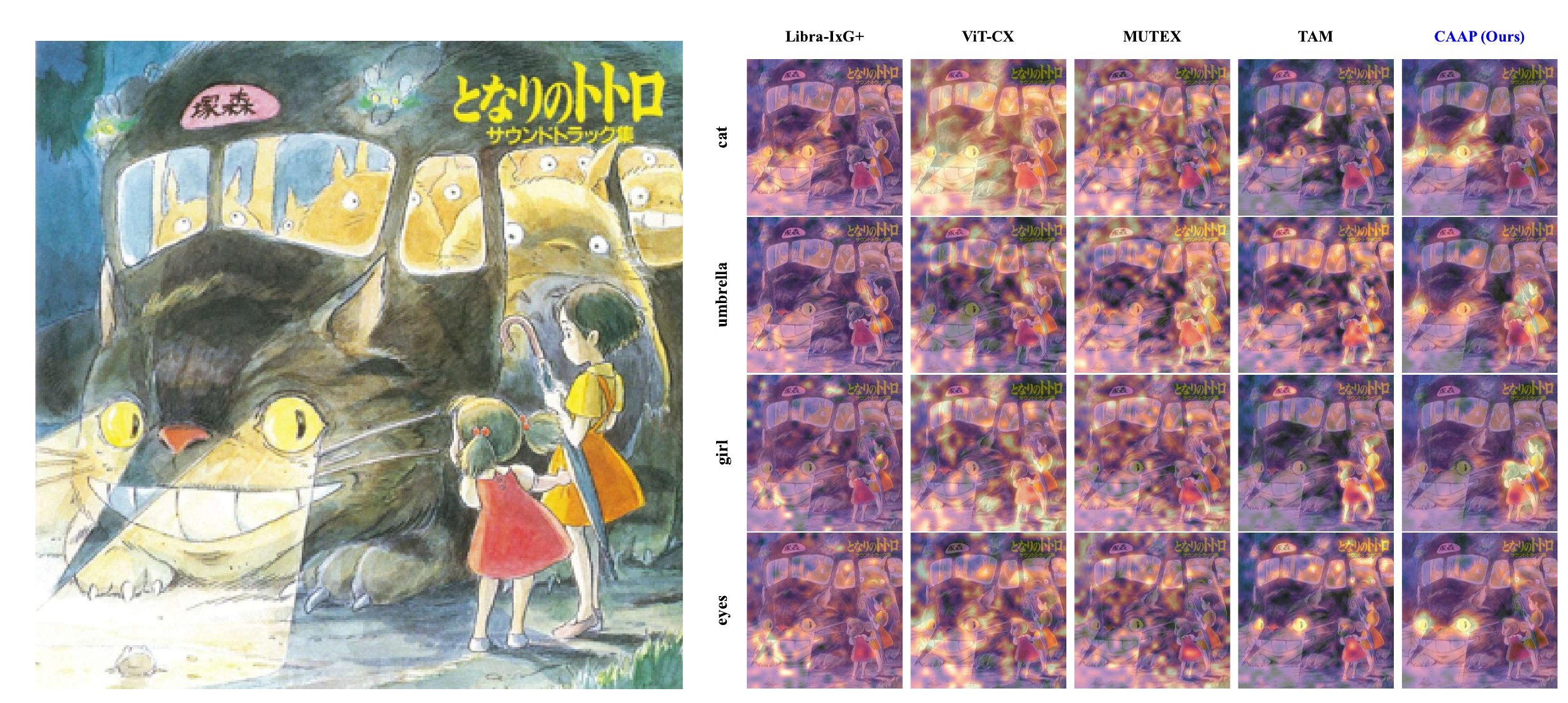}
  \vspace{-1mm}
  \caption{
    Visualization of multi-object attribution maps produced by a selected set of strong methods. The target objects are a \texttt{cat}, an \texttt{umbrella}, a \texttt{girl}, and \texttt{eyes}.
    }
  \label{fig:totoro_1}
  \vspace{-1mm}
\end{figure}

We present additional qualitative results to further illustrate the behavior of our attribution method across different object categories. As shown in Figures~\ref{fig:vis_11}--\ref{fig:totoro_1}, our method consistently produces more precise and spatially coherent attribution maps compared to the baselines. In particular, the highlighted regions align more closely with the true object extents and exhibit sharper object boundaries, while reducing spurious activations in background regions. These examples demonstrate the robustness of our approach across diverse visual concepts, including objects with varying shapes, textures, and levels of visual complexity.

\section{Efficient Implementation via Blank-Context Precomputation}
\label{app:efficiency}

A large fraction of the computation in CAAP arises from attention between each classification token ($\mathrm{CLS}$) and patch tokens originating from the blank image $x_0$. Let $K_0^l$ and $V_0^l$ denote the key and value matrices at layer $l$ computed from $x_0$. These quantities are fixed and do not depend on the source image $x$ or the selected patch $p_i$.

In the patched context $c^*$, the update of each classification token $\mathrm{CLS}_i$ follows the standard self-attention operation,
\begin{equation}
z_{\mathrm{CLS}}^{l+1}
=
\sum_{j} 
\frac{\exp(\langle q_{\mathrm{CLS}}^{l}, k_{j}^{l}\rangle / \sqrt{d})}
{\sum_{m} \exp(\langle q_{\mathrm{CLS}}^{l}, k_{m}^{l}\rangle / \sqrt{d})}
\, v_{j}^{l}.
\end{equation}
Here, tokens are divided into source-patched tokens $j \in \mathcal{S}(p_i)$ and blank tokens $j \notin \mathcal{S}(p_i)$, with keys and values drawn accordingly.

Empirically, we observe that for blank tokens the dot products $\langle q_{\mathrm{CLS}}^{c^*(p_i)}, k_{0,j}^{l}\rangle$ are nearly identical to those obtained in the blank context $c'$. As a result, the attention distribution over blank tokens is effectively invariant to the inserted source activations.

We therefore precompute two quantities for each layer and attention head using the blank context. First, the normalization contribution of blank tokens,
\begin{equation}
Z_0^{l} = \sum_{j \notin \mathcal{S}(p_i)} 
\exp(\langle q_{\mathrm{CLS}}^{c'}, k_{0,j}^{l}\rangle / \sqrt{d}),
\end{equation}
which is reused as a fixed term in the softmax denominator. Second, a prototype value vector for the blank tokens,
\begin{equation}
\bar{v}_0^{l} = \frac{1}{|\mathcal{B}|} \sum_{j \in \mathcal{B}} v_{0,j}^{l},
\end{equation}
where $\mathcal{B}$ denotes the set of blank tokens.

During attribution, we reuse $Z_0^{l}$ and $\bar{v}_0^{l}$ and compute only the terms involving the source-patched tokens dynamically. This avoids repeated evaluation of dot products and value aggregations for blank tokens across images.

This approximation substantially reduces computational complexity. A naive implementation requires approximately $\mathcal{O}(N^3)$ operations for an image with $N$ tokens, due to $N$ patch interventions combined with $N \times N$ attention interactions. With precomputation, the contribution of blank tokens becomes a constant term ($\mathcal{O}(1)$), and only the $N$ patch interventions are evaluated dynamically, reducing the overall cost to approximately $\mathcal{O}(N)$.

Tables~\ref{tab:faithfulness_caap_ecaap_comparison} and~\ref{tab:localization_caap_ecaap_comparison} compare the original CAAP implementation with the efficient version (E-CAAP). Both variants produce nearly identical results across all four ViT backbones. Faithfulness metrics (Deletion, Insertion, and Ins$-$Del) and localization metrics (AUPR$_1$, AUPR$_0$, and PG) remain very close, indicating that the approximation does not affect attribution quality while substantially reducing computational cost.

\input{tables/ablation_efficiency}

We further compare the computational efficiency of CAAP
and E-CAAP with existing attribution methods in terms
of wall-clock runtime. Table~\ref{tab:cr_runtime}
reports the average attribution time per image.
CAAP requires 8.04 seconds per image, while E-CAAP
reduces this to 1.31 seconds, achieving approximately
a sixfold speedup. The measured speedup is smaller
than the theoretical complexity reduction because
the original CAAP implementation benefits from
greater parallelism.

\input{tables/runtime}

\section{Broader Ablation Analysis}

\subsection{Ablation Analysis across ViT Models}
\label{app:models_ablation}

\begin{figure}[h]
  \centering
  \includegraphics[width=\textwidth]{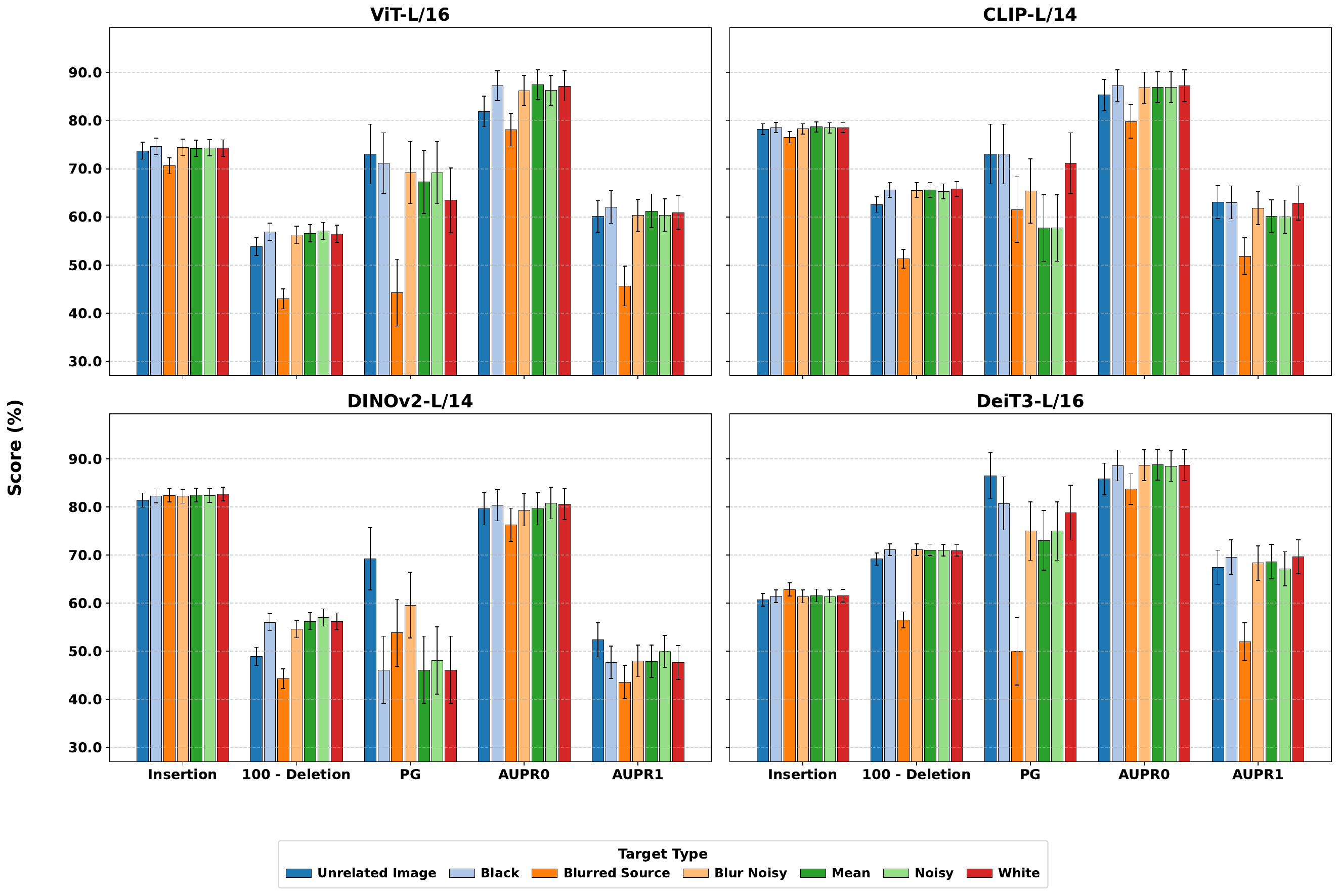}
    \caption{
    Target-context ablation on ImageNet across four
    ViT backbones (ViT-L/16, CLIP-L/14, DINOv2-L/14,
    and DeiT3-L/16). Seven target contexts are evaluated:
    Black, Blur Noisy, Mean, Noisy, White, Unrelated Image,
    and Blurred Source.
    Insertion ($\uparrow$), $100-\mathrm{Deletion}$
    ($\uparrow$), PG ($\uparrow$), AUPR$_0$ ($\uparrow$),
    and AUPR$_1$ ($\uparrow$) are reported.
    }
  \label{fig:ablation_target_type}
  \vspace{-1mm}
\end{figure}

\begin{figure}[t]
  \centering
  \includegraphics[width=\textwidth]{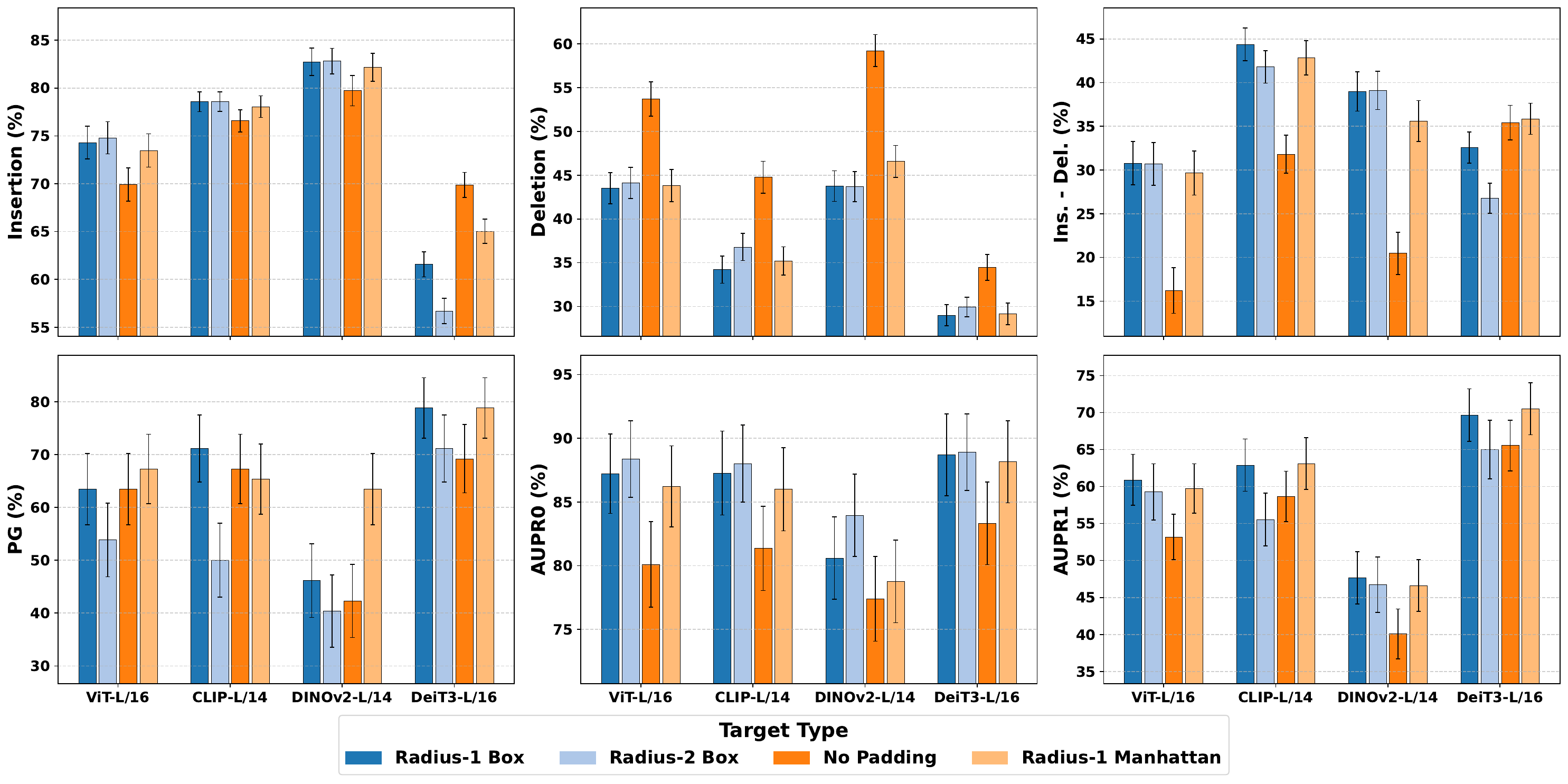}
  \vspace{-1mm}
  \caption{Selection operator ablation on ImageNet across four ViT backbones. The spatial support of the selection operator is varied by considering different neighborhood variants (No Padding, Radius-1 Box, Radius-2 Box, Radius-1 Manhattan), and faithfulness (Insertion, Deletion, Ins$-$Del) and localization (PG, AUPR$_1$, AUPR$_0$) are reported.}
  \label{fig:ablation_pad_type}
  \vspace{-1mm}
\end{figure}

\begin{figure}[t]
  \centering
  \includegraphics[width=\linewidth]{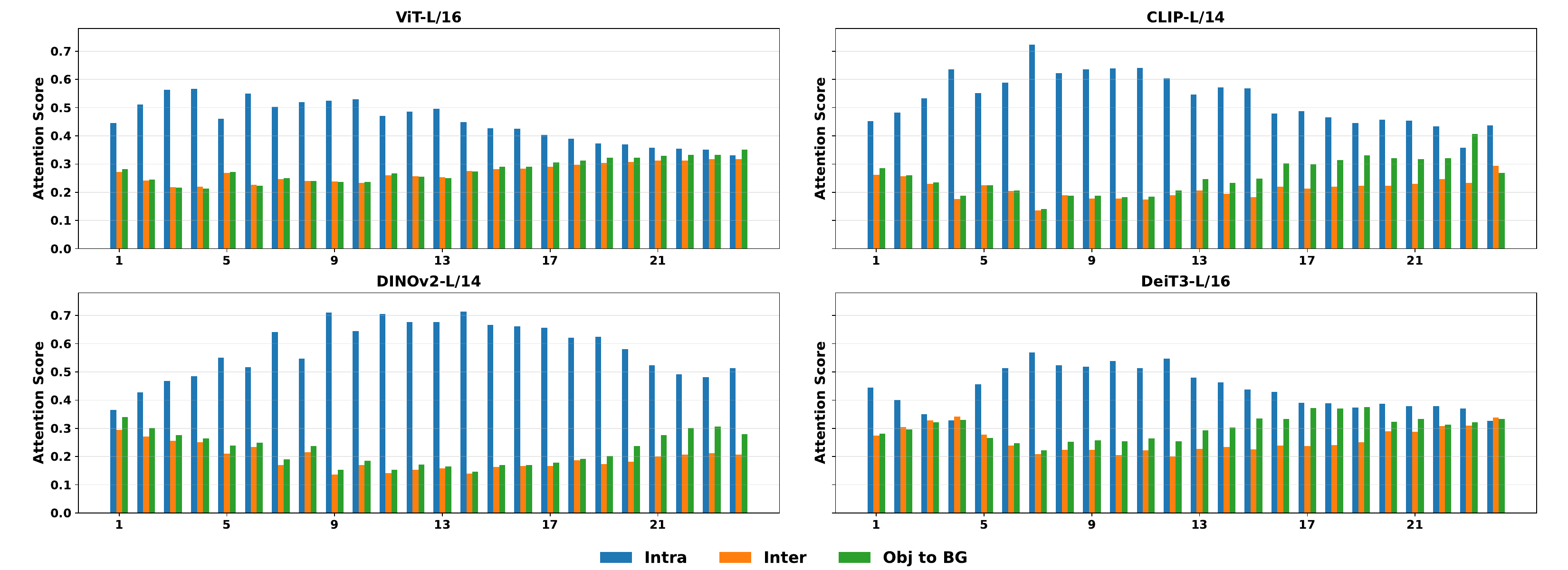}
  \vspace{-1mm}
    \caption{
    Mean attention weights across layers for different region pairs: intra-object (blue), inter-object (orange), and object-to-background (green). In early layers, intra-object attention increases and dominates, indicating strong object-level grouping.
    Around the middle layers, this gap reaches its peak. In later layers, intra-object attention decreases while object-to-background attention increases, reflecting growing extra-object contextualization and reduced spatial specificity.
    }
  \label{fig:attn-layer-sweep}
    \vspace{-1mm}
\end{figure}

\begin{figure}[t]
  \includegraphics[width=\linewidth]{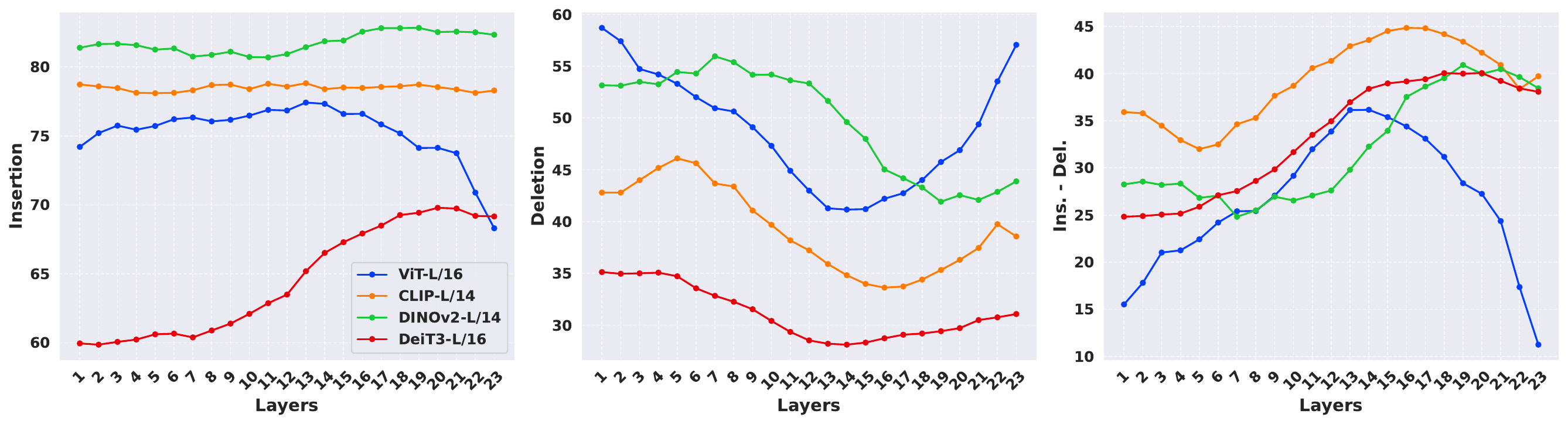}
  \vspace{-1mm}
  \caption{Intervention depth ablation on ImageNet using ViT backbones. For every cutoff point, we intervene on all transformer blocks from the initial layer up to that point and measure faithfulness metrics.}
  \label{fig:caap-layer-sweep}
  \vspace{-1mm}
\end{figure}

Here we extend the reported ablation results to all four studied ViT backbones (CLIP-L/14, ViT-L/16, DINOv2-L/14, and DeiT3-L/16).
Fig.~\ref{fig:ablation_target_type} shows that the choice of target context has generally a minor effect on faithfulness and localization metrics across architectures, with White blanks performing consistently
well. The Blurred Source shows weaker
localization, likely due to residual information from other patches leaking into the patched context.
Fig.~\ref{fig:ablation_pad_type} demonstrates that adding spatial padding to the selection operator consistently improves attribution quality compared to patching a single token, supporting the use of local neighborhoods during activation patching.
Fig.~\ref{fig:attn-layer-sweep} analyzes attention statistics across layers and shows that intra-object attention dominates in the intermediate layers before later transitioning toward extra-object contextualization.
Finally, Fig.~\ref{fig:caap-layer-sweep} presents cumulative layer-wise intervention sweeps across all backbones, showing that faithfulness peaks at intermediate depths, validating our default choice of intervening up to approximately the first two-thirds of transformer layers.

\subsection{Intervention Depth Across Datasets}
\label{app:cr_depth}

To complement the cross-backbone depth sweeps in
Fig.~\ref{fig:caap-layer-sweep}, we report additional
cumulative intervention-depth evaluations with
ViT-B/16 on Food-101, ImageNet-Hard, and Oxford-IIIT
Pet in Table~\ref{tab:cr_depth}. The results show
that intermediate intervention depths generally achieve
the strongest performance, further supporting our
choice of the default intervention range.

\input{tables/cross_dataset_depth}

\subsection{CNN Architectures}
\label{app:cnn}
\input{tables/ablation_cnn}

We further evaluate the intervention strategies on convolutional architectures, including ResNet-18, ResNet-50~\cite{resnet}, DenseNet-161~\cite{densenet}, and VGG-16~\cite{vgg}. The results are reported in Tables~\ref{tab:faithfulness_cnn_comparison} and~\ref{tab:segmentation_pg_cnn_comparison}. Compared with the transformer results reported earlier, the overall performance of intervention-based methods is lower across both faithfulness and localization metrics. In particular, the Insertion and Insertion$-$Deletion scores are substantially smaller than those observed for ViT models.
One possible explanation is that transformer architectures are generally more robust to distribution shifts than convolutional networks. Since intervention-based attribution methods modify the input or intermediate representations, they introduce out-of-distribution (OOD) patterns. Transformers appear to handle these perturbations more reliably, which makes intervention-based attribution more suitable for transformer-based models.

Within the CNN architectures, CAAP still achieves the best overall results in most cases. As shown in Table~\ref{tab:faithfulness_cnn_comparison}, CAAP produces the highest Insertion$-$Deletion scores across all four models and often improves the Insertion score, while maintaining competitive Deletion AUC. Localization metrics in Table~\ref{tab:segmentation_pg_cnn_comparison} show the same trend, where CAAP consistently achieves the highest AUPR$_1$ and AUPR$_0$ and improves PG in most cases.
Another difference from the transformer experiments is that the Input Deletion baseline generally outperforms Input Insertion on CNNs. This behavior suggests that removing pixels may better reflect the model’s reliance on local features in convolutional networks. In contrast, inserting isolated patches into a blank image provides limited evidence for the classifier, possibly due to the OOD nature of the input.

\section{Broader Impact}

This work contributes to interpretability and attribution methods for Vision Transformers by proposing a causal intervention framework based on activation patching. Improved attribution quality may help researchers better understand model behavior, identify reliance on spurious correlations, and support debugging and auditing of vision systems in sensitive applications. At the same time, attribution methods may be over-interpreted as fully faithful explanations of model reasoning, despite inherent limitations and architecture-specific assumptions. Incorrect or overly confident interpretations could lead to misplaced trust in deployed systems. Our method is designed as a diagnostic and research tool rather than a guarantee of model transparency or safety.

All experiments use publicly available datasets and pretrained models, including ImageNet-1K, ImageNet-S, Oxford-IIIT-Pet, Food-101, CLIP, DINOv2, DeiT3, and ViT models, under their respective research licenses and terms of use. The paper introduces a new attribution method but does not release new datasets or generative models.

\section{Experimental Sampling and Resources}
\label{app:resources}

For each dataset, we evaluate all methods on a fixed random subset of 500 samples. The same samples are used across baselines and ablations to ensure fair comparison. When segmentation annotations are required, samples are drawn only from images with valid object-level annotations.
All experiments were run on four NVIDIA RTX 4090 GPUs, with CPU resources used for preprocessing, metric computation, and result aggregation.

%% file: tables/base_imagenet.tex
\begin{table*}[t]
\caption{Faithfulness quantitative results on ImageNet for ViT-B/16~\cite{vit}, Clip-B/16~\cite{clip}, DINOv2-B/14~\cite{dino2}, and DeiT3-B/16~\cite{deit}. Deletion ($\downarrow$), Insertion ($\uparrow$), and Ins$-$Del ($\uparrow$) are reported. Best is \textbf{bold}, second best is \underline{underlined} in each column.}
\vspace{-1.5mm}
\label{tab:faithfulness_comparison_results_base}
\centering
\small
\setlength{\tabcolsep}{4pt}
\renewcommand{\arraystretch}{1.15}
\resizebox{\textwidth}{!}{%
\begin{tabular}{lcccccccccccc}
\toprule
& \multicolumn{3}{c}{ViT-B/16~\cite{vit}}
& \multicolumn{3}{c}{Clip-B/16~\cite{clip}}
& \multicolumn{3}{c}{DINOv2-B/14~\cite{dino2}}
& \multicolumn{3}{c}{DeiT3-B/16~\cite{deit}} \\
\cmidrule(lr){2-4}\cmidrule(lr){5-7}\cmidrule(lr){8-10}\cmidrule(lr){11-13}
Method
& Del$\downarrow$ & Ins$\uparrow$ & Ins$-$Del$\uparrow$
& Del$\downarrow$ & Ins$\uparrow$ & Ins$-$Del$\uparrow$
& Del$\downarrow$ & Ins$\uparrow$ & Ins$-$Del$\uparrow$
& Del$\downarrow$ & Ins$\uparrow$ & Ins$-$Del$\uparrow$ \\
\midrule

RISE~\cite{Petsiuk2018RISE}
& 42.65$_{\textcolor{gray}{\pm 1.51}}$ & \textbf{66.85}$_{\textcolor{gray}{\pm 1.17}}$ & 24.20$_{\textcolor{gray}{\pm 1.36}}$
& 56.22$_{\textcolor{gray}{\pm 1.12}}$ & 59.32$_{\textcolor{gray}{\pm 1.06}}$ & 3.10$_{\textcolor{gray}{\pm 0.85}}$
& 52.17$_{\textcolor{gray}{\pm 1.37}}$ & \underline{76.24}$_{\textcolor{gray}{\pm 1.00}}$ & 24.07$_{\textcolor{gray}{\pm 1.03}}$
& 43.45$_{\textcolor{gray}{\pm 1.22}}$ & 60.54$_{\textcolor{gray}{\pm 0.83}}$ & 17.09$_{\textcolor{gray}{\pm 0.90}}$ \\

Grad-CAM~\cite{grad_cam}
& 43.92$_{\textcolor{gray}{\pm 1.41}}$ & 59.07$_{\textcolor{gray}{\pm 1.25}}$ & 15.15$_{\textcolor{gray}{\pm 1.40}}$
& 55.21$_{\textcolor{gray}{\pm 1.21}}$ & 52.68$_{\textcolor{gray}{\pm 1.04}}$ & -2.53$_{\textcolor{gray}{\pm 1.30}}$
& 42.45$_{\textcolor{gray}{\pm 1.16}}$ & 74.98$_{\textcolor{gray}{\pm 1.13}}$ & 32.53$_{\textcolor{gray}{\pm 1.01}}$
& \underline{28.96}$_{\textcolor{gray}{\pm 0.89}}$ & 61.09$_{\textcolor{gray}{\pm 0.91}}$ & 32.13$_{\textcolor{gray}{\pm 0.86}}$ \\

Rollout~\cite{attn_r}
& 54.68$_{\textcolor{gray}{\pm 1.36}}$ & 49.83$_{\textcolor{gray}{\pm 1.28}}$ & -4.85$_{\textcolor{gray}{\pm 1.27}}$
& 51.65$_{\textcolor{gray}{\pm 1.12}}$ & 56.68$_{\textcolor{gray}{\pm 1.09}}$ & 5.03$_{\textcolor{gray}{\pm 1.16}}$
& 52.01$_{\textcolor{gray}{\pm 1.16}}$ & 71.06$_{\textcolor{gray}{\pm 1.18}}$ & 19.05$_{\textcolor{gray}{\pm 0.97}}$
& 46.78$_{\textcolor{gray}{\pm 1.03}}$ & 53.76$_{\textcolor{gray}{\pm 0.98}}$ & 6.98$_{\textcolor{gray}{\pm 0.94}}$ \\

T-Attr~\cite{chefer2021transformer}
& 39.49$_{\textcolor{gray}{\pm 1.32}}$ & 60.73$_{\textcolor{gray}{\pm 1.25}}$ & 21.24$_{\textcolor{gray}{\pm 1.26}}$
& 41.73$_{\textcolor{gray}{\pm 1.10}}$ & 65.18$_{\textcolor{gray}{\pm 1.04}}$ & 23.45$_{\textcolor{gray}{\pm 1.09}}$
& 45.09$_{\textcolor{gray}{\pm 1.10}}$ & 73.29$_{\textcolor{gray}{\pm 1.19}}$ & 28.20$_{\textcolor{gray}{\pm 1.02}}$
& 32.61$_{\textcolor{gray}{\pm 0.96}}$ & 62.75$_{\textcolor{gray}{\pm 0.98}}$ & 30.13$_{\textcolor{gray}{\pm 0.93}}$ \\

ViT-CX~\cite{vit_cx}
& \underline{33.94}$_{\textcolor{gray}{\pm 1.17}}$ & 62.41$_{\textcolor{gray}{\pm 1.35}}$ & \underline{28.47}$_{\textcolor{gray}{\pm 1.43}}$
& 40.78$_{\textcolor{gray}{\pm 1.21}}$ & 61.78$_{\textcolor{gray}{\pm 1.06}}$ & 21.00$_{\textcolor{gray}{\pm 1.44}}$
& 44.25$_{\textcolor{gray}{\pm 1.23}}$ & 74.66$_{\textcolor{gray}{\pm 1.13}}$ & 30.41$_{\textcolor{gray}{\pm 1.15}}$
& 31.64$_{\textcolor{gray}{\pm 1.00}}$ & 57.73$_{\textcolor{gray}{\pm 0.95}}$ & 26.10$_{\textcolor{gray}{\pm 1.10}}$ \\

TAM~\cite{tam}
& 41.81$_{\textcolor{gray}{\pm 1.34}}$ & 60.27$_{\textcolor{gray}{\pm 1.31}}$ & 18.46$_{\textcolor{gray}{\pm 1.36}}$
& 42.15$_{\textcolor{gray}{\pm 1.13}}$ & 65.05$_{\textcolor{gray}{\pm 1.02}}$ & 22.90$_{\textcolor{gray}{\pm 1.08}}$
& \underline{41.92}$_{\textcolor{gray}{\pm 1.15}}$ & 76.10$_{\textcolor{gray}{\pm 1.14}}$ & \underline{34.17}$_{\textcolor{gray}{\pm 1.04}}$
& 29.86$_{\textcolor{gray}{\pm 0.95}}$ & \underline{63.63}$_{\textcolor{gray}{\pm 0.94}}$ & 33.77$_{\textcolor{gray}{\pm 0.91}}$ \\

IxG+~\cite{mehri-skipplus-cvpr24}
& 50.41$_{\textcolor{gray}{\pm 1.53}}$ & 55.90$_{\textcolor{gray}{\pm 1.37}}$ & 5.49$_{\textcolor{gray}{\pm 1.37}}$
& 58.07$_{\textcolor{gray}{\pm 1.21}}$ & 60.46$_{\textcolor{gray}{\pm 1.11}}$ & 2.39$_{\textcolor{gray}{\pm 0.96}}$
& 61.41$_{\textcolor{gray}{\pm 1.28}}$ & 72.55$_{\textcolor{gray}{\pm 1.21}}$ & 11.14$_{\textcolor{gray}{\pm 0.77}}$
& 50.43$_{\textcolor{gray}{\pm 1.16}}$ & 60.96$_{\textcolor{gray}{\pm 1.07}}$ & 10.53$_{\textcolor{gray}{\pm 1.00}}$ \\

Libra IxG+~\cite{Mehri_2025_CVPR}
& 37.53$_{\textcolor{gray}{\pm 1.34}}$ & \underline{65.41}$_{\textcolor{gray}{\pm 1.28}}$ & 27.88$_{\textcolor{gray}{\pm 1.33}}$
& \underline{39.72}$_{\textcolor{gray}{\pm 1.10}}$ & \underline{67.34}$_{\textcolor{gray}{\pm 0.96}}$ & \underline{27.62}$_{\textcolor{gray}{\pm 1.06}}$
& 49.19$_{\textcolor{gray}{\pm 1.21}}$ & 73.53$_{\textcolor{gray}{\pm 1.21}}$ & 24.34$_{\textcolor{gray}{\pm 1.13}}$
& 31.00$_{\textcolor{gray}{\pm 0.98}}$ & \textbf{65.23}$_{\textcolor{gray}{\pm 0.91}}$ & \underline{34.22}$_{\textcolor{gray}{\pm 0.90}}$ \\

MDA~\cite{mda}
& 46.63$_{\textcolor{gray}{\pm 1.31}}$ & 56.16$_{\textcolor{gray}{\pm 1.35}}$ & 9.53$_{\textcolor{gray}{\pm 1.45}}$
& 40.70$_{\textcolor{gray}{\pm 1.10}}$ & 62.64$_{\textcolor{gray}{\pm 1.06}}$ & 21.94$_{\textcolor{gray}{\pm 1.26}}$
& 43.09$_{\textcolor{gray}{\pm 1.14}}$ & 72.81$_{\textcolor{gray}{\pm 1.17}}$ & 29.73$_{\textcolor{gray}{\pm 1.12}}$
& 38.86$_{\textcolor{gray}{\pm 0.97}}$ & 53.71$_{\textcolor{gray}{\pm 0.96}}$ & 14.85$_{\textcolor{gray}{\pm 1.04}}$ \\

MUTEX~\cite{mutex}
& 46.36$_{\textcolor{gray}{\pm 1.40}}$ & 55.79$_{\textcolor{gray}{\pm 1.27}}$ & 9.43$_{\textcolor{gray}{\pm 1.35}}$
& 52.71$_{\textcolor{gray}{\pm 1.31}}$ & 59.31$_{\textcolor{gray}{\pm 1.01}}$ & 6.60$_{\textcolor{gray}{\pm 1.22}}$
& 49.76$_{\textcolor{gray}{\pm 1.05}}$ & 70.30$_{\textcolor{gray}{\pm 1.24}}$ & 20.54$_{\textcolor{gray}{\pm 0.90}}$
& 48.91$_{\textcolor{gray}{\pm 1.23}}$ & 57.11$_{\textcolor{gray}{\pm 0.93}}$ & 8.20$_{\textcolor{gray}{\pm 1.14}}$ \\

\rowcolor{blue!10}
CAAP (Ours)
& \textbf{32.62}$_{\textcolor{gray}{\pm 1.15}}$ & 64.77$_{\textcolor{gray}{\pm 1.33}}$ & \textbf{32.15}$_{\textcolor{gray}{\pm 1.32}}$
& \textbf{32.09}$_{\textcolor{gray}{\pm 0.93}}$ & \textbf{67.40}$_{\textcolor{gray}{\pm 1.00}}$ & \textbf{35.32}$_{\textcolor{gray}{\pm 1.05}}$
& \textbf{35.42}$_{\textcolor{gray}{\pm 1.08}}$ & \textbf{76.81}$_{\textcolor{gray}{\pm 1.14}}$ & \textbf{41.38}$_{\textcolor{gray}{\pm 1.04}}$
& \textbf{26.48}$_{\textcolor{gray}{\pm 0.78}}$ & 61.23$_{\textcolor{gray}{\pm 0.91}}$ & \textbf{34.74}$_{\textcolor{gray}{\pm 0.79}}$ \\
\bottomrule
\end{tabular}
}
\end{table*}

\begin{table*}[t]
\caption{Localization quantitative results on ImageNet for ViT-B/16~\cite{vit}, Clip-B/16~\cite{clip}, DINOv2-B/14~\cite{dino2}, and DeiT3-B/16~\cite{deit}. AUPR$_1$ ($\uparrow$), AUPR$_0$ ($\uparrow$) and PG ($\uparrow$) are reported. Best is \textbf{bold}, second best is \underline{underlined} in each column.}
\vspace{-1.5mm}
\label{tab:localization_comparison_results_base}
\centering
\small
\setlength{\tabcolsep}{3.8pt}
\renewcommand{\arraystretch}{1.15}
\resizebox{\textwidth}{!}{%
\begin{tabular}{lcccccccccccc}
\toprule
& \multicolumn{3}{c}{ViT-B/16~\cite{vit}}
& \multicolumn{3}{c}{Clip-B/16~\cite{clip}}
& \multicolumn{3}{c}{DINOv2-B/14~\cite{dino2}}
& \multicolumn{3}{c}{DeiT3-B/16~\cite{deit}} \\
\cmidrule(lr){2-4}\cmidrule(lr){5-7}\cmidrule(lr){8-10}\cmidrule(lr){11-13}
Method
& AUPR$_1$ $\uparrow$ & AUPR$_0$ $\uparrow$ & PG $\uparrow$
& AUPR$_1$ $\uparrow$ & AUPR$_0$ $\uparrow$ & PG $\uparrow$
& AUPR$_1$ $\uparrow$ & AUPR$_0$ $\uparrow$ & PG $\uparrow$
& AUPR$_1$ $\uparrow$ & AUPR$_0$ $\uparrow$ & PG $\uparrow$ \\
\midrule

RISE~\cite{Petsiuk2018RISE}
& 47.82$_{\textcolor{gray}{\pm 2.14}}$ & 72.76$_{\textcolor{gray}{\pm 2.31}}$ & 53.91$_{\textcolor{gray}{\pm 4.41}}$
& 34.98$_{\textcolor{gray}{\pm 2.25}}$ & 69.84$_{\textcolor{gray}{\pm 2.15}}$ & 25.00$_{\textcolor{gray}{\pm 3.83}}$
& \underline{37.94}$_{\textcolor{gray}{\pm 2.18}}$ & 69.99$_{\textcolor{gray}{\pm 2.23}}$ & 28.91$_{\textcolor{gray}{\pm 4.01}}$
& 42.54$_{\textcolor{gray}{\pm 2.18}}$ & 70.54$_{\textcolor{gray}{\pm 2.30}}$ & 46.09$_{\textcolor{gray}{\pm 4.41}}$ \\

Grad-CAM~\cite{grad_cam}
& 46.29$_{\textcolor{gray}{\pm 2.37}}$ & 73.78$_{\textcolor{gray}{\pm 2.23}}$ & 45.31$_{\textcolor{gray}{\pm 4.40}}$
& 34.26$_{\textcolor{gray}{\pm 2.36}}$ & 66.35$_{\textcolor{gray}{\pm 2.26}}$ & 22.66$_{\textcolor{gray}{\pm 3.70}}$
& 37.07$_{\textcolor{gray}{\pm 2.31}}$ & \underline{72.09}$_{\textcolor{gray}{\pm 2.08}}$ & 46.09$_{\textcolor{gray}{\pm 4.41}}$
& \underline{66.84}$_{\textcolor{gray}{\pm 2.08}}$ & \underline{86.85}$_{\textcolor{gray}{\pm 1.78}}$ & \underline{66.41}$_{\textcolor{gray}{\pm 4.17}}$ \\

Rollout~\cite{attn_r}
& 31.62$_{\textcolor{gray}{\pm 1.83}}$ & 65.91$_{\textcolor{gray}{\pm 2.65}}$ & 14.84$_{\textcolor{gray}{\pm 3.14}}$
& 39.24$_{\textcolor{gray}{\pm 1.95}}$ & 68.69$_{\textcolor{gray}{\pm 2.66}}$ & 28.91$_{\textcolor{gray}{\pm 4.01}}$
& 33.34$_{\textcolor{gray}{\pm 2.05}}$ & 67.76$_{\textcolor{gray}{\pm 2.30}}$ & 36.72$_{\textcolor{gray}{\pm 4.26}}$
& 32.41$_{\textcolor{gray}{\pm 1.77}}$ & 67.86$_{\textcolor{gray}{\pm 2.64}}$ & 11.72$_{\textcolor{gray}{\pm 2.84}}$ \\

T-Attr~\cite{chefer2021transformer}
& 43.33$_{\textcolor{gray}{\pm 1.80}}$ & 74.48$_{\textcolor{gray}{\pm 2.43}}$ & 35.16$_{\textcolor{gray}{\pm 4.22}}$
& 54.08$_{\textcolor{gray}{\pm 1.84}}$ & \underline{78.80}$_{\textcolor{gray}{\pm 2.25}}$ & \underline{52.34}$_{\textcolor{gray}{\pm 4.41}}$
& 37.08$_{\textcolor{gray}{\pm 2.21}}$ & 71.10$_{\textcolor{gray}{\pm 2.20}}$ & 42.19$_{\textcolor{gray}{\pm 4.37}}$
& 60.34$_{\textcolor{gray}{\pm 1.75}}$ & 82.23$_{\textcolor{gray}{\pm 1.98}}$ & 53.12$_{\textcolor{gray}{\pm 4.41}}$ \\

ViT-CX~\cite{vit_cx}
& \textbf{62.87}$_{\textcolor{gray}{\pm 2.13}}$ & \underline{82.34}$_{\textcolor{gray}{\pm 2.09}}$ & \textbf{67.19}$_{\textcolor{gray}{\pm 4.15}}$
& \underline{54.13}$_{\textcolor{gray}{\pm 2.42}}$ & 74.43$_{\textcolor{gray}{\pm 2.48}}$ & 59.38$_{\textcolor{gray}{\pm 4.34}}$
& 36.43$_{\textcolor{gray}{\pm 2.25}}$ & 70.02$_{\textcolor{gray}{\pm 2.18}}$ & \underline{51.56}$_{\textcolor{gray}{\pm 4.42}}$
& 62.86$_{\textcolor{gray}{\pm 2.13}}$ & 78.92$_{\textcolor{gray}{\pm 2.48}}$ & 65.62$_{\textcolor{gray}{\pm 4.20}}$ \\

TAM~\cite{tam}
& 47.52$_{\textcolor{gray}{\pm 1.85}}$ & 76.79$_{\textcolor{gray}{\pm 2.33}}$ & 28.12$_{\textcolor{gray}{\pm 3.97}}$
& 53.00$_{\textcolor{gray}{\pm 2.01}}$ & 77.60$_{\textcolor{gray}{\pm 2.25}}$ & 50.00$_{\textcolor{gray}{\pm 4.42}}$
& 35.85$_{\textcolor{gray}{\pm 2.14}}$ & 69.98$_{\textcolor{gray}{\pm 2.25}}$ & 32.03$_{\textcolor{gray}{\pm 4.12}}$
& 62.72$_{\textcolor{gray}{\pm 1.79}}$ & 84.27$_{\textcolor{gray}{\pm 1.97}}$ & 56.25$_{\textcolor{gray}{\pm 4.38}}$ \\

IxG+~\cite{mehri-skipplus-cvpr24}
& 41.26$_{\textcolor{gray}{\pm 2.00}}$ & 67.79$_{\textcolor{gray}{\pm 2.20}}$ & 42.19$_{\textcolor{gray}{\pm 4.37}}$
& 41.49$_{\textcolor{gray}{\pm 1.94}}$ & 67.07$_{\textcolor{gray}{\pm 2.17}}$ & 46.88$_{\textcolor{gray}{\pm 4.41}}$
& 34.15$_{\textcolor{gray}{\pm 2.15}}$ & 67.99$_{\textcolor{gray}{\pm 2.16}}$ & 40.62$_{\textcolor{gray}{\pm 4.34}}$
& 38.27$_{\textcolor{gray}{\pm 1.87}}$ & 70.59$_{\textcolor{gray}{\pm 2.18}}$ & 13.28$_{\textcolor{gray}{\pm 3.00}}$ \\

Libra IxG+~\cite{Mehri_2025_CVPR}
& 46.63$_{\textcolor{gray}{\pm 1.97}}$ & 73.36$_{\textcolor{gray}{\pm 2.21}}$ & 46.88$_{\textcolor{gray}{\pm 4.41}}$
& 51.27$_{\textcolor{gray}{\pm 1.92}}$ & 77.33$_{\textcolor{gray}{\pm 2.14}}$ & 38.28$_{\textcolor{gray}{\pm 4.30}}$
& 34.45$_{\textcolor{gray}{\pm 2.14}}$ & 69.43$_{\textcolor{gray}{\pm 2.22}}$ & 27.34$_{\textcolor{gray}{\pm 3.94}}$
& 54.16$_{\textcolor{gray}{\pm 1.75}}$ & 79.75$_{\textcolor{gray}{\pm 2.20}}$ & 45.31$_{\textcolor{gray}{\pm 4.40}}$ \\

MDA~\cite{mda}
& 35.26$_{\textcolor{gray}{\pm 1.84}}$ & 67.56$_{\textcolor{gray}{\pm 2.69}}$ & 17.19$_{\textcolor{gray}{\pm 3.33}}$
& 46.90$_{\textcolor{gray}{\pm 2.01}}$ & 72.86$_{\textcolor{gray}{\pm 2.65}}$ & 33.07$_{\textcolor{gray}{\pm 4.17}}$
& 33.88$_{\textcolor{gray}{\pm 2.04}}$ & 68.75$_{\textcolor{gray}{\pm 2.30}}$ & 25.20$_{\textcolor{gray}{\pm 3.85}}$
& 37.37$_{\textcolor{gray}{\pm 1.77}}$ & 69.49$_{\textcolor{gray}{\pm 2.70}}$ & 17.19$_{\textcolor{gray}{\pm 3.33}}$ \\

MUTEX~\cite{mutex}
& 40.56$_{\textcolor{gray}{\pm 2.22}}$ & 71.89$_{\textcolor{gray}{\pm 2.38}}$ & 32.81$_{\textcolor{gray}{\pm 4.15}}$
& 39.13$_{\textcolor{gray}{\pm 2.20}}$ & 69.50$_{\textcolor{gray}{\pm 2.32}}$ & 32.03$_{\textcolor{gray}{\pm 4.12}}$
& 34.73$_{\textcolor{gray}{\pm 2.23}}$ & 70.28$_{\textcolor{gray}{\pm 2.16}}$ & 27.34$_{\textcolor{gray}{\pm 3.94}}$
& 34.76$_{\textcolor{gray}{\pm 2.19}}$ & 67.43$_{\textcolor{gray}{\pm 2.33}}$ & 25.78$_{\textcolor{gray}{\pm 3.87}}$ \\

\rowcolor{blue!10}
CAAP (Ours)
& \underline{61.88}$_{\textcolor{gray}{\pm 2.07}}$ & \textbf{82.93}$_{\textcolor{gray}{\pm 1.90}}$ & \underline{65.63}$_{\textcolor{gray}{\pm 4.20}}$
& \textbf{63.96}$_{\textcolor{gray}{\pm 1.91}}$ & \textbf{83.60}$_{\textcolor{gray}{\pm 1.93}}$ & \textbf{71.88}$_{\textcolor{gray}{\pm 3.97}}$
& \textbf{54.63}$_{\textcolor{gray}{\pm 1.94}}$ & \textbf{76.88}$_{\textcolor{gray}{\pm 2.19}}$ & \textbf{57.03}$_{\textcolor{gray}{\pm 4.38}}$
& \textbf{70.26}$_{\textcolor{gray}{\pm 1.88}}$ & \textbf{86.88}$_{\textcolor{gray}{\pm 1.77}}$ & \textbf{87.50}$_{\textcolor{gray}{\pm 2.92}}$ \\

\bottomrule
\end{tabular}
}
\end{table*}

%% file: tables/small_imagenet.tex
\begin{table*}[t]
\caption{Faithfulness quantitative results on ImageNet for ViT-S/16~\cite{vit}, DINOv2-S/14~\cite{dino2}, and DeiT3-S/16~\cite{deit}. Deletion ($\downarrow$), Insertion ($\uparrow$), and Ins$-$Del ($\uparrow$) are reported. Best is \textbf{bold}, second best is \underline{underlined} in each column.}
\vspace{-1.5mm}
\label{tab:faithfulness_comparison_results_small}
\centering
\small
\setlength{\tabcolsep}{4pt}
\renewcommand{\arraystretch}{1.15}
\resizebox{\textwidth}{!}{%
\begin{tabular}{lccccccccc}
\toprule
& \multicolumn{3}{c}{ViT-S/16~\cite{vit}}
& \multicolumn{3}{c}{DINOv2-S/14~\cite{dino2}}
& \multicolumn{3}{c}{DeiT3-S/16~\cite{deit}} \\
\cmidrule(lr){2-4}\cmidrule(lr){5-7}\cmidrule(lr){8-10}
Method
& Del$\downarrow$ & Ins$\uparrow$ & Ins$-$Del$\uparrow$
& Del$\downarrow$ & Ins$\uparrow$ & Ins$-$Del$\uparrow$
& Del$\downarrow$ & Ins$\uparrow$ & Ins$-$Del$\uparrow$ \\
\midrule

RISE~\cite{Petsiuk2018RISE} & 35.38$_{\textcolor{gray}{\pm 1.31}}$ & \underline{63.68}$_{\textcolor{gray}{\pm 1.12}}$ & 28.29$_{\textcolor{gray}{\pm 1.19}}$ & 39.25$_{\textcolor{gray}{\pm 1.30}}$ & \textbf{71.03}$_{\textcolor{gray}{\pm 1.02}}$ & 31.78$_{\textcolor{gray}{\pm 1.05}}$ & 39.65$_{\textcolor{gray}{\pm 1.45}}$ & \underline{68.01}$_{\textcolor{gray}{\pm 1.21}}$ & 28.36$_{\textcolor{gray}{\pm 1.23}}$ \\
Grad-CAM~\cite{grad_cam} & 30.70$_{\textcolor{gray}{\pm 1.07}}$ & 61.41$_{\textcolor{gray}{\pm 1.22}}$ & 30.71$_{\textcolor{gray}{\pm 1.11}}$ & \underline{30.94}$_{\textcolor{gray}{\pm 0.99}}$ & 68.28$_{\textcolor{gray}{\pm 1.15}}$ & 37.35$_{\textcolor{gray}{\pm 1.02}}$ & \underline{27.35}$_{\textcolor{gray}{\pm 1.04}}$ & 67.17$_{\textcolor{gray}{\pm 1.32}}$ & \underline{39.83}$_{\textcolor{gray}{\pm 1.17}}$ \\
Rollout~\cite{attn_r} & 43.36$_{\textcolor{gray}{\pm 1.16}}$ & 50.29$_{\textcolor{gray}{\pm 1.26}}$ & 6.93$_{\textcolor{gray}{\pm 1.15}}$ & 40.01$_{\textcolor{gray}{\pm 1.09}}$ & 64.05$_{\textcolor{gray}{\pm 1.20}}$ & 24.04$_{\textcolor{gray}{\pm 1.05}}$ & 51.77$_{\textcolor{gray}{\pm 1.33}}$ & 52.32$_{\textcolor{gray}{\pm 1.29}}$ & 0.54$_{\textcolor{gray}{\pm 1.12}}$ \\
T-Attr~\cite{chefer2021transformer} & 30.23$_{\textcolor{gray}{\pm 1.04}}$ & 60.06$_{\textcolor{gray}{\pm 1.24}}$ & 29.83$_{\textcolor{gray}{\pm 1.10}}$ & 34.39$_{\textcolor{gray}{\pm 1.01}}$ & 66.21$_{\textcolor{gray}{\pm 1.20}}$ & 31.82$_{\textcolor{gray}{\pm 1.04}}$ & 33.29$_{\textcolor{gray}{\pm 1.18}}$ & 65.70$_{\textcolor{gray}{\pm 1.34}}$ & 32.41$_{\textcolor{gray}{\pm 1.14}}$ \\
ViT-CX~\cite{vit_cx} & \underline{28.84}$_{\textcolor{gray}{\pm 1.01}}$ & 61.42$_{\textcolor{gray}{\pm 1.27}}$ & \underline{32.58}$_{\textcolor{gray}{\pm 1.22}}$ & 33.44$_{\textcolor{gray}{\pm 1.09}}$ & 66.93$_{\textcolor{gray}{\pm 1.17}}$ & 33.49$_{\textcolor{gray}{\pm 1.17}}$ & 33.98$_{\textcolor{gray}{\pm 1.22}}$ & 63.77$_{\textcolor{gray}{\pm 1.38}}$ & 29.79$_{\textcolor{gray}{\pm 1.36}}$ \\
TAM~\cite{tam} & 30.11$_{\textcolor{gray}{\pm 1.06}}$ & 61.52$_{\textcolor{gray}{\pm 1.26}}$ & 31.41$_{\textcolor{gray}{\pm 1.13}}$ & 31.74$_{\textcolor{gray}{\pm 1.04}}$ & 69.20$_{\textcolor{gray}{\pm 1.16}}$ & \underline{37.45}$_{\textcolor{gray}{\pm 1.06}}$ & 31.72$_{\textcolor{gray}{\pm 1.18}}$ & 66.55$_{\textcolor{gray}{\pm 1.33}}$ & 34.82$_{\textcolor{gray}{\pm 1.19}}$ \\
IxG+~\cite{mehri-skipplus-cvpr24} & 45.73$_{\textcolor{gray}{\pm 1.34}}$ & 51.75$_{\textcolor{gray}{\pm 1.28}}$ & 6.01$_{\textcolor{gray}{\pm 1.01}}$ & 50.67$_{\textcolor{gray}{\pm 1.26}}$ & 61.95$_{\textcolor{gray}{\pm 1.29}}$ & 11.28$_{\textcolor{gray}{\pm 0.89}}$ & 50.63$_{\textcolor{gray}{\pm 1.42}}$ & 59.43$_{\textcolor{gray}{\pm 1.39}}$ & 8.81$_{\textcolor{gray}{\pm 1.05}}$ \\
Libra IxG+~\cite{Mehri_2025_CVPR} & 32.70$_{\textcolor{gray}{\pm 1.11}}$ & 62.25$_{\textcolor{gray}{\pm 1.26}}$ & 29.55$_{\textcolor{gray}{\pm 1.19}}$ & 36.94$_{\textcolor{gray}{\pm 1.13}}$ & 66.86$_{\textcolor{gray}{\pm 1.22}}$ & 29.92$_{\textcolor{gray}{\pm 1.20}}$ & 35.72$_{\textcolor{gray}{\pm 1.31}}$ & 66.88$_{\textcolor{gray}{\pm 1.35}}$ & 31.17$_{\textcolor{gray}{\pm 1.18}}$ \\
MDA~\cite{mda} & 38.09$_{\textcolor{gray}{\pm 1.14}}$ & 54.93$_{\textcolor{gray}{\pm 1.27}}$ & 16.84$_{\textcolor{gray}{\pm 1.26}}$ & 34.53$_{\textcolor{gray}{\pm 1.06}}$ & 66.46$_{\textcolor{gray}{\pm 1.19}}$ & 31.93$_{\textcolor{gray}{\pm 1.16}}$ & 40.97$_{\textcolor{gray}{\pm 1.27}}$ & 57.86$_{\textcolor{gray}{\pm 1.33}}$ & 16.89$_{\textcolor{gray}{\pm 1.34}}$ \\
MUTEX~\cite{mutex} & 38.46$_{\textcolor{gray}{\pm 1.19}}$ & 55.48$_{\textcolor{gray}{\pm 1.22}}$ & 17.02$_{\textcolor{gray}{\pm 1.07}}$ & 42.51$_{\textcolor{gray}{\pm 0.98}}$ & 60.55$_{\textcolor{gray}{\pm 1.24}}$ & 18.04$_{\textcolor{gray}{\pm 0.93}}$ & 43.21$_{\textcolor{gray}{\pm 1.34}}$ & 59.95$_{\textcolor{gray}{\pm 1.28}}$ & 16.74$_{\textcolor{gray}{\pm 1.10}}$ \\
\rowcolor{blue!10}
CAAP (Ours) & \textbf{27.25}$_{\textcolor{gray}{\pm 0.98}}$ & \textbf{64.19}$_{\textcolor{gray}{\pm 1.25}}$ & \textbf{36.94}$_{\textcolor{gray}{\pm 1.13}}$ & \textbf{29.14}$_{\textcolor{gray}{\pm 0.96}}$ & \underline{69.90}$_{\textcolor{gray}{\pm 1.16}}$ & \textbf{40.76}$_{\textcolor{gray}{\pm 1.03}}$ & \textbf{26.66}$_{\textcolor{gray}{\pm 1.01}}$ & \textbf{68.23}$_{\textcolor{gray}{\pm 0.91}}$ & \textbf{41.57}$_{\textcolor{gray}{\pm 0.79}}$ \\

\bottomrule
\end{tabular}
}
\end{table*}

\begin{table*}[t]
\caption{Localization quantitative results on ImageNet for ViT-S/16~\cite{vit}, DINOv2-S/14~\cite{dino2}, and DeiT3-S/16~\cite{deit}. AUPR$_1$ ($\uparrow$), AUPR$_0$ ($\uparrow$) and PG ($\uparrow$) are reported. Best is \textbf{bold}, second best is \underline{underlined} in each column.}
\vspace{-1.5mm}
\label{tab:localization_comparison_results_small}
\centering
\small
\setlength{\tabcolsep}{3.8pt}
\renewcommand{\arraystretch}{1.15}
\resizebox{\textwidth}{!}{%
\begin{tabular}{lccccccccc}
\toprule
& \multicolumn{3}{c}{ViT-S/16~\cite{vit}}
& \multicolumn{3}{c}{DINOv2-S/14~\cite{dino2}}
& \multicolumn{3}{c}{DeiT3-S/16~\cite{deit}} \\
\cmidrule(lr){2-4}\cmidrule(lr){5-7}\cmidrule(lr){8-10}
Method
& AUPR$_1$ $\uparrow$ & AUPR$_0$ $\uparrow$ & PG $\uparrow$
& AUPR$_1$ $\uparrow$ & AUPR$_0$ $\uparrow$ & PG $\uparrow$
& AUPR$_1$ $\uparrow$ & AUPR$_0$ $\uparrow$ & PG $\uparrow$ \\
\midrule

RISE~\cite{Petsiuk2018RISE} & 47.43$_{\textcolor{gray}{\pm 2.05}}$ & 72.46$_{\textcolor{gray}{\pm 2.30}}$ & 50.78$_{\textcolor{gray}{\pm 4.42}}$ & 42.35$_{\textcolor{gray}{\pm 2.07}}$ & 70.93$_{\textcolor{gray}{\pm 2.36}}$ & 41.41$_{\textcolor{gray}{\pm 4.35}}$ & 47.43$_{\textcolor{gray}{\pm 2.15}}$ & 72.66$_{\textcolor{gray}{\pm 2.31}}$ & 49.22$_{\textcolor{gray}{\pm 4.42}}$ \\
Grad-CAM~\cite{grad_cam} & 54.18$_{\textcolor{gray}{\pm 2.08}}$ & 79.60$_{\textcolor{gray}{\pm 2.23}}$ & 50.00$_{\textcolor{gray}{\pm 4.42}}$ & 53.34$_{\textcolor{gray}{\pm 2.20}}$ & \textbf{81.38}$_{\textcolor{gray}{\pm 2.05}}$ & 50.78$_{\textcolor{gray}{\pm 4.42}}$ & \textbf{61.08}$_{\textcolor{gray}{\pm 2.00}}$ & \underline{81.44}$_{\textcolor{gray}{\pm 2.09}}$ & 59.38$_{\textcolor{gray}{\pm 4.34}}$ \\
Rollout~\cite{attn_r} & 34.28$_{\textcolor{gray}{\pm 1.91}}$ & 66.55$_{\textcolor{gray}{\pm 2.67}}$ & 17.19$_{\textcolor{gray}{\pm 3.33}}$ & 38.69$_{\textcolor{gray}{\pm 1.86}}$ & 69.65$_{\textcolor{gray}{\pm 2.54}}$ & 42.19$_{\textcolor{gray}{\pm 4.37}}$ & 34.06$_{\textcolor{gray}{\pm 2.00}}$ & 66.57$_{\textcolor{gray}{\pm 2.48}}$ & 24.22$_{\textcolor{gray}{\pm 3.79}}$ \\
T-Attr~\cite{chefer2021transformer} & 44.47$_{\textcolor{gray}{\pm 1.68}}$ & 76.60$_{\textcolor{gray}{\pm 2.39}}$ & 18.75$_{\textcolor{gray}{\pm 3.45}}$ & 49.48$_{\textcolor{gray}{\pm 2.07}}$ & 76.20$_{\textcolor{gray}{\pm 2.31}}$ & 43.75$_{\textcolor{gray}{\pm 4.38}}$ & 55.54$_{\textcolor{gray}{\pm 1.85}}$ & 75.75$_{\textcolor{gray}{\pm 2.19}}$ & \underline{62.50}$_{\textcolor{gray}{\pm 4.28}}$ \\
ViT-CX~\cite{vit_cx} & \underline{59.01}$_{\textcolor{gray}{\pm 2.16}}$ & \underline{80.91}$_{\textcolor{gray}{\pm 2.24}}$ & \underline{53.91}$_{\textcolor{gray}{\pm 4.41}}$ & \textbf{53.98}$_{\textcolor{gray}{\pm 2.16}}$ & 75.50$_{\textcolor{gray}{\pm 2.53}}$ & \textbf{59.38}$_{\textcolor{gray}{\pm 4.34}}$ & \underline{56.21}$_{\textcolor{gray}{\pm 2.16}}$ & 78.51$_{\textcolor{gray}{\pm 2.28}}$ & 54.69$_{\textcolor{gray}{\pm 4.40}}$ \\
TAM~\cite{tam} & 52.79$_{\textcolor{gray}{\pm 1.83}}$ & 77.93$_{\textcolor{gray}{\pm 2.25}}$ & 46.88$_{\textcolor{gray}{\pm 4.41}}$ & 48.80$_{\textcolor{gray}{\pm 1.89}}$ & 76.01$_{\textcolor{gray}{\pm 2.36}}$ & 43.75$_{\textcolor{gray}{\pm 4.38}}$ & 54.73$_{\textcolor{gray}{\pm 1.88}}$ & 78.08$_{\textcolor{gray}{\pm 2.17}}$ & 51.56$_{\textcolor{gray}{\pm 4.42}}$ \\
IxG+~\cite{mehri-skipplus-cvpr24} & 38.25$_{\textcolor{gray}{\pm 1.95}}$ & 65.73$_{\textcolor{gray}{\pm 2.10}}$ & 48.44$_{\textcolor{gray}{\pm 4.42}}$ & 38.62$_{\textcolor{gray}{\pm 1.97}}$ & 66.87$_{\textcolor{gray}{\pm 2.10}}$ & 43.75$_{\textcolor{gray}{\pm 4.38}}$ & 37.84$_{\textcolor{gray}{\pm 2.04}}$ & 67.48$_{\textcolor{gray}{\pm 2.12}}$ & 42.97$_{\textcolor{gray}{\pm 4.38}}$ \\
Libra IxG+~\cite{Mehri_2025_CVPR} & 44.69$_{\textcolor{gray}{\pm 1.92}}$ & 72.95$_{\textcolor{gray}{\pm 2.28}}$ & 33.59$_{\textcolor{gray}{\pm 4.17}}$ & 44.52$_{\textcolor{gray}{\pm 2.04}}$ & 73.13$_{\textcolor{gray}{\pm 2.29}}$ & 42.19$_{\textcolor{gray}{\pm 4.37}}$ & 48.78$_{\textcolor{gray}{\pm 1.86}}$ & 73.75$_{\textcolor{gray}{\pm 2.23}}$ & 47.66$_{\textcolor{gray}{\pm 4.41}}$ \\
MDA~\cite{mda} & 36.10$_{\textcolor{gray}{\pm 1.85}}$ & 67.55$_{\textcolor{gray}{\pm 2.72}}$ & 19.69$_{\textcolor{gray}{\pm 3.53}}$ & 44.00$_{\textcolor{gray}{\pm 2.04}}$ & 71.70$_{\textcolor{gray}{\pm 2.74}}$ & 37.60$_{\textcolor{gray}{\pm 4.33}}$ & 38.08$_{\textcolor{gray}{\pm 2.07}}$ & 69.09$_{\textcolor{gray}{\pm 2.62}}$ & 21.88$_{\textcolor{gray}{\pm 3.65}}$ \\
MUTEX~\cite{mutex} & 41.61$_{\textcolor{gray}{\pm 2.17}}$ & 71.82$_{\textcolor{gray}{\pm 2.39}}$ & 39.06$_{\textcolor{gray}{\pm 4.31}}$ & 37.63$_{\textcolor{gray}{\pm 2.25}}$ & 73.12$_{\textcolor{gray}{\pm 2.29}}$ & 25.00$_{\textcolor{gray}{\pm 3.83}}$ & 37.79$_{\textcolor{gray}{\pm 2.06}}$ & 71.01$_{\textcolor{gray}{\pm 2.38}}$ & 28.91$_{\textcolor{gray}{\pm 4.01}}$ \\
\rowcolor{blue!10}
CAAP (Ours) & \textbf{61.34}$_{\textcolor{gray}{\pm 1.99}}$ & \textbf{82.69}$_{\textcolor{gray}{\pm 1.99}}$ & \textbf{67.97}$_{\textcolor{gray}{\pm 4.12}}$ & \underline{53.49}$_{\textcolor{gray}{\pm 1.94}}$ & \underline{76.23}$_{\textcolor{gray}{\pm 2.20}}$ & \underline{57.81}$_{\textcolor{gray}{\pm 4.37}}$ & \textbf{61.08}$_{\textcolor{gray}{\pm 2.07}}$ & \textbf{81.77}$_{\textcolor{gray}{\pm 2.00}}$ & \textbf{65.63}$_{\textcolor{gray}{\pm 4.20}}$ \\

\bottomrule
\end{tabular}
}
\end{table*}

%% file: tables/huge_imagenet.tex
\begin{table*}[t]
\caption{Faithfulness quantitative results on ImageNet for CLIP-H/14~\cite{clip}, DINOv2-G/14~\cite{dino2}, and DeiT3-H/14~\cite{deit}. Deletion ($\downarrow$), Insertion ($\uparrow$), and Ins$-$Del ($\uparrow$) are reported. Best is \textbf{bold}, second best is \underline{underlined} in each column.}
\vspace{-1.5mm}
\label{tab:faithfulness_comparison_results_largest}
\centering
\small
\setlength{\tabcolsep}{4pt}
\renewcommand{\arraystretch}{1.15}
\resizebox{\textwidth}{!}{%
\begin{tabular}{lccccccccc}
\toprule
& \multicolumn{3}{c}{CLIP-H/14~\cite{clip}}
& \multicolumn{3}{c}{DINOv2-G/14~\cite{dino2}}
& \multicolumn{3}{c}{DeiT3-H/14~\cite{deit}} \\
\cmidrule(lr){2-4}\cmidrule(lr){5-7}\cmidrule(lr){8-10}
Method
& Del$\downarrow$ & Ins$\uparrow$ & Ins$-$Del$\uparrow$
& Del$\downarrow$ & Ins$\uparrow$ & Ins$-$Del$\uparrow$
& Del$\downarrow$ & Ins$\uparrow$ & Ins$-$Del$\uparrow$ \\
\midrule

RISE~\cite{Petsiuk2018RISE}
& 45.28$_{\textcolor{gray}{\pm 1.74}}$ & \textbf{63.61}$_{\textcolor{gray}{\pm 0.93}}$ & 18.33$_{\textcolor{gray}{\pm 1.58}}$
& 64.41$_{\textcolor{gray}{\pm 2.81}}$ & 81.11$_{\textcolor{gray}{\pm 1.88}}$ & 16.70$_{\textcolor{gray}{\pm 1.59}}$
& 63.56$_{\textcolor{gray}{\pm 2.34}}$ & 55.43$_{\textcolor{gray}{\pm 2.25}}$ & -8.13$_{\textcolor{gray}{\pm 1.91}}$ \\

Grad-CAM~\cite{grad_cam}
& 43.90$_{\textcolor{gray}{\pm 1.63}}$ & 54.91$_{\textcolor{gray}{\pm 1.28}}$ & 11.01$_{\textcolor{gray}{\pm 1.64}}$
& 58.01$_{\textcolor{gray}{\pm 1.86}}$ & 79.67$_{\textcolor{gray}{\pm 1.67}}$ & 21.65$_{\textcolor{gray}{\pm 1.54}}$
& \underline{35.53}$_{\textcolor{gray}{\pm 1.58}}$ & 64.04$_{\textcolor{gray}{\pm 1.43}}$ & 28.51$_{\textcolor{gray}{\pm 1.31}}$ \\

Rollout~\cite{attn_r}
& 44.64$_{\textcolor{gray}{\pm 1.46}}$ & 54.02$_{\textcolor{gray}{\pm 1.35}}$ & 9.37$_{\textcolor{gray}{\pm 1.45}}$
& 69.34$_{\textcolor{gray}{\pm 1.79}}$ & 76.50$_{\textcolor{gray}{\pm 1.76}}$ & 7.16$_{\textcolor{gray}{\pm 1.10}}$
& 63.67$_{\textcolor{gray}{\pm 1.48}}$ & 52.83$_{\textcolor{gray}{\pm 1.62}}$ & -10.84$_{\textcolor{gray}{\pm 1.24}}$ \\

T-Attr~\cite{chefer2021transformer}
& 41.07$_{\textcolor{gray}{\pm 1.58}}$ & 57.55$_{\textcolor{gray}{\pm 1.27}}$ & 16.48$_{\textcolor{gray}{\pm 1.58}}$
& \underline{53.01}$_{\textcolor{gray}{\pm 1.99}}$ & 81.48$_{\textcolor{gray}{\pm 1.55}}$ & 28.47$_{\textcolor{gray}{\pm 1.64}}$
& 40.30$_{\textcolor{gray}{\pm 1.70}}$ & 67.09$_{\textcolor{gray}{\pm 1.45}}$ & 26.78$_{\textcolor{gray}{\pm 1.51}}$ \\

CausalX~\cite{vit_cx}
& 39.61$_{\textcolor{gray}{\pm 1.67}}$ & 56.52$_{\textcolor{gray}{\pm 1.34}}$ & 16.91$_{\textcolor{gray}{\pm 2.00}}$
& 56.65$_{\textcolor{gray}{\pm 2.07}}$ & 81.24$_{\textcolor{gray}{\pm 1.55}}$ & 24.59$_{\textcolor{gray}{\pm 1.74}}$
& 41.69$_{\textcolor{gray}{\pm 1.82}}$ & 60.09$_{\textcolor{gray}{\pm 1.51}}$ & 18.40$_{\textcolor{gray}{\pm 1.80}}$ \\

TAM~\cite{tam}
& 43.26$_{\textcolor{gray}{\pm 1.66}}$ & 58.45$_{\textcolor{gray}{\pm 1.28}}$ & 15.19$_{\textcolor{gray}{\pm 1.62}}$
& 58.32$_{\textcolor{gray}{\pm 1.93}}$ & 79.47$_{\textcolor{gray}{\pm 1.67}}$ & 21.15$_{\textcolor{gray}{\pm 1.55}}$
& 40.66$_{\textcolor{gray}{\pm 1.83}}$ & \underline{67.74}$_{\textcolor{gray}{\pm 1.43}}$ & 27.09$_{\textcolor{gray}{\pm 1.54}}$ \\

IxG+~\cite{mehri-skipplus-cvpr24}
& 50.64$_{\textcolor{gray}{\pm 1.59}}$ & 54.10$_{\textcolor{gray}{\pm 1.21}}$ & 3.46$_{\textcolor{gray}{\pm 1.25}}$
& 69.22$_{\textcolor{gray}{\pm 1.93}}$ & 80.90$_{\textcolor{gray}{\pm 1.64}}$ & 11.68$_{\textcolor{gray}{\pm 1.16}}$
& 59.78$_{\textcolor{gray}{\pm 1.80}}$ & 65.12$_{\textcolor{gray}{\pm 1.67}}$ & 5.34$_{\textcolor{gray}{\pm 1.47}}$ \\

Libra IxG+~\cite{Mehri_2025_CVPR}
& \underline{37.75}$_{\textcolor{gray}{\pm 1.65}}$ & 61.33$_{\textcolor{gray}{\pm 1.26}}$ & \underline{23.58}$_{\textcolor{gray}{\pm 1.66}}$
& 54.09$_{\textcolor{gray}{\pm 2.11}}$ & \textbf{83.32}$_{\textcolor{gray}{\pm 1.53}}$ & \underline{29.23}$_{\textcolor{gray}{\pm 1.74}}$
& 38.23$_{\textcolor{gray}{\pm 1.70}}$ & \textbf{70.67}$_{\textcolor{gray}{\pm 1.29}}$ & \underline{32.44}$_{\textcolor{gray}{\pm 1.53}}$ \\

MDA~\cite{mda}
& 38.62$_{\textcolor{gray}{\pm 1.53}}$ & 58.08$_{\textcolor{gray}{\pm 1.37}}$ & 19.47$_{\textcolor{gray}{\pm 1.72}}$
& 56.50$_{\textcolor{gray}{\pm 2.03}}$ & 80.94$_{\textcolor{gray}{\pm 1.56}}$ & 24.44$_{\textcolor{gray}{\pm 1.56}}$
& 51.85$_{\textcolor{gray}{\pm 1.67}}$ & 53.08$_{\textcolor{gray}{\pm 1.49}}$ & 1.24$_{\textcolor{gray}{\pm 1.58}}$ \\

MUTEX~\cite{mutex}
& 46.57$_{\textcolor{gray}{\pm 1.62}}$ & 54.85$_{\textcolor{gray}{\pm 1.31}}$ & 8.29$_{\textcolor{gray}{\pm 1.54}}$
& 56.67$_{\textcolor{gray}{\pm 1.81}}$ & 78.56$_{\textcolor{gray}{\pm 1.67}}$ & 21.89$_{\textcolor{gray}{\pm 1.42}}$
& 52.54$_{\textcolor{gray}{\pm 1.93}}$ & 63.90$_{\textcolor{gray}{\pm 1.50}}$ & 11.36$_{\textcolor{gray}{\pm 1.67}}$ \\

\rowcolor{blue!10}
CAAP (Ours)
& \textbf{32.44}$_{\textcolor{gray}{\pm 1.46}}$ & \underline{62.28}$_{\textcolor{gray}{\pm 1.25}}$ & \textbf{29.84}$_{\textcolor{gray}{\pm 1.61}}$
& \textbf{44.42}$_{\textcolor{gray}{\pm 1.84}}$ & \underline{82.04}$_{\textcolor{gray}{\pm 1.58}}$ & \textbf{37.62}$_{\textcolor{gray}{\pm 1.58}}$
& \textbf{29.55}$_{\textcolor{gray}{\pm 1.34}}$ & 64.49$_{\textcolor{gray}{\pm 1.44}}$ & \textbf{34.94}$_{\textcolor{gray}{\pm 1.25}}$ \\
\bottomrule
\end{tabular}
}
\end{table*}

\begin{table*}[t]
\caption{Localization quantitative results on ImageNet for CLIP-H/14~\cite{clip}, DINOv2-G/14~\cite{dino2}, and DeiT3-H/14~\cite{deit}. AUPR$_1$ ($\uparrow$), AUPR$_0$ ($\uparrow$), and PG ($\uparrow$) are reported. Best is \textbf{bold}, second best is \underline{underlined} in each column.}
\vspace{-1.5mm}
\label{tab:localization_comparison_results_largest}
\centering
\small
\setlength{\tabcolsep}{3.8pt}
\renewcommand{\arraystretch}{1.15}
\resizebox{\textwidth}{!}{%
\begin{tabular}{lccccccccc}
\toprule
& \multicolumn{3}{c}{CLIP-H/14~\cite{clip}}
& \multicolumn{3}{c}{DINOv2-G/14~\cite{dino2}}
& \multicolumn{3}{c}{DeiT3-H/14~\cite{deit}} \\
\cmidrule(lr){2-4}\cmidrule(lr){5-7}\cmidrule(lr){8-10}
Method
& AUPR$_1$ $\uparrow$ & AUPR$_0$ $\uparrow$ & PG $\uparrow$
& AUPR$_1$ $\uparrow$ & AUPR$_0$ $\uparrow$ & PG $\uparrow$
& AUPR$_1$ $\uparrow$ & AUPR$_0$ $\uparrow$ & PG $\uparrow$ \\
\midrule

RISE~\cite{Petsiuk2018RISE}
& 47.27$_{\textcolor{gray}{\pm 3.03}}$ & 71.21$_{\textcolor{gray}{\pm 3.39}}$ & 51.85$_{\textcolor{gray}{\pm 6.80}}$
& 32.43$_{\textcolor{gray}{\pm 4.27}}$ & 68.07$_{\textcolor{gray}{\pm 4.43}}$ & 25.00$_{\textcolor{gray}{\pm 8.84}}$
& 34.94$_{\textcolor{gray}{\pm 4.43}}$ & 68.90$_{\textcolor{gray}{\pm 4.55}}$ & 50.00$_{\textcolor{gray}{\pm 10.21}}$ \\

Grad-CAM~\cite{grad_cam}
& 39.31$_{\textcolor{gray}{\pm 2.24}}$ & 71.89$_{\textcolor{gray}{\pm 2.32}}$ & 27.34$_{\textcolor{gray}{\pm 3.94}}$
& 47.64$_{\textcolor{gray}{\pm 2.25}}$ & \underline{77.12}$_{\textcolor{gray}{\pm 2.20}}$ & 44.53$_{\textcolor{gray}{\pm 4.39}}$
& \underline{64.60}$_{\textcolor{gray}{\pm 1.88}}$ & \underline{85.08}$_{\textcolor{gray}{\pm 1.89}}$ & \underline{62.50}$_{\textcolor{gray}{\pm 4.28}}$ \\

Rollout~\cite{attn_r}
& 41.40$_{\textcolor{gray}{\pm 2.15}}$ & 71.28$_{\textcolor{gray}{\pm 2.54}}$ & 37.50$_{\textcolor{gray}{\pm 4.28}}$
& 33.66$_{\textcolor{gray}{\pm 1.97}}$ & 65.90$_{\textcolor{gray}{\pm 2.38}}$ & 38.28$_{\textcolor{gray}{\pm 4.30}}$
& 28.49$_{\textcolor{gray}{\pm 1.95}}$ & 64.97$_{\textcolor{gray}{\pm 2.34}}$ & 2.34$_{\textcolor{gray}{\pm 1.34}}$ \\

T-Attr~\cite{chefer2021transformer}
& 43.87$_{\textcolor{gray}{\pm 1.76}}$ & 73.82$_{\textcolor{gray}{\pm 2.44}}$ & 33.59$_{\textcolor{gray}{\pm 4.17}}$
& \underline{52.91}$_{\textcolor{gray}{\pm 1.79}}$ & 76.06$_{\textcolor{gray}{\pm 2.46}}$ & \underline{46.88}$_{\textcolor{gray}{\pm 4.41}}$
& 54.06$_{\textcolor{gray}{\pm 1.63}}$ & 80.76$_{\textcolor{gray}{\pm 2.06}}$ & 28.91$_{\textcolor{gray}{\pm 4.01}}$ \\

CausalX~\cite{vit_cx}
& 51.98$_{\textcolor{gray}{\pm 2.32}}$ & 75.55$_{\textcolor{gray}{\pm 2.35}}$ & \underline{58.59}$_{\textcolor{gray}{\pm 4.35}}$
& 45.94$_{\textcolor{gray}{\pm 2.37}}$ & 71.65$_{\textcolor{gray}{\pm 2.38}}$ & 41.41$_{\textcolor{gray}{\pm 4.35}}$
& 52.49$_{\textcolor{gray}{\pm 2.40}}$ & 76.23$_{\textcolor{gray}{\pm 2.26}}$ & 50.78$_{\textcolor{gray}{\pm 4.42}}$ \\

TAM~\cite{tam}
& 46.87$_{\textcolor{gray}{\pm 1.91}}$ & 73.12$_{\textcolor{gray}{\pm 2.47}}$ & 42.19$_{\textcolor{gray}{\pm 4.37}}$
& 40.35$_{\textcolor{gray}{\pm 1.87}}$ & 71.75$_{\textcolor{gray}{\pm 2.53}}$ & 28.12$_{\textcolor{gray}{\pm 3.97}}$
& 59.52$_{\textcolor{gray}{\pm 1.79}}$ & 80.91$_{\textcolor{gray}{\pm 2.09}}$ & 53.91$_{\textcolor{gray}{\pm 4.41}}$ \\

IxG+~\cite{mehri-skipplus-cvpr24}
& 32.55$_{\textcolor{gray}{\pm 2.13}}$ & 68.64$_{\textcolor{gray}{\pm 2.10}}$ & 22.66$_{\textcolor{gray}{\pm 3.70}}$
& 39.26$_{\textcolor{gray}{\pm 1.97}}$ & 67.08$_{\textcolor{gray}{\pm 2.12}}$ & 39.06$_{\textcolor{gray}{\pm 4.31}}$
& 34.94$_{\textcolor{gray}{\pm 1.89}}$ & 69.09$_{\textcolor{gray}{\pm 2.18}}$ & 8.59$_{\textcolor{gray}{\pm 2.48}}$ \\

Libra IxG+~\cite{Mehri_2025_CVPR}
& \underline{57.57}$_{\textcolor{gray}{\pm 1.82}}$ & \underline{78.46}$_{\textcolor{gray}{\pm 2.09}}$ & 57.03$_{\textcolor{gray}{\pm 4.38}}$
& 46.72$_{\textcolor{gray}{\pm 1.80}}$ & 73.03$_{\textcolor{gray}{\pm 2.33}}$ & 47.66$_{\textcolor{gray}{\pm 4.41}}$
& 52.22$_{\textcolor{gray}{\pm 1.72}}$ & 79.70$_{\textcolor{gray}{\pm 2.03}}$ & 2.34$_{\textcolor{gray}{\pm 1.34}}$ \\

MDA~\cite{mda}
& 48.22$_{\textcolor{gray}{\pm 3.25}}$ & 70.00$_{\textcolor{gray}{\pm 4.03}}$ & 30.77$_{\textcolor{gray}{\pm 6.40}}$
& 40.89$_{\textcolor{gray}{\pm 2.92}}$ & 64.86$_{\textcolor{gray}{\pm 4.08}}$ & 13.21$_{\textcolor{gray}{\pm 4.65}}$
& 31.18$_{\textcolor{gray}{\pm 2.00}}$ & 66.16$_{\textcolor{gray}{\pm 2.57}}$ & 14.17$_{\textcolor{gray}{\pm 3.09}}$ \\

MUTEX~\cite{mutex}
& 39.40$_{\textcolor{gray}{\pm 2.24}}$ & 69.56$_{\textcolor{gray}{\pm 2.29}}$ & 39.84$_{\textcolor{gray}{\pm 4.33}}$
& 43.24$_{\textcolor{gray}{\pm 2.17}}$ & \textbf{77.42}$_{\textcolor{gray}{\pm 2.18}}$ & 35.16$_{\textcolor{gray}{\pm 4.22}}$
& 38.87$_{\textcolor{gray}{\pm 2.06}}$ & 69.99$_{\textcolor{gray}{\pm 2.44}}$ & 19.53$_{\textcolor{gray}{\pm 3.50}}$ \\

\rowcolor{blue!10}
CAAP (Ours)
& \textbf{63.05}$_{\textcolor{gray}{\pm 1.83}}$ & \textbf{81.30}$_{\textcolor{gray}{\pm 2.00}}$ & \textbf{65.63}$_{\textcolor{gray}{\pm 4.20}}$
& \textbf{56.22}$_{\textcolor{gray}{\pm 3.01}}$ & 77.03$_{\textcolor{gray}{\pm 3.29}}$ & \textbf{50.91}$_{\textcolor{gray}{\pm 6.74}}$
& \textbf{68.70}$_{\textcolor{gray}{\pm 1.87}}$ & \textbf{86.24}$_{\textcolor{gray}{\pm 1.83}}$ & \textbf{72.66}$_{\textcolor{gray}{\pm 3.94}}$ \\

\bottomrule
\end{tabular}
}
\end{table*}

%% file: tables/compactness_scores.tex
\begin{table*}[t]
\caption{Complementary faithfulness and compactness quantitative results on ImageNet for ViT-L/16~\cite{vit}, CLIP-L/14~\cite{clip}, DINOv2-L/14~\cite{dino2}, and DeiT3-L/16~\cite{deit}. Infidelity ($\downarrow$, $\times 10^3$), Entropy ($\downarrow$), and Gini ($\uparrow$) are reported.  Best is \textbf{bold}, second best is \underline{underlined} in each column.}
\vspace{-1.5mm}
\label{tab:compactness_comparison_results_main}
\centering
\small
\setlength{\tabcolsep}{3pt}
\renewcommand{\arraystretch}{1.15}
\resizebox{\textwidth}{!}{%
\begin{tabular}{lcccccccccccc}
\toprule
& \multicolumn{3}{c}{ViT-L/16~\cite{vit}}
& \multicolumn{3}{c}{CLIP-L/14~\cite{clip}}
& \multicolumn{3}{c}{DINOv2-L/14~\cite{dino2}}
& \multicolumn{3}{c}{DeiT3-L/16~\cite{deit}} \\
\cmidrule(lr){2-4}\cmidrule(lr){5-7}\cmidrule(lr){8-10}\cmidrule(lr){11-13}
Method
& Ent$\downarrow$ & Gini$\uparrow$ & Infid.$\downarrow$
& Ent$\downarrow$ & Gini$\uparrow$ & Infid.$\downarrow$
& Ent$\downarrow$ & Gini$\uparrow$ & Infid.$\downarrow$
& Ent$\downarrow$ & Gini$\uparrow$ & Infid.$\downarrow$ \\
\midrule
RISE~\cite{Petsiuk2018RISE}
& 97.93$_{\textcolor{gray}{\pm 0.04}}$ & 24.58$_{\textcolor{gray}{\pm 0.24}}$ & 2.71$_{\textcolor{gray}{\pm 0.25}}$ & 98.03$_{\textcolor{gray}{\pm 0.03}}$ & 24.80$_{\textcolor{gray}{\pm 0.18}}$ & 2.20$_{\textcolor{gray}{\pm 0.22}}$ & 98.35$_{\textcolor{gray}{\pm 0.03}}$ & 22.65$_{\textcolor{gray}{\pm 0.18}}$ & 1.37$_{\textcolor{gray}{\pm 0.15}}$ & 98.02$_{\textcolor{gray}{\pm 0.04}}$ & 24.02$_{\textcolor{gray}{\pm 0.23}}$ & \underline{2.03}$_{\textcolor{gray}{\pm 0.18}}$ \\

Grad-CAM~\cite{grad_cam}
& 93.33$_{\textcolor{gray}{\pm 0.20}}$ & 42.73$_{\textcolor{gray}{\pm 0.59}}$ & 2.42$_{\textcolor{gray}{\pm 0.21}}$ & \textbf{73.09}$_{\textcolor{gray}{\pm 0.50}}$ & \textbf{79.85}$_{\textcolor{gray}{\pm 0.50}}$ & 1.83$_{\textcolor{gray}{\pm 0.18}}$ & 94.39$_{\textcolor{gray}{\pm 0.14}}$ & 41.03$_{\textcolor{gray}{\pm 0.49}}$ & 1.41$_{\textcolor{gray}{\pm 0.16}}$ & 85.90$_{\textcolor{gray}{\pm 0.30}}$ & 60.21$_{\textcolor{gray}{\pm 0.59}}$ & 3.21$_{\textcolor{gray}{\pm 0.19}}$ \\

Rollout~\cite{attn_r}
& 96.42$_{\textcolor{gray}{\pm 0.03}}$ & 33.53$_{\textcolor{gray}{\pm 0.15}}$ & 2.78$_{\textcolor{gray}{\pm 0.27}}$ & 96.59$_{\textcolor{gray}{\pm 0.05}}$ & 33.09$_{\textcolor{gray}{\pm 0.22}}$ & 2.22$_{\textcolor{gray}{\pm 0.23}}$ & \underline{83.98}$_{\textcolor{gray}{\pm 0.12}}$ & \underline{60.53}$_{\textcolor{gray}{\pm 0.20}}$ & 1.46$_{\textcolor{gray}{\pm 0.16}}$ & 93.28$_{\textcolor{gray}{\pm 0.08}}$ & 41.32$_{\textcolor{gray}{\pm 0.27}}$ & 2.32$_{\textcolor{gray}{\pm 0.20}}$ \\

T-Attr~\cite{chefer2021transformer}
& 88.93$_{\textcolor{gray}{\pm 0.12}}$ & 53.99$_{\textcolor{gray}{\pm 0.30}}$ & 2.49$_{\textcolor{gray}{\pm 0.22}}$ & 89.15$_{\textcolor{gray}{\pm 0.41}}$ & 55.24$_{\textcolor{gray}{\pm 0.88}}$ & 1.88$_{\textcolor{gray}{\pm 0.18}}$ & 95.89$_{\textcolor{gray}{\pm 0.12}}$ & 32.75$_{\textcolor{gray}{\pm 0.37}}$ & 1.41$_{\textcolor{gray}{\pm 0.15}}$ & \underline{81.01}$_{\textcolor{gray}{\pm 0.31}}$ & \underline{69.68}$_{\textcolor{gray}{\pm 0.42}}$ & 3.25$_{\textcolor{gray}{\pm 0.17}}$ \\

ViT-CX~\cite{vit_cx}
& 97.14$_{\textcolor{gray}{\pm 0.07}}$ & 27.79$_{\textcolor{gray}{\pm 0.44}}$ & 2.73$_{\textcolor{gray}{\pm 0.26}}$ & 97.63$_{\textcolor{gray}{\pm 0.07}}$ & 25.87$_{\textcolor{gray}{\pm 0.43}}$ & 2.22$_{\textcolor{gray}{\pm 0.23}}$ & 97.31$_{\textcolor{gray}{\pm 0.05}}$ & 28.51$_{\textcolor{gray}{\pm 0.32}}$ & 1.37$_{\textcolor{gray}{\pm 0.15}}$ & 95.90$_{\textcolor{gray}{\pm 0.10}}$ & 33.95$_{\textcolor{gray}{\pm 0.44}}$ & 2.06$_{\textcolor{gray}{\pm 0.18}}$ \\

TAM~\cite{tam}
& 94.50$_{\textcolor{gray}{\pm 0.10}}$ & 40.66$_{\textcolor{gray}{\pm 0.33}}$ & 2.42$_{\textcolor{gray}{\pm 0.22}}$ & 94.42$_{\textcolor{gray}{\pm 0.13}}$ & 41.18$_{\textcolor{gray}{\pm 0.42}}$ & 1.94$_{\textcolor{gray}{\pm 0.20}}$ & 96.60$_{\textcolor{gray}{\pm 0.06}}$ & 32.98$_{\textcolor{gray}{\pm 0.28}}$ & \underline{1.34}$_{\textcolor{gray}{\pm 0.15}}$ & 86.52$_{\textcolor{gray}{\pm 0.24}}$ & 60.22$_{\textcolor{gray}{\pm 0.48}}$ & 2.95$_{\textcolor{gray}{\pm 0.17}}$ \\

IxG+~\cite{mehri-skipplus-cvpr24}
& 99.42$_{\textcolor{gray}{\pm 0.02}}$ & 10.31$_{\textcolor{gray}{\pm 0.22}}$ & 2.92$_{\textcolor{gray}{\pm 0.29}}$ & 99.55$_{\textcolor{gray}{\pm 0.02}}$ & 8.85$_{\textcolor{gray}{\pm 0.19}}$ & 2.07$_{\textcolor{gray}{\pm 0.21}}$ & 99.47$_{\textcolor{gray}{\pm 0.01}}$ & 10.74$_{\textcolor{gray}{\pm 0.17}}$ & 2.05$_{\textcolor{gray}{\pm 0.23}}$ & 96.51$_{\textcolor{gray}{\pm 0.20}}$ & 18.93$_{\textcolor{gray}{\pm 0.57}}$ & 6.35$_{\textcolor{gray}{\pm 0.29}}$ \\

Libra IxG+~\cite{Mehri_2025_CVPR}
& 97.57$_{\textcolor{gray}{\pm 0.09}}$ & 23.50$_{\textcolor{gray}{\pm 0.46}}$ & \underline{2.27}$_{\textcolor{gray}{\pm 0.17}}$ & 98.99$_{\textcolor{gray}{\pm 0.05}}$ & 12.77$_{\textcolor{gray}{\pm 0.35}}$ & \underline{1.63}$_{\textcolor{gray}{\pm 0.15}}$ & 98.52$_{\textcolor{gray}{\pm 0.06}}$ & 19.10$_{\textcolor{gray}{\pm 0.40}}$ & 1.51$_{\textcolor{gray}{\pm 0.17}}$ & 94.88$_{\textcolor{gray}{\pm 0.17}}$ & 34.49$_{\textcolor{gray}{\pm 0.60}}$ & 2.96$_{\textcolor{gray}{\pm 0.19}}$ \\

MDA~\cite{mda}
& \underline{86.89}$_{\textcolor{gray}{\pm 0.29}}$ & \underline{58.24}$_{\textcolor{gray}{\pm 0.57}}$ & 2.89$_{\textcolor{gray}{\pm 0.29}}$ & 85.47$_{\textcolor{gray}{\pm 0.31}}$ & 62.69$_{\textcolor{gray}{\pm 0.57}}$ & 2.04$_{\textcolor{gray}{\pm 0.22}}$ & 86.37$_{\textcolor{gray}{\pm 0.29}}$ & 59.97$_{\textcolor{gray}{\pm 0.53}}$ & 1.58$_{\textcolor{gray}{\pm 0.17}}$ & 90.40$_{\textcolor{gray}{\pm 0.20}}$ & 51.83$_{\textcolor{gray}{\pm 0.52}}$ & 3.52$_{\textcolor{gray}{\pm 0.22}}$ \\

MUTEX~\cite{mutex}
& 98.40$_{\textcolor{gray}{\pm 0.03}}$ & 21.63$_{\textcolor{gray}{\pm 0.18}}$ & 2.68$_{\textcolor{gray}{\pm 0.25}}$ & 98.38$_{\textcolor{gray}{\pm 0.02}}$ & 22.48$_{\textcolor{gray}{\pm 0.18}}$ & 2.20$_{\textcolor{gray}{\pm 0.22}}$ & 97.58$_{\textcolor{gray}{\pm 0.05}}$ & 27.08$_{\textcolor{gray}{\pm 0.31}}$ & 1.38$_{\textcolor{gray}{\pm 0.15}}$ & 98.40$_{\textcolor{gray}{\pm 0.02}}$ & 21.55$_{\textcolor{gray}{\pm 0.17}}$ & \textbf{2.02}$_{\textcolor{gray}{\pm 0.18}}$ \\

\rowcolor{blue!10}
CAAP (Ours)
& \textbf{85.83}$_{\textcolor{gray}{\pm 0.36}}$ & \textbf{60.55}$_{\textcolor{gray}{\pm 0.64}}$ & \textbf{2.19}$_{\textcolor{gray}{\pm 0.19}}$ & \underline{79.08}$_{\textcolor{gray}{\pm 0.41}}$ & \underline{72.69}$_{\textcolor{gray}{\pm 0.57}}$ & \textbf{1.62}$_{\textcolor{gray}{\pm 0.15}}$ & \textbf{76.00}$_{\textcolor{gray}{\pm 0.46}}$ & \textbf{75.67}$_{\textcolor{gray}{\pm 0.58}}$ & \textbf{1.27}$_{\textcolor{gray}{\pm 0.17}}$ & \textbf{75.59}$_{\textcolor{gray}{\pm 0.47}}$ & \textbf{74.30}$_{\textcolor{gray}{\pm 0.63}}$ & 2.50$_{\textcolor{gray}{\pm 0.18}}$ \\

\bottomrule
\end{tabular}
}
\end{table*}

%% file: tables/non_cls_readouts.tex
\begin{table*}[t]
\caption{Faithfulness and localization quantitative results for BEiT~\cite{bao2022beit} and SigLIP~\cite{Zhai_2023_ICCV}. Deletion ($\downarrow$), Insertion ($\uparrow$), Ins$-$Del ($\uparrow$), AUPR$_1$ ($\uparrow$), AUPR$_0$ ($\uparrow$), and PG ($\uparrow$) are reported. Best is \textbf{bold}, second best is \underline{underlined} in each column.}
\vspace{-1.5mm}
\label{tab:cr_readouts}
\centering
\small
\setlength{\tabcolsep}{3.8pt}
\renewcommand{\arraystretch}{1.15}
\resizebox{\textwidth}{!}{%
\begin{tabular}{l*{12}{c}}
\toprule
& \multicolumn{6}{c}{BEiT~\cite{bao2022beit}} & \multicolumn{6}{c}{SigLIP~\cite{Zhai_2023_ICCV}} \\
\cmidrule(lr){2-7}\cmidrule(lr){8-13}
Method & Del$\downarrow$ & Ins$\uparrow$ & Ins$-$Del$\uparrow$ & AUPR$_1\uparrow$ & AUPR$_0\uparrow$ & PG$\uparrow$ & Del$\downarrow$ & Ins$\uparrow$ & Ins$-$Del$\uparrow$ & AUPR$_1\uparrow$ & AUPR$_0\uparrow$ & PG$\uparrow$ \\
\midrule
Grad-CAM~\cite{grad_cam} & 75.2$_{\textcolor{gray}{\pm 1.51}}$ & 65.3$_{\textcolor{gray}{\pm 1.89}}$ & -9.9$_{\textcolor{gray}{\pm 1.54}}$ & 31.5$_{\textcolor{gray}{\pm 2.24}}$ & 65.7$_{\textcolor{gray}{\pm 2.17}}$ & 26.6$_{\textcolor{gray}{\pm 3.90}}$ & 71.8$_{\textcolor{gray}{\pm 1.45}}$ & 66.5$_{\textcolor{gray}{\pm 3.34}}$ & -4.5$_{\textcolor{gray}{\pm 1.46}}$ & 36.8$_{\textcolor{gray}{\pm 3.50}}$ & 63.4$_{\textcolor{gray}{\pm 3.39}}$ & 17.0$_{\textcolor{gray}{\pm 5.16}}$ \\
Rollout~\cite{attn_r} & 73.5$_{\textcolor{gray}{\pm 1.61}}$ & 67.7$_{\textcolor{gray}{\pm 1.81}}$ & -5.8$_{\textcolor{gray}{\pm 1.36}}$ & 36.7$_{\textcolor{gray}{\pm 1.99}}$ & 66.3$_{\textcolor{gray}{\pm 2.40}}$ & 37.5$_{\textcolor{gray}{\pm 4.28}}$ & 56.2$_{\textcolor{gray}{\pm 1.57}}$ & 67.7$_{\textcolor{gray}{\pm 3.14}}$ & 14.3$_{\textcolor{gray}{\pm 1.49}}$ & 42.1$_{\textcolor{gray}{\pm 3.13}}$ & 73.0$_{\textcolor{gray}{\pm 3.50}}$ & 5.7$_{\textcolor{gray}{\pm 3.17}}$ \\
T-Attr~\cite{chefer2021transformer} & 75.9$_{\textcolor{gray}{\pm 1.54}}$ & 63.0$_{\textcolor{gray}{\pm 1.88}}$ & -12.9$_{\textcolor{gray}{\pm 1.43}}$ & 29.8$_{\textcolor{gray}{\pm 2.19}}$ & 64.2$_{\textcolor{gray}{\pm 2.15}}$ & 35.9$_{\textcolor{gray}{\pm 4.24}}$ & 66.4$_{\textcolor{gray}{\pm 1.52}}$ & 69.6$_{\textcolor{gray}{\pm 3.08}}$ & 5.0$_{\textcolor{gray}{\pm 3.59}}$ & 36.1$_{\textcolor{gray}{\pm 3.45}}$ & 65.3$_{\textcolor{gray}{\pm 3.37}}$ & 28.3$_{\textcolor{gray}{\pm 6.19}}$ \\
ViT-CX~\cite{vit_cx} & 56.3$_{\textcolor{gray}{\pm 1.91}}$ & 74.2$_{\textcolor{gray}{\pm 1.54}}$ & 17.9$_{\textcolor{gray}{\pm 1.70}}$ & 51.7$_{\textcolor{gray}{\pm 2.28}}$ & 76.0$_{\textcolor{gray}{\pm 2.27}}$ & 58.6$_{\textcolor{gray}{\pm 4.35}}$ & 46.6$_{\textcolor{gray}{\pm 1.91}}$ & \textbf{78.7}$_{\textcolor{gray}{\pm 1.94}}$ & 33.8$_{\textcolor{gray}{\pm 2.93}}$ & 61.8$_{\textcolor{gray}{\pm 3.95}}$ & 75.7$_{\textcolor{gray}{\pm 3.78}}$ & 67.9$_{\textcolor{gray}{\pm 6.41}}$ \\
TAM~\cite{tam} & 59.8$_{\textcolor{gray}{\pm 1.90}}$ & 73.0$_{\textcolor{gray}{\pm 1.62}}$ & 13.3$_{\textcolor{gray}{\pm 1.71}}$ & 45.8$_{\textcolor{gray}{\pm 1.96}}$ & 73.7$_{\textcolor{gray}{\pm 2.50}}$ & 40.6$_{\textcolor{gray}{\pm 4.34}}$ & 52.7$_{\textcolor{gray}{\pm 1.63}}$ & 69.8$_{\textcolor{gray}{\pm 3.17}}$ & 18.9$_{\textcolor{gray}{\pm 3.71}}$ & 44.5$_{\textcolor{gray}{\pm 3.12}}$ & 72.6$_{\textcolor{gray}{\pm 3.74}}$ & 26.4$_{\textcolor{gray}{\pm 6.06}}$ \\
IxG+~\cite{mehri-skipplus-cvpr24} & 67.8$_{\textcolor{gray}{\pm 1.81}}$ & 73.7$_{\textcolor{gray}{\pm 1.56}}$ & 5.9$_{\textcolor{gray}{\pm 1.37}}$ & 39.8$_{\textcolor{gray}{\pm 1.94}}$ & 72.7$_{\textcolor{gray}{\pm 2.15}}$ & 27.3$_{\textcolor{gray}{\pm 3.94}}$ & 59.1$_{\textcolor{gray}{\pm 6.11}}$ & 75.5$_{\textcolor{gray}{\pm 4.74}}$ & 18.2$_{\textcolor{gray}{\pm 7.81}}$ & 45.9$_{\textcolor{gray}{\pm 4.70}}$ & 72.4$_{\textcolor{gray}{\pm 5.54}}$ & 14.1$_{\textcolor{gray}{\pm 14.87}}$ \\
Libra IxG+~\cite{Mehri_2025_CVPR} & \underline{49.7}$_{\textcolor{gray}{\pm 2.09}}$ & \textbf{77.3}$_{\textcolor{gray}{\pm 1.52}}$ & \underline{27.6}$_{\textcolor{gray}{\pm 1.90}}$ & \underline{59.3}$_{\textcolor{gray}{\pm 1.74}}$ & \underline{80.3}$_{\textcolor{gray}{\pm 2.09}}$ & \underline{65.6}$_{\textcolor{gray}{\pm 4.20}}$ & \underline{40.1}$_{\textcolor{gray}{\pm 6.20}}$ & \underline{78.6}$_{\textcolor{gray}{\pm 4.71}}$ & \underline{40.3}$_{\textcolor{gray}{\pm 7.85}}$ & \underline{67.2}$_{\textcolor{gray}{\pm 4.51}}$ & \underline{79.7}$_{\textcolor{gray}{\pm 5.49}}$ & \underline{72.2}$_{\textcolor{gray}{\pm 15.02}}$ \\
MDA~\cite{mda} & 67.4$_{\textcolor{gray}{\pm 1.67}}$ & 69.5$_{\textcolor{gray}{\pm 1.68}}$ & 2.1$_{\textcolor{gray}{\pm 1.62}}$ & 38.0$_{\textcolor{gray}{\pm 2.27}}$ & 68.1$_{\textcolor{gray}{\pm 2.83}}$ & 35.7$_{\textcolor{gray}{\pm 4.84}}$ & 58.7$_{\textcolor{gray}{\pm 6.08}}$ & 71.7$_{\textcolor{gray}{\pm 4.85}}$ & 14.8$_{\textcolor{gray}{\pm 7.85}}$ & 43.9$_{\textcolor{gray}{\pm 5.06}}$ & 67.9$_{\textcolor{gray}{\pm 6.24}}$ & 26.8$_{\textcolor{gray}{\pm 15.42}}$ \\
MUTEX~\cite{mutex} & 54.7$_{\textcolor{gray}{\pm 1.95}}$ & 74.2$_{\textcolor{gray}{\pm 1.54}}$ & 19.6$_{\textcolor{gray}{\pm 1.70}}$ & 47.9$_{\textcolor{gray}{\pm 2.18}}$ & 75.5$_{\textcolor{gray}{\pm 2.43}}$ & 44.5$_{\textcolor{gray}{\pm 4.39}}$ & 46.0$_{\textcolor{gray}{\pm 1.90}}$ & 76.7$_{\textcolor{gray}{\pm 2.20}}$ & 32.5$_{\textcolor{gray}{\pm 3.09}}$ & 53.8$_{\textcolor{gray}{\pm 3.29}}$ & 71.7$_{\textcolor{gray}{\pm 3.84}}$ & 52.8$_{\textcolor{gray}{\pm 6.86}}$ \\
\rowcolor{blue!10}
CAAP (Ours) & \textbf{45.7}$_{\textcolor{gray}{\pm 1.98}}$ & \underline{76.2}$_{\textcolor{gray}{\pm 1.56}}$ & \textbf{30.5}$_{\textcolor{gray}{\pm 1.91}}$ & \textbf{65.0}$_{\textcolor{gray}{\pm 1.76}}$ & \textbf{82.2}$_{\textcolor{gray}{\pm 1.97}}$ & \textbf{70.3}$_{\textcolor{gray}{\pm 4.04}}$ & \textbf{34.7}$_{\textcolor{gray}{\pm 1.63}}$ & 78.1$_{\textcolor{gray}{\pm 2.46}}$ & \textbf{45.1}$_{\textcolor{gray}{\pm 1.55}}$ & \textbf{73.8}$_{\textcolor{gray}{\pm 2.60}}$ & \textbf{83.1}$_{\textcolor{gray}{\pm 3.03}}$ & \textbf{75.5}$_{\textcolor{gray}{\pm 5.91}}$ \\
\bottomrule
\end{tabular}
}
\end{table*}

%% file: tables/ablation_efficiency.tex
\begin{table*}[t]
\caption{Faithfulness quantitative results for CAAP and Efficient-CAAP (E-CAAP) on ImageNet. Deletion ($\downarrow$), Insertion ($\uparrow$), and Ins$-$Del ($\uparrow$) are reported.}
\vspace{-1.5mm}
\label{tab:faithfulness_caap_ecaap_comparison}
\centering
\small
\setlength{\tabcolsep}{4pt}
\renewcommand{\arraystretch}{1.15}
\resizebox{\textwidth}{!}{%
\begin{tabular}{lcccccccccccc}
\toprule
& \multicolumn{3}{c}{ViT-L/16~\cite{vit}}
& \multicolumn{3}{c}{CLIP-L/14~\cite{clip}}
& \multicolumn{3}{c}{DINOv2-L/14~\cite{dino2}}
& \multicolumn{3}{c}{DeiT3-L/16~\cite{deit}} \\
\cmidrule(lr){2-4}\cmidrule(lr){5-7}\cmidrule(lr){8-10}\cmidrule(lr){11-13}
Method
& Del$\downarrow$ & Ins$\uparrow$ & Ins$-$Del$\uparrow$
& Del$\downarrow$ & Ins$\uparrow$ & Ins$-$Del$\uparrow$
& Del$\downarrow$ & Ins$\uparrow$ & Ins$-$Del$\uparrow$
& Del$\downarrow$ & Ins$\uparrow$ & Ins$-$Del$\uparrow$ \\
\midrule

CAAP
& 43.61$_{\textcolor{gray}{\pm 2.73}}$ & 71.10$_{\textcolor{gray}{\pm 2.61}}$ & 27.50$_{\textcolor{gray}{\pm 2.71}}$
& 34.20$_{\textcolor{gray}{\pm 2.21}}$ & 74.92$_{\textcolor{gray}{\pm 2.01}}$ & 40.73$_{\textcolor{gray}{\pm 2.15}}$
& 43.48$_{\textcolor{gray}{\pm 2.76}}$ & 79.05$_{\textcolor{gray}{\pm 2.50}}$ & 35.57$_{\textcolor{gray}{\pm 2.16}}$
& 28.93$_{\textcolor{gray}{\pm 1.84}}$ & 65.79$_{\textcolor{gray}{\pm 2.06}}$ & 36.86$_{\textcolor{gray}{\pm 1.82}}$ \\

E-CAAP
&42.56$_{\textcolor{gray}{\pm 2.80}}$ &70.23$_{\textcolor{gray}{\pm 2.56}}$ &27.67$_{\textcolor{gray}{\pm 2.63}}$ 
&33.56$_{\textcolor{gray}{\pm 2.42}}$
&74.68$_{\textcolor{gray}{\pm 2.58}}$ 
&41.12$_{\textcolor{gray}{\pm 2.39}}$
&41.73$_{\textcolor{gray}{\pm 2.79}}$
&78.44$_{\textcolor{gray}{\pm 2.75}}$
&36.71$_{\textcolor{gray}{\pm 2.65}}$
&29.72$_{\textcolor{gray}{\pm 1.87}}$
&65.55$_{\textcolor{gray}{\pm 2.33}}$
&35.83$_{\textcolor{gray}{\pm 1.97}}$  \\

\bottomrule
\end{tabular}
}
\end{table*}

\begin{table*}[t]
\caption{Localization quantitative results for CAAP and Efficient-CAAP (E-CAAP) on ImageNet. AUPR$_1$ ($\uparrow$), AUPR$_0$ ($\uparrow$), and PG ($\uparrow$) are reported.}
\vspace{-1.5mm}
\label{tab:localization_caap_ecaap_comparison}
\centering
\small
\setlength{\tabcolsep}{3.8pt}
\renewcommand{\arraystretch}{1.15}
\resizebox{\textwidth}{!}{%
\begin{tabular}{lcccccccccccc}
\toprule
& \multicolumn{3}{c}{ViT-L/16~\cite{vit}}
& \multicolumn{3}{c}{CLIP-L/14~\cite{clip}}
& \multicolumn{3}{c}{DINOv2-L/14~\cite{dino2}}
& \multicolumn{3}{c}{DeiT3-L/16~\cite{deit}} \\
\cmidrule(lr){2-4}\cmidrule(lr){5-7}\cmidrule(lr){8-10}\cmidrule(lr){11-13}
Method
& AUPR$_1$ $\uparrow$ & AUPR$_0$ $\uparrow$ & PG $\uparrow$
& AUPR$_1$ $\uparrow$ & AUPR$_0$ $\uparrow$ & PG $\uparrow$
& AUPR$_1$ $\uparrow$ & AUPR$_0$ $\uparrow$ & PG $\uparrow$
& AUPR$_1$ $\uparrow$ & AUPR$_0$ $\uparrow$ & PG $\uparrow$ \\
\midrule

CAAP
& 62.29$_{\textcolor{gray}{\pm 3.70}}$ & 80.83$_{\textcolor{gray}{\pm 3.75}}$ & 77.78$_{\textcolor{gray}{\pm 8.00}}$
& 64.30$_{\textcolor{gray}{\pm 3.68}}$ & 84.44$_{\textcolor{gray}{\pm 3.29}}$ & 74.07$_{\textcolor{gray}{\pm 8.43}}$
& 53.04$_{\textcolor{gray}{\pm 3.88}}$ & 74.96$_{\textcolor{gray}{\pm 4.52}}$ & 37.04$_{\textcolor{gray}{\pm 9.29}}$
& 73.53$_{\textcolor{gray}{\pm 3.10}}$ & 87.19$_{\textcolor{gray}{\pm 3.17}}$ & 85.19$_{\textcolor{gray}{\pm 6.84}}$ \\

E-CAAP
&61.17$_{\textcolor{gray}{\pm 3.11}}$
&80.28$_{\textcolor{gray}{\pm 3.23}}$
&75.66$_{\textcolor{gray}{\pm 4.71}}$
&66.59$_{\textcolor{gray}{\pm 2.41}}$
&83.35$_{\textcolor{gray}{\pm 3.11}}$
&73.14$_{\textcolor{gray}{\pm 4.99}}$
&54.45$_{\textcolor{gray}{\pm 3.16}}$
&75.99$_{\textcolor{gray}{\pm 3.14}}$
&38.75$_{\textcolor{gray}{\pm 4.96}}$
&72.41$_{\textcolor{gray}{\pm 2.84}}$
&86.28$_{\textcolor{gray}{\pm 2.23}}$
&87.67$_{\textcolor{gray}{\pm 3.16}}$  \\

\bottomrule
\end{tabular}
}
\end{table*}

%% file: tables/runtime.tex
\begin{table*}[t]
\caption{Average wall-clock attribution runtime comparison. Time in seconds per image is reported.}
\vspace{-1.5mm}
\label{tab:cr_runtime}
\centering
\small
\setlength{\tabcolsep}{4pt}
\renewcommand{\arraystretch}{1.15}
\resizebox{\textwidth}{!}{%
\begin{tabular}{l*{11}{c}}
\toprule
 & \shortstack{Grad-CAM\\\cite{grad_cam}} & \shortstack{Rollout\\\cite{attn_r}} & \shortstack{T-Attr\\\cite{chefer2021transformer}} & \shortstack{ViT-CX\\\cite{vit_cx}} & \shortstack{TAM\\\cite{tam}} & \shortstack{IxG+\\\cite{mehri-skipplus-cvpr24}} & \shortstack{Libra IxG+\\\cite{Mehri_2025_CVPR}} & \shortstack{MDA\\\cite{mda}} & \shortstack{MUTEX\\\cite{mutex}} & \cellcolor{blue!10}CAAP & \cellcolor{blue!10}E-CAAP \\
\midrule
Seconds/image & 0.43 & 0.22 & 0.29 & 0.87 & 1.00 & 1.27 & 0.36 & 45.93 & 0.84 & \cellcolor{blue!10}8.04 & \cellcolor{blue!10}1.31 \\
\bottomrule
\end{tabular}
}
\end{table*}

%% file: tables/cross_dataset_depth.tex
\begin{table*}[t]
\caption{Faithfulness quantitative results for cumulative intervention-depth sweeps on Food-101~\cite{food101}, ImageNet-Hard~\cite{imagenet_hard}, and Oxford-IIIT Pet~\cite{oxford_pet} using ViT-B/16~\cite{vit}. Deletion ($\downarrow$), Insertion ($\uparrow$), and Ins$-$Del ($\uparrow$) are reported. Best is \textbf{bold}, second best is \underline{underlined} in each column.}
\vspace{-1.5mm}
\label{tab:cr_depth}
\centering
\small
\setlength{\tabcolsep}{4pt}
\renewcommand{\arraystretch}{1.15}
\resizebox{\textwidth}{!}{%
\begin{tabular}{l*{9}{c}}
\toprule
& \multicolumn{3}{c}{Food-101~\cite{food101}} & \multicolumn{3}{c}{ImageNet-Hard~\cite{imagenet_hard}} & \multicolumn{3}{c}{Oxford-IIIT Pet~\cite{oxford_pet}} \\
\cmidrule(lr){2-4}\cmidrule(lr){5-7}\cmidrule(lr){8-10}
Layer range & Del$\downarrow$ & Ins$\uparrow$ & Ins$-$Del$\uparrow$ & Del$\downarrow$ & Ins$\uparrow$ & Ins$-$Del$\uparrow$ & Del$\downarrow$ & Ins$\uparrow$ & Ins$-$Del$\uparrow$ \\
\midrule
0--2 & 46.8 & 66.8 & 20.0 & 19.0 & 27.7 & 8.7 & 61.5 & 83.6 & 22.1 \\
0--4 & 43.7 & \underline{68.0} & 24.2 & 15.2 & 29.8 & 14.6 & 54.9 & 84.8 & 29.9 \\
0--6 & 39.8 & \textbf{68.7} & \textbf{28.9} & \textbf{12.3} & \textbf{34.0} & \textbf{21.7} & 49.6 & \textbf{86.1} & \underline{36.5} \\
0--8 & \textbf{38.1} & 66.8 & \underline{28.7} & \underline{12.6} & \underline{33.4} & \underline{20.8} & \textbf{47.2} & \underline{85.4} & \textbf{38.2} \\
0--10 & \underline{39.4} & 65.1 & 25.7 & 13.9 & 31.1 & 17.2 & \underline{47.8} & 83.9 & 36.1 \\
\bottomrule
\end{tabular}
}
\end{table*}

%% file: tables/ablation_cnn.tex
\begin{table*}[t]
\caption{Faithfulness quantitative results for Input Insertion, Input Deletion, and CAAP on ResNet-18, ResNet-50, DenseNet-161, and VGG-16. Deletion ($\downarrow$), Insertion ($\uparrow$), and Ins$-$Del ($\uparrow$) are reported. Best is \textbf{bold}, second best is \underline{underlined} in each column.}
\vspace{-1.5mm}
\label{tab:faithfulness_cnn_comparison}
\centering
\small
\setlength{\tabcolsep}{4pt}
\renewcommand{\arraystretch}{1.15}
\resizebox{\textwidth}{!}{%
\begin{tabular}{lcccccccccccc}
\toprule
& \multicolumn{3}{c}{ResNet-18~\cite{resnet}}
& \multicolumn{3}{c}{ResNet-50~\cite{resnet}}
& \multicolumn{3}{c}{DenseNet-161~\cite{densenet}}
& \multicolumn{3}{c}{VGG-16~\cite{vgg}} \\
\cmidrule(lr){2-4}\cmidrule(lr){5-7}\cmidrule(lr){8-10}\cmidrule(lr){11-13}
Method
& Del$\downarrow$ & Ins$\uparrow$ & Ins$-$Del$\uparrow$
& Del$\downarrow$ & Ins$\uparrow$ & Ins$-$Del$\uparrow$
& Del$\downarrow$ & Ins$\uparrow$ & Ins$-$Del$\uparrow$
& Del$\downarrow$ & Ins$\uparrow$ & Ins$-$Del$\uparrow$ \\
\midrule

Input Insertion
& 16.05$_{\textcolor{gray}{\pm 0.67}}$ & 38.75$_{\textcolor{gray}{\pm 1.20}}$ & 22.70$_{\textcolor{gray}{\pm 1.13}}$
& 15.65$_{\textcolor{gray}{\pm 0.64}}$ & 49.11$_{\textcolor{gray}{\pm 1.28}}$ & 33.46$_{\textcolor{gray}{\pm 1.25}}$
& 20.83$_{\textcolor{gray}{\pm 0.78}}$ & 52.87$_{\textcolor{gray}{\pm 1.34}}$ & 32.04$_{\textcolor{gray}{\pm 1.26}}$
& 14.60$_{\textcolor{gray}{\pm 0.62}}$ & 41.01$_{\textcolor{gray}{\pm 1.22}}$ & 26.41$_{\textcolor{gray}{\pm 1.14}}$ \\

Input Deletion
& \textbf{12.54}$_{\textcolor{gray}{\pm 0.57}}$ & \underline{54.44}$_{\textcolor{gray}{\pm 1.19}}$ & \underline{41.90}$_{\textcolor{gray}{\pm 1.09}}$
& \textbf{14.74}$_{\textcolor{gray}{\pm 0.78}}$ & \underline{62.19}$_{\textcolor{gray}{\pm 1.18}}$ & \underline{47.45}$_{\textcolor{gray}{\pm 1.22}}$
& \underline{20.35}$_{\textcolor{gray}{\pm 0.95}}$ & \underline{65.98}$_{\textcolor{gray}{\pm 1.16}}$ & \underline{45.64}$_{\textcolor{gray}{\pm 1.17}}$
& \underline{11.66}$_{\textcolor{gray}{\pm 0.60}}$ & \underline{57.54}$_{\textcolor{gray}{\pm 1.17}}$ & \underline{45.88}$_{\textcolor{gray}{\pm 1.09}}$ \\

\rowcolor{blue!10}
CAAP (Ours)
& \underline{13.53}$_{\textcolor{gray}{\pm 0.58}}$ & \textbf{55.64}$_{\textcolor{gray}{\pm 1.19}}$ & \textbf{42.11}$_{\textcolor{gray}{\pm 1.04}}$
& \underline{15.61}$_{\textcolor{gray}{\pm 0.66}}$ & \textbf{64.85}$_{\textcolor{gray}{\pm 1.14}}$ & \textbf{49.24}$_{\textcolor{gray}{\pm 1.00}}$
& \textbf{20.33}$_{\textcolor{gray}{\pm 0.77}}$ & \textbf{67.43}$_{\textcolor{gray}{\pm 1.12}}$ & \textbf{47.10}$_{\textcolor{gray}{\pm 0.97}}$
& \textbf{9.53}$_{\textcolor{gray}{\pm 0.44}}$ & \textbf{57.62}$_{\textcolor{gray}{\pm 1.21}}$ & \textbf{48.08}$_{\textcolor{gray}{\pm 1.08}}$ \\

\bottomrule
\end{tabular}
}
\end{table*}

\begin{table*}[t]
\caption{Localization quantitative results for Input Insertion, Input Deletion, and CAAP on ResNet-18, ResNet-50, DenseNet-161, and VGG-16. AUPR$_1$ ($\uparrow$), AUPR$_0$ ($\uparrow$), and PG ($\uparrow$) are reported. Best is \textbf{bold}, second best is \underline{underlined} in each column.}
\vspace{-1.5mm}
\label{tab:segmentation_pg_cnn_comparison}
\centering
\small
\setlength{\tabcolsep}{4pt}
\renewcommand{\arraystretch}{1.15}
\resizebox{\textwidth}{!}{%
\begin{tabular}{lcccccccccccc}
\toprule
& \multicolumn{3}{c}{ResNet-18~\cite{resnet}}
& \multicolumn{3}{c}{ResNet-50~\cite{resnet}}
& \multicolumn{3}{c}{DenseNet-161~\cite{densenet}}
& \multicolumn{3}{c}{VGG-16~\cite{vgg}} \\
\cmidrule(lr){2-4}\cmidrule(lr){5-7}\cmidrule(lr){8-10}\cmidrule(lr){11-13}
Method
& AUPR$_1$ $\uparrow$ & AUPR$_0$ $\uparrow$ & PG $\uparrow$
& AUPR$_1$ $\uparrow$ & AUPR$_0$ $\uparrow$ & PG $\uparrow$
& AUPR$_1$ $\uparrow$ & AUPR$_0$ $\uparrow$ & PG $\uparrow$
& AUPR$_1$ $\uparrow$ & AUPR$_0$ $\uparrow$ & PG $\uparrow$ \\
\midrule

Input Insertion
& 36.91$_{\textcolor{gray}{\pm 2.23}}$ & 65.50$_{\textcolor{gray}{\pm 2.37}}$ & 47.66$_{\textcolor{gray}{\pm 4.41}}$
& 42.00$_{\textcolor{gray}{\pm 2.21}}$ & 69.03$_{\textcolor{gray}{\pm 2.28}}$ & 56.25$_{\textcolor{gray}{\pm 4.38}}$
& 39.70$_{\textcolor{gray}{\pm 2.24}}$ & 67.83$_{\textcolor{gray}{\pm 2.19}}$ & 49.22$_{\textcolor{gray}{\pm 4.42}}$
& 35.12$_{\textcolor{gray}{\pm 2.15}}$ & 64.95$_{\textcolor{gray}{\pm 2.31}}$ & 42.19$_{\textcolor{gray}{\pm 4.37}}$ \\

Input Deletion
& \underline{50.49}$_{\textcolor{gray}{\pm 2.20}}$ & \underline{72.09}$_{\textcolor{gray}{\pm 2.31}}$ & \underline{60.94}$_{\textcolor{gray}{\pm 4.31}}$
& \underline{49.40}$_{\textcolor{gray}{\pm 2.08}}$ & \underline{69.96}$_{\textcolor{gray}{\pm 2.35}}$ & \underline{64.06}$_{\textcolor{gray}{\pm 4.24}}$
& \underline{49.54}$_{\textcolor{gray}{\pm 2.14}}$ & \underline{71.87}$_{\textcolor{gray}{\pm 2.25}}$ & \underline{62.50}$_{\textcolor{gray}{\pm 4.28}}$
& \underline{51.61}$_{\textcolor{gray}{\pm 2.23}}$ & \underline{73.62}$_{\textcolor{gray}{\pm 2.32}}$ & \textbf{64.84}$_{\textcolor{gray}{\pm 4.22}}$ \\

\rowcolor{blue!10}
CAAP (Ours)
& \textbf{61.10}$_{\textcolor{gray}{\pm 2.18}}$ & \textbf{84.68}$_{\textcolor{gray}{\pm 1.80}}$ & \textbf{69.53}$_{\textcolor{gray}{\pm 4.07}}$
& \textbf{59.32}$_{\textcolor{gray}{\pm 2.07}}$ & \textbf{84.55}$_{\textcolor{gray}{\pm 1.82}}$ & \textbf{64.84}$_{\textcolor{gray}{\pm 4.22}}$
& \textbf{62.85}$_{\textcolor{gray}{\pm 2.22}}$ & \textbf{84.72}$_{\textcolor{gray}{\pm 1.85}}$ & \textbf{75.78}$_{\textcolor{gray}{\pm 3.79}}$
& \textbf{55.82}$_{\textcolor{gray}{\pm 2.07}}$ & \textbf{79.39}$_{\textcolor{gray}{\pm 2.10}}$ & \underline{62.50}$_{\textcolor{gray}{\pm 4.28}}$ \\

\bottomrule
\end{tabular}
}
\end{table*}

%% file: main.bib
@misc{vit_cx,
  title        = {ViT-CX: Causal Explanation of Vision Transformers},
  author       = {Xie, Weiyan and Li, Xiao-Hui and Cao, Caleb Chen and Zhang, Nevin L.},
  year         = {2023},
  eprint       = {2211.03064},
  archivePrefix= {arXiv},
  primaryClass = {cs.CV},
  url          = {https://arxiv.org/abs/2211.03064}
}

@inproceedings{mda,
  author    = {Walker, Chase and Jha, Sumit and Ewetz, Rickard},
  title     = {Metric-Driven Attributions for Vision Transformers},
  booktitle = {International Conference on Learning Representations (ICLR)},
  year      = {2025},
  url       = {https://proceedings.iclr.cc/paper_files/paper/2025/file/4e21153e79aff242492146d78d09fcdb-Paper-Conference.pdf}
}

@InProceedings{Mehri_2025_CVPR,
  author    = {Mehri, Faridoun and Baghshah, Mahdieh Soleymani and Pilehvar, Mohammad Taher},
  title     = {LibraGrad: Balancing Gradient Flow for Universally Better Vision Transformer Attributions},
  booktitle = {Proceedings of the IEEE/CVF Conference on Computer Vision and Pattern Recognition (CVPR)},
  month     = {June},
  year      = {2025},
  pages     = {67--78}
}

@inproceedings{mehri-skipplus-cvpr24,
  title     = {{SkipPLUS}: Skip the First Few Layers to Better Explain Vision Transformers},
  author    = {Mehri, Faridoun and Fayyaz, Mohsen and Baghshah, Mahdieh Soleymani and Pilehvar, Mohammad Taher},
  booktitle = {Proceedings of the IEEE/CVF Conference on Computer Vision and Pattern Recognition (CVPR) Workshops},
  year      = {2024},
  month     = {June},
  pages     = {204--215},
  doi       = {10.1109/CVPRW63382.2024.00025},
  url       = {https://openaccess.thecvf.com/content/CVPR2024W/TCV2024/html/Mehri_SkipPLUS_Skip_the_First_Few_Layers_to_Better_Explain_Vision_CVPRW_2024_paper.html}
}

@article{imagenet_s,
  title     = {Large-Scale Unsupervised Semantic Segmentation},
  author    = {Gao, Shanghua and Li, Zhong-Yu and Yang, Ming-Hsuan and Cheng, Ming-Ming and Han, Junwei and Torr, Philip},
  journal   = {IEEE Transactions on Pattern Analysis and Machine Intelligence},
  year      = {2023},
  volume    = {45},
  number    = {6},
  pages     = {7457--7476},
  doi       = {10.1109/TPAMI.2022.3218275},
  url       = {https://doi.org/10.1109/TPAMI.2022.3218275}
}

@misc{dino2,
  title        = {DINOv2: Learning Robust Visual Features without Supervision},
  author       = {Oquab, Maxime and Darcet, Timoth{\'e}e and Moutakanni, Th{\'e}o and Vo, Huy and Szafraniec, Marc and Khalidov, Vasil and Fernandez, Pierre and Haziza, Daniel and Massa, Francisco and El-Nouby, Alaaeldin and Assran, Mahmoud and Ballas, Nicolas and Galuba, Wojciech and Howes, Russell and Huang, Po-Yao and Li, Shang-Wen and Misra, Ishan and Rabbat, Michael and Sharma, Vasu and Synnaeve, Gabriel and Xu, Hu and Jegou, Herv{\'e} and Mairal, Julien and Labatut, Patrick and Joulin, Armand and Bojanowski, Piotr},
  year         = {2024},
  eprint       = {2304.07193},
  archivePrefix= {arXiv},
  primaryClass = {cs.CV},
  url          = {https://arxiv.org/abs/2304.07193}
}

@misc{clip,
  title        = {Learning Transferable Visual Models From Natural Language Supervision},
  author       = {Radford, Alec and Kim, Jong Wook and Hallacy, Chris and Ramesh, Aditya and Goh, Gabriel and Agarwal, Sandhini and Sastry, Girish and Askell, Amanda and Mishkin, Pamela and Clark, Jack and Krueger, Gretchen and Sutskever, Ilya},
  year         = {2021},
  eprint       = {2103.00020},
  archivePrefix= {arXiv},
  primaryClass = {cs.CV},
  url          = {https://arxiv.org/abs/2103.00020}
}

@misc{russakovsky2015imagenetlargescalevisual,
  title        = {ImageNet Large Scale Visual Recognition Challenge},
  author       = {Russakovsky, Olga and Deng, Jia and Su, Hao and Krause, Jonathan and Satheesh, Sanjeev and Ma, Sean and Huang, Zhiheng and Karpathy, Andrej and Khosla, Aditya and Bernstein, Michael and Berg, Alexander C. and Fei-Fei, Li},
  year         = {2015},
  eprint       = {1409.0575},
  archivePrefix= {arXiv},
  primaryClass = {cs.CV},
  url          = {https://arxiv.org/abs/1409.0575}
}

@inproceedings{Petsiuk2018RISE,
  title     = {RISE: Randomized Input Sampling for Explanation of Black-box Models},
  author    = {Petsiuk, Vitali and Das, Abir and Saenko, Kate},
  booktitle = {BMVC},
  year      = {2018}
}

@inproceedings{Margolin2014Foreground,
  title     = {How to Evaluate Foreground Maps?},
  author    = {Margolin, Ran and Zelnik-Manor, Lihi and Tal, Ayellet},
  booktitle = {CVPR},
  year      = {2014}
}

@misc{vit,
  title        = {How to train your ViT? Data, Augmentation, and Regularization in Vision Transformers},
  author       = {Steiner, Andreas and Kolesnikov, Alexander and Zhai, Xiaohua and Wightman, Ross and Uszkoreit, Jakob and Beyer, Lucas},
  year         = {2021},
  eprint       = {2106.10270},
  archivePrefix= {arXiv},
  primaryClass = {cs.CV},
  url          = {https://arxiv.org/abs/2106.10270}
}

@misc{deit,
  title        = {DeiT III: Revenge of the ViT},
  author       = {Touvron, Hugo and Cord, Matthieu and J{\'e}gou, Herv{\'e}},
  year         = {2022},
  eprint       = {2204.07118},
  archivePrefix= {arXiv},
  primaryClass = {cs.CV},
  url          = {https://arxiv.org/abs/2204.07118}
}

@misc{imagenet_hard,
  title        = {ImageNet-Hard: The Hardest Images Remaining from a Study of the Power of Zoom and Spatial Biases in Image Classification},
  author       = {Taesiri, Mohammad Reza and Nguyen, Giang and Habchi, Sarra and Bezemer, Cor-Paul and Nguyen, Anh},
  year         = {2023},
  eprint       = {2304.05538},
  archivePrefix= {arXiv},
  primaryClass = {cs.CV},
  url          = {https://arxiv.org/abs/2304.05538}
}

@InProceedings{oxford_pet,
  author    = {Parkhi, Omkar M. and Vedaldi, Andrea and Zisserman, Andrew and Jawahar, C. V.},
  title     = {Cats and Dogs},
  booktitle = {IEEE Conference on Computer Vision and Pattern Recognition (CVPR)},
  year      = {2012}
}

@article{samek2021explaining,
  title     = {Explaining deep neural networks and beyond: A review of methods and applications},
  author    = {Samek, Wojciech and Montavon, Gr{\'e}goire and Lapuschkin, Sebastian and Anders, Christopher J. and M{\"u}ller, Klaus-Robert},
  journal   = {Proceedings of the IEEE},
  year      = {2021},
  volume    = {109},
  number    = {3},
  pages     = {247--278},
  publisher = {IEEE}
}

@misc{shrikumar2016not,
  title        = {Not Just a Black Box: Learning Important Features Through Propagating Activation Differences},
  author       = {Shrikumar, Avanti and Greenside, Peyton and Shcherbina, Anna and Kundaje, Anshul},
  year         = {2016},
  eprint       = {1605.01713},
  archivePrefix= {arXiv},
  primaryClass = {cs.LG},
  url          = {https://arxiv.org/abs/1605.01713}
}

@inproceedings{sundararajan2017axiomatic,
  title     = {Axiomatic Attribution for Deep Networks},
  author    = {Sundararajan, Mukund and Taly, Ankur and Yan, Qiqi},
  booktitle = {Proceedings of the 34th International Conference on Machine Learning (ICML)},
  year      = {2017},
  pages     = {3319--3328},
  organization={PMLR}
}

@article{grad_cam,
  title     = {Grad-CAM: Visual Explanations from Deep Networks via Gradient-Based Localization},
  author    = {Selvaraju, Ramprasaath R. and Cogswell, Michael and Das, Abhishek and Vedantam, Ramakrishna and Parikh, Devi and Batra, Dhruv},
  journal   = {International Journal of Computer Vision},
  year      = {2019},
  volume    = {128},
  number    = {2},
  pages     = {336--359},
  doi       = {10.1007/s11263-019-01228-7},
  url       = {https://doi.org/10.1007/s11263-019-01228-7}
}

@inproceedings{chefer2021transformer,
  title     = {Transformer Interpretability Beyond Attention Visualization},
  author    = {Chefer, Hila and Gur, Shir and Wolf, Lior},
  booktitle = {Proceedings of the IEEE/CVF Conference on Computer Vision and Pattern Recognition (CVPR)},
  year      = {2021},
  pages     = {782--791}
}

@inproceedings{qiang2022attcat,
  title     = {AttCAT: Explaining Transformers via Attentive Class Activation Tokens},
  author    = {Qiang, Yao and Pan, Deng and Li, Chengyin and Li, Xin and Jang, Rhongho and Zhu, Dongxiao},
  booktitle = {Advances in Neural Information Processing Systems (NeurIPS)},
  year      = {2022}
}

@inproceedings{Jain2019Attention,
  title     = {Attention is not Explanation},
  author    = {Jain, Sarthak and Wallace, Byron C.},
  booktitle = {Proceedings of NAACL},
  year      = {2019},
  url       = {https://aclanthology.org/N19-1357/}
}

@inproceedings{kobayashi2020attention,
  title     = {Attention is not only a weight: Analyzing transformers with vector norms},
  author    = {Kobayashi, Goro and Kuribayashi, Tatsuki and Yokoi, Sho and Inui, Kentaro},
  booktitle = {Proceedings of EMNLP},
  year      = {2020},
  pages     = {7057--7075}
}

@inproceedings{attn_r,
  title     = {Quantifying Attention Flow in Transformers},
  author    = {Abnar, Samira and Zuidema, Willem},
  booktitle = {Proceedings of the 58th Annual Meeting of the Association for Computational Linguistics},
  year      = {2020},
  pages     = {4190--4197},
  doi       = {10.18653/v1/2020.acl-main.385},
  url       = {https://aclanthology.org/2020.acl-main.385/}
}

@inproceedings{tam,
  title     = {Explaining Information Flow Inside Vision Transformers Using Markov Chain},
  author    = {Yuan, Tingyi and Li, Xuhong and Xiong, Haoyi and Cao, Hui and Dou, Dejing},
  booktitle = {NeurIPS 2021 Workshop on eXplainable AI Approaches for Debugging and Diagnosis (XAI4Debugging)},
  year      = {2021},
  url       = {https://openreview.net/forum?id=TT-cf6QSDaQ}
}

@article{mutex,
  author    = {Marchetti, Michele and Traini, Davide and Ursino, Domenico and Virgili, Luca},
  title     = {Multiplex network-based representation of vision transformers for visual explainability},
  journal   = {Neural Computing and Applications},
  year      = {2025},
  volume    = {37},
  number    = {29},
  pages     = {24385--24420},
  doi       = {10.1007/s00521-025-11591-x},
  url       = {https://doi.org/10.1007/s00521-025-11591-x}
}

@article{deb2023atman,
  title   = {AtMan: Understanding Transformer Predictions through Memory Efficient Attention Manipulation},
  author  = {Deb, Mayukh and Deiseroth, Bj{\"o}rn and Weinbach, Samuel and Schramowski, Patrick and Kersting, Kristian},
  journal = {arXiv preprint arXiv:2301.08110},
  year    = {2023},
  url     = {https://arxiv.org/abs/2301.08110}
}

@inproceedings{fong2017meaningful,
  title     = {Interpretable Explanations of Black Boxes by Meaningful Perturbation},
  author    = {Fong, Ruth C. and Vedaldi, Andrea},
  booktitle = {Proceedings of ICCV},
  year      = {2017},
  url       = {https://www.robots.ox.ac.uk/~vgg/publications/2017/Fong17/}
}

@inproceedings{fong2019extremal,
  title     = {Understanding Deep Networks via Extremal Perturbations and Smooth Masks},
  author    = {Fong, Ruth C. and Patrick, Mandela and Vedaldi, Andrea},
  booktitle = {Proceedings of ICCV},
  year      = {2019}
}

@inproceedings{chattopadhyay2019causal,
  title     = {Neural Network Attributions: A Causal Perspective},
  author    = {Chattopadhyay, Aditya and Sarkar, Anirban and Howlader, Prantik and Balasubramanian, Vineeth N.},
  booktitle = {Proceedings of ICML},
  year      = {2019},
  url       = {https://proceedings.mlr.press/v97/chattopadhyay19a.html}
}

@inproceedings{ZhangNanda2024Patching,
  title     = {Towards Best Practices of Activation Patching in Language Models: Metrics and Methods},
  author    = {Zhang, Fred and Nanda, Neel},
  booktitle = {International Conference on Learning Representations (ICLR)},
  year      = {2024},
  url       = {https://openreview.net/forum?id=Hf17y6u9BC}
}

@misc{dosovitskiy2021imageworth16x16words,
  title        = {An Image is Worth 16x16 Words: Transformers for Image Recognition at Scale},
  author       = {Dosovitskiy, Alexey and Beyer, Lucas and Kolesnikov, Alexander and Weissenborn, Dirk and Zhai, Xiaohua and Unterthiner, Thomas and Dehghani, Mostafa and Minderer, Matthias and Heigold, Georg and Gelly, Sylvain and Uszkoreit, Jakob and Houlsby, Neil},
  year         = {2021},
  eprint       = {2010.11929},
  archivePrefix= {arXiv},
  primaryClass = {cs.CV},
  url          = {https://arxiv.org/abs/2010.11929}
}

@misc{smilkov2017smoothgrad,
  title        = {SmoothGrad: removing noise by adding noise},
  author       = {Smilkov, Daniel and Thorat, Nikhil and Kim, Been and Vi{\'e}gas, Fernanda and Wattenberg, Martin},
  year         = {2017},
  eprint       = {1706.03825},
  archivePrefix= {arXiv},
  primaryClass = {cs.LG},
  url          = {https://arxiv.org/abs/1706.03825}
}

@misc{chattopadhyay2017gradcampp,
  title        = {Grad-CAM++: Improved Visual Explanations for Deep Convolutional Networks},
  author       = {Chattopadhyay, Aditya and Sarkar, Anirban and Howlader, Prantik and Balasubramanian, Vineeth N.},
  year         = {2017},
  eprint       = {1710.11063},
  archivePrefix= {arXiv},
  primaryClass = {cs.CV},
  url          = {https://arxiv.org/abs/1710.11063}
}

@inproceedings{wang2020scorecam,
  title     = {Score-CAM: Score-Weighted Visual Explanations for Convolutional Neural Networks},
  author    = {Wang, Haofan and Wang, Zifan and Du, Mengnan and Yang, Fan and Zhang, Zijian and Ding, Sirui and Mardziel, Piotr and Hu, Xia},
  booktitle = {Proceedings of the IEEE/CVF Conference on Computer Vision and Pattern Recognition Workshops (CVPRW)},
  year      = {2020}
}

@inproceedings{Englebert_2023_ICCV,
  title     = {Explaining Through Transformer Input Sampling},
  author    = {Englebert, Alexandre and Stassin, S{\'e}drick and Nanfack, G{\'e}raldin and Mahmoudi, Sidi Ahmed and Siebert, Xavier and Cornu, Olivier and De Vleeschouwer, Christophe},
  booktitle = {Proceedings of the IEEE/CVF International Conference on Computer Vision Workshops (ICCVW)},
  year      = {2023},
  url       = {https://openaccess.thecvf.com/content/ICCV2023W/NIVT/html/Englebert_Explaining_Through_Transformer_Input_Sampling_ICCVW_2023_paper.html}
}

@inproceedings{barkan2021gradsam,
  title     = {Grad-SAM: Explaining Transformers via Gradient Self-Attention Maps},
  author    = {Barkan, Oren and Hauon, Edan and Caciularu, Avi and Katz, Ori and Malkiel, Itzik and Armstrong, Omri and Koenigstein, Noam},
  booktitle = {Proceedings of the 30th ACM International Conference on Information and Knowledge Management (CIKM)},
  year      = {2021},
  pages     = {2882--2887},
  doi       = {10.1145/3459637.3482126}
}

@misc{bearman2016whatspointsemanticsegmentation,
  title        = {What's the Point: Semantic Segmentation with Point Supervision},
  author       = {Bearman, Amy and Russakovsky, Olga and Ferrari, Vittorio and Fei-Fei, Li},
  year         = {2016},
  eprint       = {1506.02106},
  archivePrefix= {arXiv},
  primaryClass = {cs.CV},
  url          = {https://arxiv.org/abs/1506.02106}
}

@inproceedings{
pan2024dissecting,
title={Dissecting Query-Key Interaction in Vision Transformers},
author={Xu Pan and Aaron Philip and Ziqian Xie and Odelia Schwartz},
booktitle={The Thirty-eighth Annual Conference on Neural Information Processing Systems},
year={2024},
url={https://openreview.net/forum?id=dIktpSgK4F}
}

@inproceedings{food101,
  title = {Food-101 -- Mining Discriminative Components with Random Forests},
  author = {Bossard, Lukas and Guillaumin, Matthieu and Van Gool, Luc},
  booktitle = {European Conference on Computer Vision},
  year = {2014}
}

@article{resnet,
  title={Deep Residual Learning for Image Recognition},
  author={Kaiming He and X. Zhang and Shaoqing Ren and Jian Sun},
  journal={2016 IEEE Conference on Computer Vision and Pattern Recognition (CVPR)},
  year={2015},
  pages={770-778},
  url={https://api.semanticscholar.org/CorpusID:206594692}
}

@article{vgg,
  title={Very Deep Convolutional Networks for Large-Scale Image Recognition},
  author={Karen Simonyan and Andrew Zisserman},
  journal={CoRR},
  year={2014},
  volume={abs/1409.1556},
  url={https://api.semanticscholar.org/CorpusID:14124313}
}

@article{densenet,
  title={Densely Connected Convolutional Networks},
  author={Gao Huang and Zhuang Liu and Kilian Q. Weinberger},
  journal={2017 IEEE Conference on Computer Vision and Pattern Recognition (CVPR)},
  year={2016},
  pages={2261-2269},
  url={https://api.semanticscholar.org/CorpusID:9433631}
}

@article{causality,
  author  = {Atticus Geiger and Duligur Ibeling and Amir Zur and Maheep Chaudhary and Sonakshi Chauhan and Jing Huang and Aryaman Arora and Zhengxuan Wu and Noah Goodman and Christopher Potts and Thomas Icard},
  title   = {Causal Abstraction: A Theoretical Foundation for Mechanistic Interpretability},
  journal = {Journal of Machine Learning Research},
  year    = {2025},
  volume  = {26},
  number  = {83},
  pages   = {1--64},
  url     = {http://jmlr.org/papers/v26/23-0058.html}
}

@inproceedings{bao2022beit,
  title     = {BEiT: BERT Pre-Training of Image Transformers},
  author    = {Bao, Hangbo and Dong, Li and Piao, Songhao and Wei, Furu},
  booktitle = {International Conference on Learning Representations},
  year      = {2022}
}

@inproceedings{Zhai_2023_ICCV,
  title     = {Sigmoid Loss for Language Image Pre-Training},
  author    = {Zhai, Xiaohua and Mustafa, Basil and Kolesnikov, Alexander and Beyer, Lucas},
  booktitle = {Proceedings of the IEEE/CVF International Conference on Computer Vision (ICCV)},
  pages     = {11975--11986},
  year      = {2023}
}

@inproceedings{yeh2019infidelity,
  title     = {On the (In)fidelity and Sensitivity of Explanations},
  author    = {Yeh, Chih-Kuan and Hsieh, Cheng-Yu and
               Suggala, Arun Sai and Inouye, David I. and
               Ravikumar, Pradeep},
  booktitle = {Advances in Neural Information Processing Systems},
  volume    = {32},
  year      = {2019}
}
